\documentclass[12pt,twoside]{report}

\newcommand{\reporttitle}{Bayesian Mixture-of-Experts: Towards Making LLMs Know What They Don't Know}
\newcommand{\reportauthor}{Albus Yizhuo Li}
\newcommand{\supervisor}{Dr Matthew Wicker}
\newcommand{\secondmarker}{Dr Yingzhen Li}
\newcommand{\degreetype}{\textit{Computing (Artificial Intelligence and Machine Learning)}}

\usepackage[a4paper,hmargin=2cm,vmargin=1.0cm,includeheadfoot]{geometry}
\usepackage{textpos}
\usepackage{tabularx,longtable,multirow,caption}
\usepackage{fncylab} 
\usepackage{fancyhdr} 
\usepackage{url} 
\usepackage[english]{babel}
\usepackage{graphicx}
\usepackage{dsfont}
\usepackage{epstopdf} 
\usepackage{backref} 
\usepackage{array}
\usepackage{latexsym}
\usepackage[pdftex,pagebackref,hypertexnames=false,colorlinks]{hyperref}
\usepackage{xcolor}
\usepackage{geometry}
\usepackage{graphicx}
\usepackage{titlesec}
\usepackage{hyperref}
\usepackage{amsfonts}
\usepackage{bbm}
\usepackage{float}
\usepackage{tikz}
\usepackage{bm}
\usepackage[square,numbers]{natbib}
\usepackage{tikz}
\usepackage{amsmath, amssymb, amsthm, mathtools}
\usepackage{booktabs} 
\usepackage{wrapfig}      
\usepackage{multirow} 
\usepackage{siunitx}
\usepackage{lipsum}
\usepackage[table]{xcolor}
\usepackage{caption}
\usepackage{subcaption}
\usepackage{array} 
\usepackage{multirow} 
\usepackage{algorithm}
\usepackage[noend]{algpseudocode}
\usepackage{siunitx}
\definecolor{diffcolor}{HTML}{C00000} 

\usetikzlibrary{positioning, arrows.meta, fit, shapes.geometric}
\usepackage[colorlinks]{hyperref} 

\usepackage{enumitem}
\setlist[itemize]{leftmargin=*}
\setlist[enumerate]{leftmargin=*}

\RequirePackage{booktabs} 
\RequirePackage{caption}  
\RequirePackage{longtable} 
\RequirePackage{pdflscape}
\RequirePackage{makecell}

\newcommand{\KL}{D_{\mathbb{KL}}} 

\usetikzlibrary{shapes.geometric, arrows.meta, positioning, calc}
\definecolor{trunk}{RGB}{245,180,180}   
\definecolor{headA}{RGB}{175,200,245}   
\definecolor{headB}{RGB}{180,215,180}   
\definecolor{gaussFill}{gray}{0.85}     

\definecolor{darkblue}{RGB}{0, 51, 102}
\definecolor{ImperialBlue}{RGB}{0, 0, 205}
\definecolor{lightgray}{RGB}{100, 100, 100}

\newcommand{\impb}[1]{\textcolor{ImperialBlue}{#1}}

\hypersetup{pdftitle={},
  pdfsubject={}, 
  pdfauthor={},
  pdfkeywords={}, 
  pdfstartview=FitH,
  pdfpagemode={UseOutlines},
  bookmarksnumbered=true, bookmarksopen=true, colorlinks,
    citecolor=black,%
    filecolor=black,%
    linkcolor=black,%
    urlcolor=black}

\usepackage[all]{hypcap}

\usepackage{tcolorbox}
\tcbuselibrary{skins, breakable}
\newtcolorbox{authorcommentbox}{
    enhanced,
    breakable,
    colback=green!10,
    colframe=black!60,
    boxrule=0.5pt,
    left=2mm,
    right=2mm,
    top=1mm,
    bottom=1mm,
    rounded corners,
    before skip=0pt, after skip=10pt,
}
\newtcolorbox{matthewfeedbackbox}{
    enhanced,
    breakable,
    colback=blue!10,
    colframe=black!60,
    boxrule=0.5pt,
    left=2mm,
    right=2mm,
    top=1mm,
    bottom=1mm,
    rounded corners,
    before skip=0pt, after skip=10pt,
}

\usepackage{libertinus}

\def\@makechapterhead#1{%
  \vspace*{10\p@}%
  {\parindent \z@ \raggedright \sffamily
    \interlinepenalty\@M
    \Huge\bfseries \thechapter \space\space #1\par\nobreak
    \vskip 30\p@
  }}

\def\@makeschapterhead#1{%
  \vspace*{10\p@}%
  {\parindent \z@ \raggedright \sffamily
    \interlinepenalty\@M
    \Huge \bfseries  #1\par\nobreak
    \vskip 30\p@
  }}

\allowdisplaybreaks

\date{September 2025}

\begin{document}

\begin{titlepage}

\newcommand{\HRule}{\rule{\linewidth}{0.5mm}} 


\includegraphics[width = 7cm]{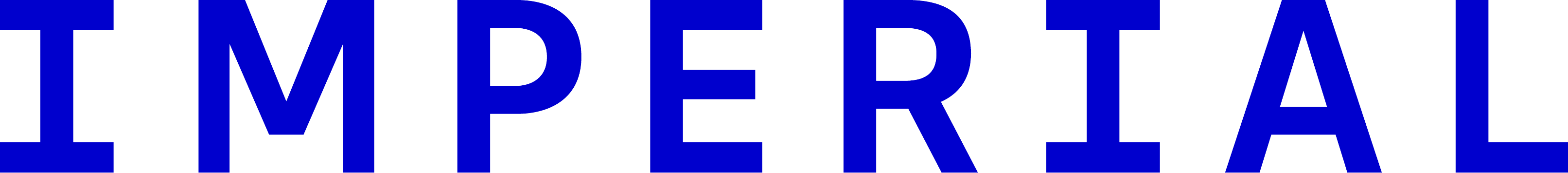}\\[0.5cm] 

\center 


\textsc{\Large Imperial College London}\\[0.5cm] 
\textsc{\large Department of Computing}\\[0.5cm] 


\HRule \\[0.4cm]
{ \huge \bfseries \reporttitle}\\ 
\HRule \\[1.5cm]
 

\begin{minipage}{0.4\textwidth}
\begin{flushleft} \large
\emph{Author:}\\
\reportauthor 
\end{flushleft}
\end{minipage}
~
\begin{minipage}{0.4\textwidth}
\begin{flushright} \large
\emph{Supervisor:} \\
\supervisor\\
\vspace{5pt}
\emph{Second Marker:}\\
\secondmarker 
\end{flushright}
\end{minipage}\\[2.5cm]

\begin{center}
    \includegraphics[width=5cm]{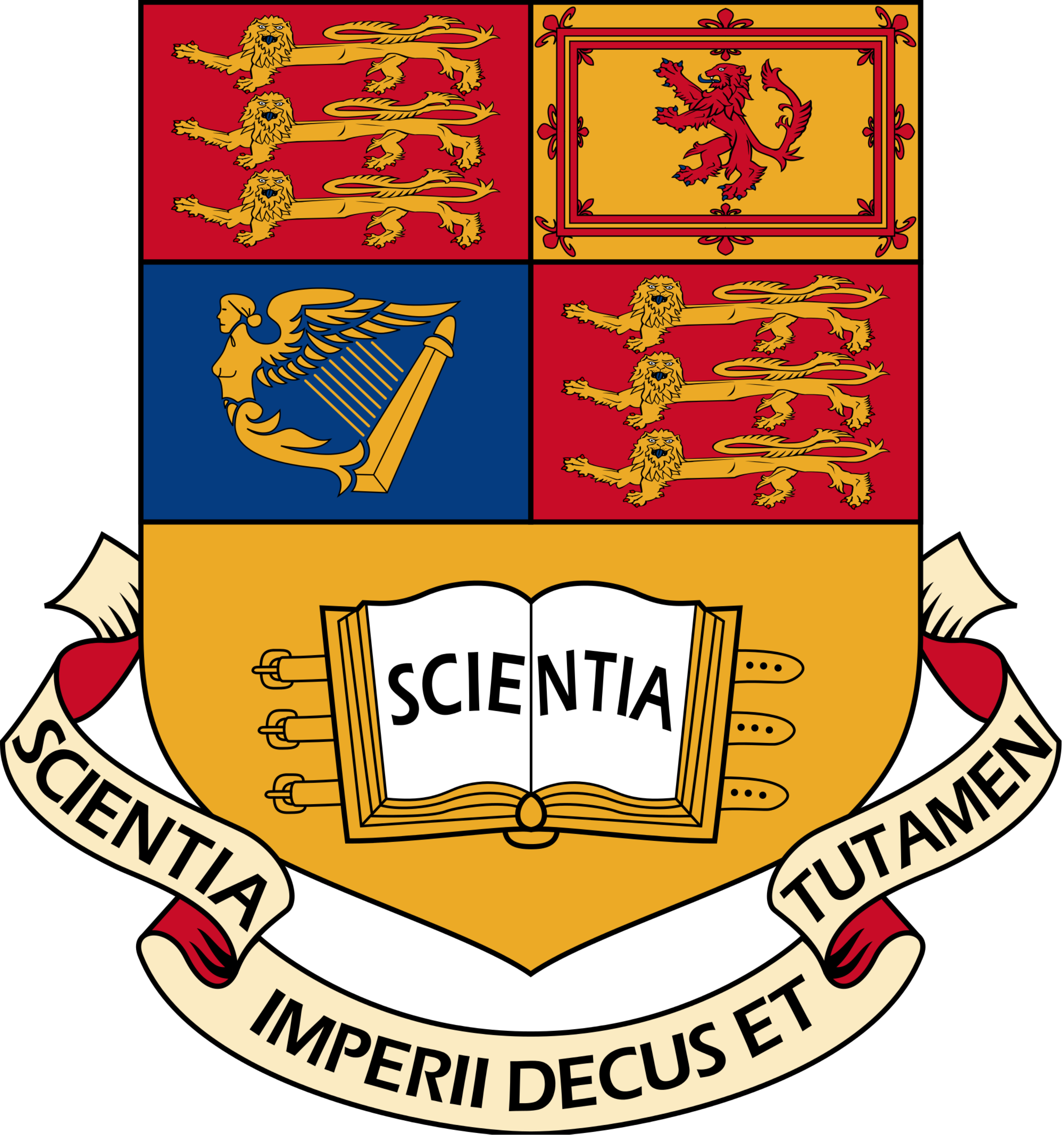}\\[0.5cm]
\end{center}

\vfill 
Submitted in partial fulfillment of the requirements for the MSc degree in
\degreetype~of Imperial College London\\[0.5cm]

\makeatletter
\@date
\makeatother

\end{titlepage}

\pagenumbering{roman}
\clearpage{\pagestyle{empty}\cleardoublepage}
\setcounter{page}{1}
\pagestyle{fancy}

\begin{abstract}
The Mixture-of-Experts (MoE) architecture has enabled the creation of massive yet efficient Large Language Models (LLMs). 
However, the standard deterministic routing mechanism presents a significant limitation: 
its inherent brittleness is a key contributor to model miscalibration and overconfidence, 
resulting in systems that often do not know what they don't know.

This thesis confronts this challenge by proposing a structured \textbf{Bayesian MoE routing framework}. 
Instead of forcing a single, deterministic expert selection, our approach models a probability distribution over the routing decision itself. 
We systematically investigate three families of methods that introduce this principled uncertainty at different stages of the routing pipeline: 
in the \textbf{weight-space}, the \textbf{logit-space}, and the final \textbf{selection-space}.

Through a series of controlled experiments on a 3-billion parameter MoE model, 
we demonstrate that this framework significantly improves routing stability, in-distribution calibration, and out-of-distribution (OoD) detection. 
The results show that by targeting this core architectural component, we can create a more reliable internal uncertainty signal. 
This work provides a practical and computationally tractable pathway towards building more robust and self-aware LLMs, 
taking a crucial step towards making them know what they don't know.
\end{abstract}

\clearpage
\section*{Acknowledgments}

\begin{center}
\textit{This thesis is dedicated to my demanding, fulfilling and joyous year at Imperial College London,\\ my Hogwarts.}
\end{center}

This journey to this thesis was made possible by the support, guidance, and inspiration of many people, to whom I owe my deepest gratitude:
\vspace{0.4cm}

First and foremost, I would like to express my sincere gratitude to my supervisor, \textit{Dr. Matthew Wicker}. 
His amazing \textit{70015: Mathematics for Machine Learning} module lured me down the rabbit hole of Probabilistic \& Bayesian Machine Learning, a journey from which I have happily not returned. 
His initial ideation of Bayesianfying Mixture-of-Experts provides the foundation of this thesis.
Since mid-stage of this project, his careful guidance and detailed feedback on both experiments and writing were invaluable. 
Thank you for being a great supervisor and friend.
\vspace{0.4cm}

My thanks also extend to my second marker, \textit{Dr. Yingzhen Li}, 
whose lecture notes on Variational Inference and Introduction to BNNs are the best I have ever seen. 
I am grateful for her interest in this project and for the insightful meeting she arranged with her PhD student, Wenlong, which provided crucial perspective at a key stage.
\vspace{0.4cm}

The work was sharpened by the weekly discussions of \textit{LLM Shilling Crew}, 
a reading group I had the pleasure of co-founding with my best friend at Imperial, \textit{James Kerns}. 
Thank you all for the stimulating discussion and the fun we had, which were instrumental during the early research phase of this project.
\vspace{0.4cm}

To my parents, Yuhan and Wei, thank you for the unconditional love and the unwavering financial and emotional support you have provided for the past 22 years.
\vspace{0.4cm}

Last but certainly not least, I must thank my close friends at the Department of Computing, 
fellow habitants of the deep, dark, and cold basement of the Huxley building (you know who you are). 
You are a priceless treasure in my life.

\clearpage{}
\setcounter{tocdepth}{2}
\fancyhead[RE,LO]{\sffamily {Table of Contents}}
\tableofcontents 

\clearpage{\pagestyle{empty}\cleardoublepage}
\pagenumbering{arabic}
\setcounter{page}{1}
\fancyhead[LE,RO]{\slshape \rightmark}
\fancyhead[LO,RE]{}
\fancyhead[LO,RE]{\slshape \leftmark}

\chapter{Introduction}
\label{chap:introduction}

\section{Overview}

Modern Large Language Models (LLMs) have achieved remarkable success through clever techniques for scaling both dataset and model size. 
A key architectural innovation enabling this progress is the Mixture-of-Experts (MoE) model \cite{shazeer2017outrageously, lepikhin2020gshard}. 
The computational cost of dense, all-parameter activation in traditional LLMs creates a bottleneck that limits further scaling and hinders wider, more accessible deployment. 
The MoE architecture elegantly circumvents this by using a routing network (gating network) to activate only a fraction of the model's parameters for any given input. 
This sparsity allows for a massive increase in the total number of parameters, enhancing the model's capacity for specialised knowledge without a proportional increase in computational cost. 
This dual benefit of specilisation and sparsity has made MoE a cornerstone of state-of-the-art LLMs.

Despite their power, the practical deployment of LLMs is hindered by fundamental challenges in robustness and calibration \cite{guo2017calibration}. 
These models often produce highly confident yet incorrect outputs, a phenomenon known as overconfidence, which has been shown to be a persistent issue across a wide range of models and tasks \cite{mielke2022reducing}. 
This unreliability frequently manifests as hallucination, the generation of plausible but factually fictitious content, 
which poses a significant barrier to their adoption in high-stake domains \cite{ji2023survey}, such as medicine and the law. 
At its core, this untrustworthiness stems from the models' inability to quantify their own predictive uncertainty.

This thesis argues that in an MoE model, the classic deterministic routing mechanism represents a critical point of failure. 
The router's decision is not a minor adjustment, but dictates which specialised sub-networks are activated for inference. 
An incorrect or brittle routing choice means the wrong knowledge-domain expert is applied to a token, leading to a flawed output. 
In modern LLMs with dozens of stacked MoE layers, this problem is magnified: A single routing error in an early layer creates a corrupted representation that is then passed to all subsequent layers, initiating a cascading failure.

This thesis proposes to address potential failure mode by introducing a \textbf{Bayesian routing framework}. 
Instead of forcing the router to make a single, deterministic choice, our approach is to model a probability distribution over the routing decisions themselves. 
This allows us to perform principled uncertainty quantification directly at the point of expert selection, drawing on foundational concepts in Bayesian deep learning~\cite{blundell2015weight, bishop2006pattern, murphy2024pml}. 
While applying Bayesian methods to an entire multi-billion parameter LLM is often computationally daunting, focusing this treatment only on the lightweight routing networks is a highly pragmatic and tractable approach. The ultimate purpose is to leverage this targeted uncertainty to enable better calibrated and robust LLM inference, creating models that are not only powerful but also aware of the limits of their own knowledge.

\section{Contributions}

This thesis makes the following primary contributions to the study of reliable and calibrated Mixture-of-Experts models:

\begin{enumerate}
    \setlength\itemsep{0.05em}
    \item \textbf{Diagnosis of Router Brittleness and Rationale for Probabilistic Routing:} 
          We establish the empirical foundation for this thesis with a two-part investigation, which reveals the inherent brittleness of standard deterministic routing and potentials for probablistic approaches respectively.
    \item \textbf{A Structured Framework for Bayesian Routing:} We formulate and evaluate a novel framework that categorises Bayesian routing methods based on where uncertainty is introduced. This taxonomy provides a clear and structured landscape for analysis, 
          focussed on Bayesian modelling of weight-space, logit-space and routing-space respectively.
    \item \textbf{Rigorous Evaluation of Calibration and Robustness:} 
    We conduct a series of controlled experiments on a pre-trained MoE model with 3B parameters, then systematically measure the impact of our proposed methods on in-distribution (ID) performance and calibration, out-of-distribution (OoD) detection, and overall router stability.
    \item \textbf{Memory and Computation Overhead Analysis:} We assess the practical feasibility of the proposed Bayesian routing methods by performing a detailed analysis of their memory and computational overhead. This provides a clear picture of the trade-offs involved, demonstrating which methods are most viable for deployment in large-scale systems.
\end{enumerate}

\section{Thesis Outline}
\label{sec:outline}

The remainder of this thesis is organised as follows. 
\textbf{Chapter 2} provides a review of the foundational literature on Mixture-of-Experts models, uncertainty in LLMs, and Bayesian machine learning. 
\textbf{Chapter 3} presents the motivational experiments that quantitatively establish the problem of router instability. 
\textbf{Chapter 4} details the methodology behind our proposed Bayesian Routing Networks framework. 
\textbf{Chapter 5} is dedicated to the main experiments and analysis, evaluating the impact of our methods on stability, calibration, and robustness, with further efficiency analysis. 
Finally, \textbf{Chapter 6} concludes the thesis with a discussion that includes the limitations of this study, and promising directions for future work.
\chapter{Background}
\label{chap:background}

\section{Mixture-of-Experts (MoE) Architecture}
\label{sec:background_moe}

\subsection{Modern LLM: A Primer}
\label{sec:background_primer}

To understand the innovation of the Mixture-of-Experts (MoE) architecture, one must first understand the standard model it enhances. 
The foundational architecture for virtually all modern Large Language Models (LLMs) is the Transformer~\cite{vaswani2017attention}. 
This section provides a brief but essential overview of the key components of a contemporary, dense LLM, establishing a baseline before we introduce the concept of sparsity.

\subsubsection{Decoder-Only Transformer Blueprint}

The dominant architecture for modern generative LLMs, such as those in the GPT family~\cite{radford2018improving}, is the Decoder-only Transformer~\cite{brown2020language}. As illustrated in Figure~\ref{fig:docoder_stack_to_transformer_block}~(A), this model processes text through a sequential pipeline. 
The process begins with an input sequence of tokens, which are represented in the form of indices from the vocabulary by Tokeniser. 
These discrete IDs are first converted into continuous vector representations by an Embedding layer, which is a learnable lookup table.
Positional encoding is also usually incorporated at the embedding stage.

The resulting embeddings are then processed by the core of the model: 
a stack of $N$ identical \textbf{Decoder Layers}. 
The output of one layer serves as the input to the next, allowing the model to build progressively more abstract and contextually rich representations of the sequence. 
After the final decoder layer, a concluding LayerNorm is applied. 
This final hidden state is then projected into the vocabulary space by a linear layer known as the Language Modelling Head~\cite{hf_lm_head}, which produces a logit for every possible token from the vocabulary. 
Finally, a softmax function is applied to these logits to generate a probability distribution, from which the output Token ID is predicted. 
Each of these decoder blocks contains the same set of internal sub-layers, which we will describe next.

\subsubsection{Inside the Transformer Block}

\begin{figure}[H]
    \centering
    \includegraphics[width=0.6\textwidth]{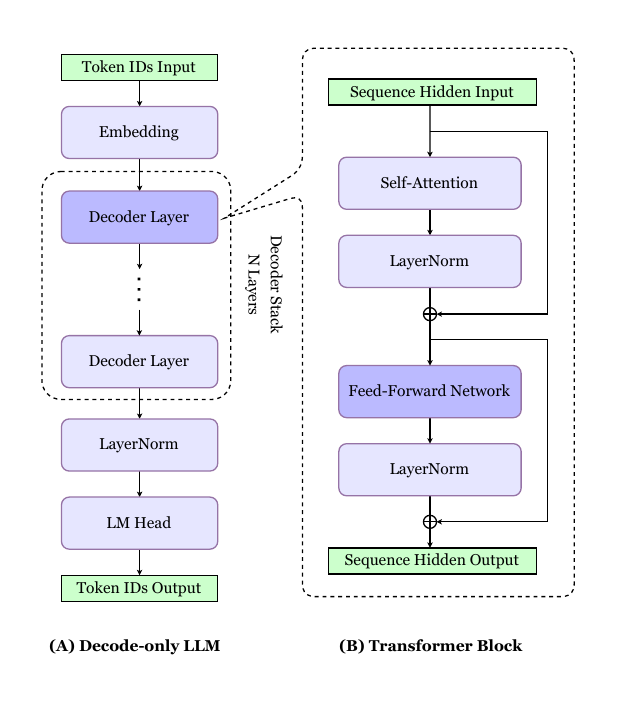}
    \caption{From Decoder-only LLM to Transformer Block. 
    (A) The high-level of a decoder-only LLM, composed of a stack of identical Transformer blocks.
    (B) The internal structure of a single Transformer block.
    }
    \label{fig:docoder_stack_to_transformer_block}
\end{figure}

As shown in Figure~\ref{fig:docoder_stack_to_transformer_block}~(B), each identical decoder block is composed of two primary sub-layers, wrapped with essential components that enable stable training of deep networks.

The first sub-layer is the \textbf{Multi-Head Self-Attention} mechanism. This is the core innovation of the Transformer, allowing each token to weigh the importance of all other preceding tokens in the sequence. The output of this sub-layer, $\mathbf{u}$, is computed by applying the attention function to the block's input, $\mathbf{h}$, with residual connection and layer normalisation added:
\begin{equation}
    \mathbf{u} = \text{LayerNorm}(\text{SA}(\mathbf{h}) + \mathbf{h})
    \label{eq:sa_layer}
\end{equation}

As the attention mechanism is not the primary focus of this thesis, we will not detail its internal mechanics.

The second sub-layer is a position-wise \textbf{Feed-Forward Network (FFN)}. 
This is a non-linear transformation that is applied independently to each token representation $\mathbf{u}_t$ after it has been updated by the attention mechanism.
Skip connections and layer normalisation are again applied, yielding the final output of the Transformer block, $\mathbf{h'}$: 
\begin{equation}
\mathbf{h'} = \text{LayerNorm}(\text{FFN}(\mathbf{u}) + \mathbf{u})
\end{equation}
In modern LLMs, this is typically implemented as a \textbf{Gated Linear Unit (GLU)} variant such as SwiGLU~\cite{shazeer2020glu}, which has been shown to be highly effective:
\begin{equation}
    \text{FFN}(\mathbf{u}_t) = \left(\sigma(\mathbf{u}_t W_{\text{Up}}) \odot \mathbf{u}_t W_{\text{Gate}}\right)W_{\text{Down}}
    \label{eq:dense_glu_ffn}
\end{equation}
This FFN is the specific component that the Mixture-of-Experts architecture modifies and enhances.

Crucially, as stated, each of these two sub-layers is wrapped by two other components: 
a \textbf{residual connection} (or skip connection) and a \textbf{layer normalisation} step. 
The residual connection is vital for preventing the vanishing gradient problem. 
Layer normalisation stabilises the activations, ensuring that the training of dozens or even hundreds of stacked layers remains feasible.

\subsubsection{Architectural Advances}

Beyond the core components, the performance of modern LLMs relies on several key innovations, including:
\begin{itemize}
    \setlength\itemsep{0.05em} 
    \item \textbf{Root Mean Square Normalisation (RMSNorm):} A computationally efficient alternative to LayerNorm that stabilises training by normalising activations based on their root-mean-square magnitude~\cite{zhang2019root}.
    \item \textbf{Rotary Position Embeddings (RoPE):} A method for encoding the relative positions of tokens by rotating their vector representations, which has been shown to improve generalisation to longer sequences~\cite{su2021roformer}.
    \item \textbf{Advanced Attention Mechanisms:} Techniques such as Latent Attention are used to handle longer contexts more efficiently by first compressing the input sequence into a smaller set of latent representations~\cite{deepseek-v3}.
\end{itemize}
While these techniques optimise existing components of the Transformer, a more fundamental architectural shift for scaling model capacity involves reimagining the Feed-Forward Network (FFN) itself. 
This leads us directly to the Mixture-of-Experts paradigm, which is a sparsity-inducing modification of the FFN.

\subsection{MoE: From Dense Layers to Sparse Experts}
\label{sec:background_moe_intro}

The architectural innovations described previously optimise existing components of the Transformer. The Mixture-of-Experts (MoE) paradigm introduces a more fundamental change by completely redesigning the Feed-Forward Network (FFN), the primary source of a dense model's parameter count and computational cost~\cite{cai2025moesurvey, shazeer2017outrageously, lepikhin2020gshard}.
\subsubsection{Motivation and Key Benefits}

The core idea of an MoE layer is to replace a single FFN with a collection of many smaller, independent FFNs called \textbf{experts}. 
For each incoming token, a lightweight routing mechanism dynamically selects a small subset of these experts (e.g., 2 or 4 out of 64) to process it. 
This strategy of sparse activation yields two significant benefits:

\textbf{Massive Parameter Count for Specialised Knowledge.} 
The first benefit is a dramatic increase in the model's total number of learnable parameters. The total knowledge capacity of the model is the sum of all experts, enabling different experts to learn specialised functions for different types of data or tasks.

\textbf{Constant Computational Cost for Efficient Inference.} 
The second benefit is that this increased capacity does not come with a proportional rise in computational cost. 
Despite the vast number of total parameters, the cost (in FLOPs) per token remains constant and manageable, as it only depends on the small number of activated experts. 
This breaks the direct link between model size and inference cost, enabling a new frontier of scale. 

This paradigm has been successfully adopted by many state-of-the-art open-source LLMs.
A detailed comparison of their respective sizes and expert configurations is presented in Table~\ref{tab:moe_models_full_comparison}, Appendix~\ref{app:models_and_datasets}.

\subsubsection{The MoE Routing Mechanism}

The core of the MoE layer is a deterministic routing mechanism, which decides which subset of experts to activate during inference for each individual tokens. 
The entire MoE FFN layer's working procedure is demonstrated in Figure~\ref{fig:from-transformer-to-moe}. We can break this process down into three distinct stages:

\begin{figure}
    \centering
    \includegraphics[width=\textwidth]{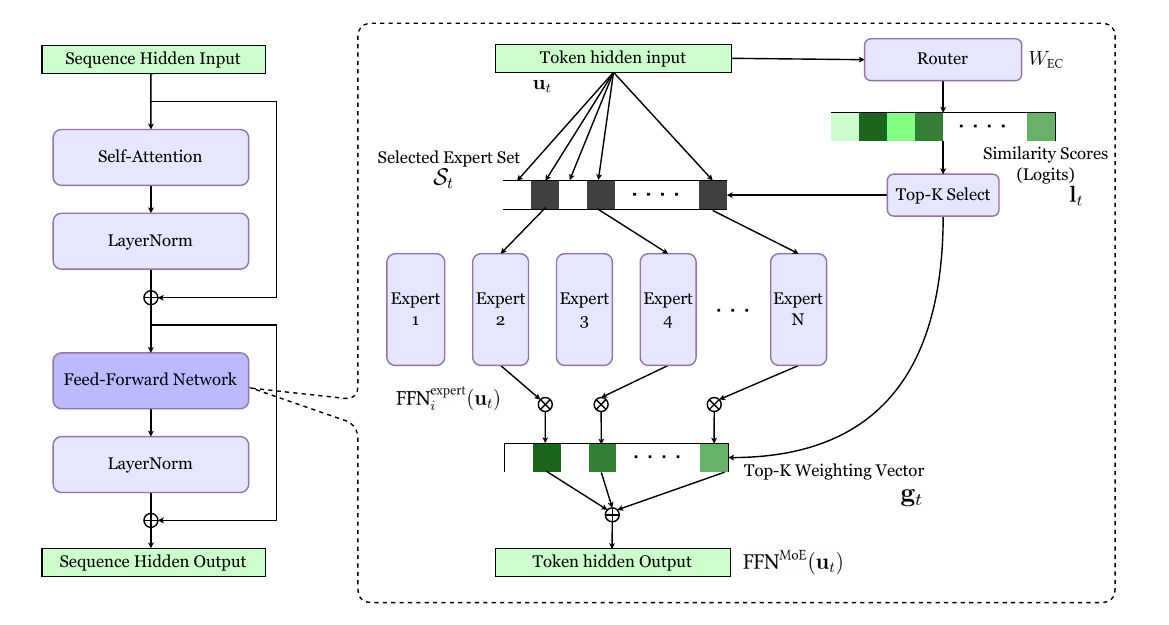}
    \caption{Routing Mechanism in MoE Feed-Forward Network Layer.}
    \label{fig:from-transformer-to-moe}
\end{figure}

\textbf{Stage 1: Expert Similarity Scoring.} 
First, the router computes a similarity score between the input token's hidden state, $\mathbf{u}_t \in \mathbb{R}^D$, and each of the $N$ unique, learnable \textbf{expert centroid vectors}, 
$\mathbf{e}_i \in \mathbb{R}^D$. 
This is achieved using a dot product to measure the alignment between the token's representation and each expert's specialised focus. 
For computational efficiency, these $N$ centroid vectors are collected as the columns of a single weight matrix:
\begin{equation}
W_{\text{EC}} = [\mathbf{e}_1, \dots, \mathbf{e}_N]
\end{equation}
The similarity calculation for all experts is then performed with a single matrix multiplication. 
In neural network terms, this is a simple linear projection that produces a vector of unnormalised scores, or \textbf{logits} ($\mathbf{l}_t \in \mathbb{R}^N$):
\begin{equation}
    \mathbf{l}_t = \mathbf{u}_t W_{\text{EC}}
\end{equation}

\textbf{Stage 2: Probability Transformation.}
Next, these raw logit scores are transformed into a discrete probability distribution over all $N$ experts using the softmax function:
\begin{equation}
    \mathbf{s}_t = \text{softmax}(\mathbf{l}_t)
\end{equation}
Taken together, this two-step process of a linear projection followed by a softmax function is a \textbf{multinomial logistic regression}~\cite{wiki:multinomial_logistic_regression} model.

\textbf{Stage 3: Top-K Expert Selection.}
Finally, to enforce sparse activation, a hard, deterministic Top-K selection mechanism is applied to this probability vector $\mathbf{s}_t$. 
This operation identifies the indices of the $K$ experts with the highest probabilities.
\footnote{Many practical implementations select the Top-K experts directly from the logits before applying a renormalising softmax to the scores of only the selected experts~\cite{deepseek-v3}. 
Since the softmax function is monotonic, this yields the exact same set of chosen experts. 
Our softmax $\rightarrow$ Top-K framing is mathematically equivalent for the final selection and provides a more natural foundation for the probabilistic methods developed in this thesis.} 
\begin{equation}
g'_{t, i} = 
\begin{cases} 
s_{t, i} & \text{if } s_{t, i} \in \textsc{Top-K}(\{s_{t, j}\}_{j=1}^{N}) \\
0 & \text{otherwise}
\end{cases}
\label{eq:moe_expert_selection}
\end{equation}
Let $\mathcal{S}_t$ be the set of the Top-K expert indices selected for token $\mathbf{u}_t$, which contains $K$ indices. The probabilities for these selected experts are then renormalised to sum to one,
\begin{equation}
    \mathbf{g}_t = \frac{\mathbf{g}'_t}{\sum_{i=1}^{N} g'_{t, i}}
\end{equation}
forming the final sparse \textbf{gating weights}, $\mathbf{g}_t$, which are used to compute the weighted sum of expert outputs.
\begin{equation}
\text{FFN}^\text{MoE}(\mathbf{u}_t) = \sum_{i \in \mathcal{S}_t} g_{t, i} \cdot \text{FFN}^\text{expert}_i(\mathbf{u}_t)
\end{equation}

\subsubsection{Auxiliary Losses for Router Training}

The hard, competitive nature of the Top-K selection mechanism can lead to a training pathology known as \textbf{routing collapse}~\cite{shazeer2017outrageously}. This occurs when a positive feedback loop causes the router to consistently send the majority of tokens to a small, favored subset of experts. The remaining experts are starved of data and fail to learn, rendering a large portion of the model's capacity useless. 
To counteract this and ensure all experts are trained effectively, auxiliary loss functions are added to the main training task objective with a scaling hyperparameter $\beta$:
\begin{equation}
    \mathcal{L} = \mathcal{L}_{\text{task}} + \beta \cdot \mathcal{L}_{\text{auxiliary}}
\end{equation}

Numerous auxillary losses for stablising and balancing router training have been proposed over the past few years~\cite{pham2024competesmoe, dai2022stablemoe, wang2024auxiliary}. 
Here we only introuduced two most famous ones:

\paragraph{Load-Balancing Loss}
The most common auxiliary loss is a \textbf{load-balancing loss} designed to incentivise the router to distribute tokens evenly across all $N$ experts. For a batch of $T$ tokens, this loss is typically calculated as the dot product of two quantities for each expert $i$: the fraction of tokens in the batch routed to it ($f_i$), and the average router probability it received over those tokens ($P_i$)~\cite{fedus2022switch}:
\begin{equation}
    \mathcal{L}_{\text{balance}} = N \sum_{i=1}^{N} f_i \cdot P_i
\end{equation}
This loss is minimised when each expert receives an equal share of the routing responsibility.

\paragraph{Router Z-Loss}
Some models also employ a \textbf{router Z-loss} to regularise the magnitude of the pre-softmax logits~\cite{zoph2022st}. This loss penalises large logit values, which helps to prevent the router from becoming overly confident in its selections early in training. This can improve training stability and encourage a smoother distribution of routing scores. The loss is calculated as the mean squared log-sum-exp of the logits over a batch:
\begin{equation}
    \mathcal{L}_{\text{Z}} = \frac{1}{T} \sum_{t=1}^{T} \left( \log \sum_{i=1}^{N} \exp(l_{t,i}) \right)^2
\end{equation}

These auxiliary losses are combined with the primary task loss to guide the model towards a stable and balanced routing policy.

\section{Uncertainty and Calibration in Large Language Models}
\label{sec:background_uncertainty}

Having detailed the architecture of a modern LLM, we now turn to the fundamental challenges of reliability that motivate our work. To understand the need for a Bayesian MoE router, it is crucial to first understand the general problems of overconfidence and miscalibration inherent in standard, deterministic models.

\subsection{The Problem of Overconfidence and Miscalibration}

A fundamental challenge in modern LLMs is the frequent mismatch between the model's predictive probabilities and its true underlying knowledge. 
The softmax outputs of a well-trained network cannot be reliably interpreted as a true measure of the model's confidence. 
This phenomenon is known as \textbf{miscalibration}, and for most modern deep networks, it manifests as consistent \textbf{overconfidence}, a tendency to produce high-probability predictions that are, in fact, incorrect~\cite{guo2017calibration}.

This overconfidence is a primary driver of one of the most significant failure modes in LLMs: \textbf{hallucination}. 
Defined as the generation of plausible-sounding but factually baseless or fictitious content, hallucination makes models fundamentally untrustworthy~\cite{ji2023survey}. 
In high-stakes domains such as medicine or law, the tendency to state falsehoods with unwavering certainty poses a critical safety risk and a major barrier to adoption.

The formal goal is to achieve good \textbf{calibration}. 
A model is considered perfectly calibrated if its predicted confidence aligns with its empirical accuracy. 
For instance, across the set of all predictions to which the model assigns an 80\% confidence, a calibrated model will be correct on 80\% of them. 
Achieving better calibration is therefore a central objective in the pursuit of safe and reliable AI, and it is a primary motivation for the methods developed in this thesis.

\subsection{Evaluating Uncertainty: From Sequences to Controlled Predictions}

Quantifying the uncertainty of an LLM's output is a complex task, especially for open-ended, autoregressive generation. The output space is vast, and uncertainty can accumulate at each step, making it difficult to obtain a reliable and interpretable measure. This remains an active and challenging area of research, with various proposed methods.

The most traditional metric is \textbf{Perplexity (PPL)}, the exponentiated average negative log-likelihood of a sequence, which measures how ``surprised'' a model is by the text:
\begin{equation}
    \text{PPL}(\mathbf{s}) = \exp\left\{-\frac{1}{T}\sum_{t=1}^{T} \log p(s_t|s_{<t})\right\}
\end{equation}
More advanced approaches, like \textbf{Semantic Entropy}, aim to measure uncertainty by clustering the semantic meaning of many possible generated sequences~\cite{kuhn2023semantic, farquhar2024detecting}. 
The entropy is calculated over the probability of these semantic clusters rather than individual tokens. Each semantic cluster $\mathbf{c}$ is defined as $\forall \mathbf{s}, \mathbf{s}' \in \mathbf{c}: E(\mathbf{s}, \mathbf{s}')$, where $E$ is a semantic equivalence relation. 
$\mathcal{C}$ is semantic cluster space. The semantic entropy is then given by:
\begin{equation}
    \mathcal{H}_{\text{sem}}(p(y|\mathbf{x})) = -\sum_{\mathbf{c} \in \mathcal{C}} p(\mathbf{c}|\mathbf{x}) \log p(\mathbf{c}|\mathbf{x})
\end{equation}
Other methods focus on explicitly teaching the model to assess its own confidence, either through direct \textbf{prompting} or by using \textbf{Supervised Fine-Tuning (SFT)} to train the model to state when it does not know the answer~\cite{kapoor2024large}. An example of such prompting strategies is shown in Table~\ref{tab:prompting_examples}.

\begin{table}[H]
    \centering
    \caption{Examples of prompting strategies for outputing model confidence.}
    \label{tab:prompting_examples}
    \begin{tabular}{l l l}
        \toprule
        \textbf{Name} & \textbf{Format} & \textbf{Confidence} \\
        \midrule
        Zero-Shot Classifier& ``Question. Answer. \impb{True/False:} \textcolor{red}{True}'' & $\frac{P(\text{``\textcolor{red}{True}''})}{P(\text{``\textcolor{red}{True}''}) + P(\text{``\textcolor{red}{False}''})}$ \\
        \addlinespace 
        Verbalised & ``Question. Answer. \impb{Confidence: \textcolor{red}{90\%}}'' & \texttt{float(``\textcolor{red}{90\%}'')} \\
        \bottomrule
    \end{tabular}
\end{table}

While these methods are valuable for sequence-level analysis, in order to rigorously and quantitatively evaluate the impact of the architectural changes proposed in this thesis, a more controlled and standardised evaluation setting is required. A common and effective strategy is to simplify the task to the fundamental problem of \textbf{next-token prediction} in a constrained environment.

For this purpose, \textbf{Multiple-Choice Question Answering (MCQA)}
\footnote{A detailed summary of the MCQA datasets used later in this thesis is provided in Table~\ref{tab:mcqa_datasets_summary}, Appendix~\ref{app:models_and_datasets}.}
provides an ideal testbed. In this setting, the model's task is reduced to assigning probabilities over a small, discrete set of predefined answer choices. This allows for a direct and unambiguous comparison between the model's assigned probability for the correct answer (its confidence) and the actual outcome. This provides a clean, reliable signal for measuring the model's calibration, which is the focus of our evaluation.

\subsection{Formal Metrics for Calibration}

Within the controlled setting of Multiple-Choice Question Answering (MCQA), we can use a suite of formal metrics to quantify a model's performance and, more importantly, its calibration.

A primary metric for any probabilistic classifier is the \textbf{Negative Log-Likelihood (NLL)}, also known as the cross-entropy loss. It measures how well the model's predicted probability distribution aligns with the ground-truth outcome. A lower NLL indicates that the model is not only accurate but also assigns high confidence to the correct answers.

\begin{figure}[H]
    \centering
    \includegraphics[width=\textwidth]{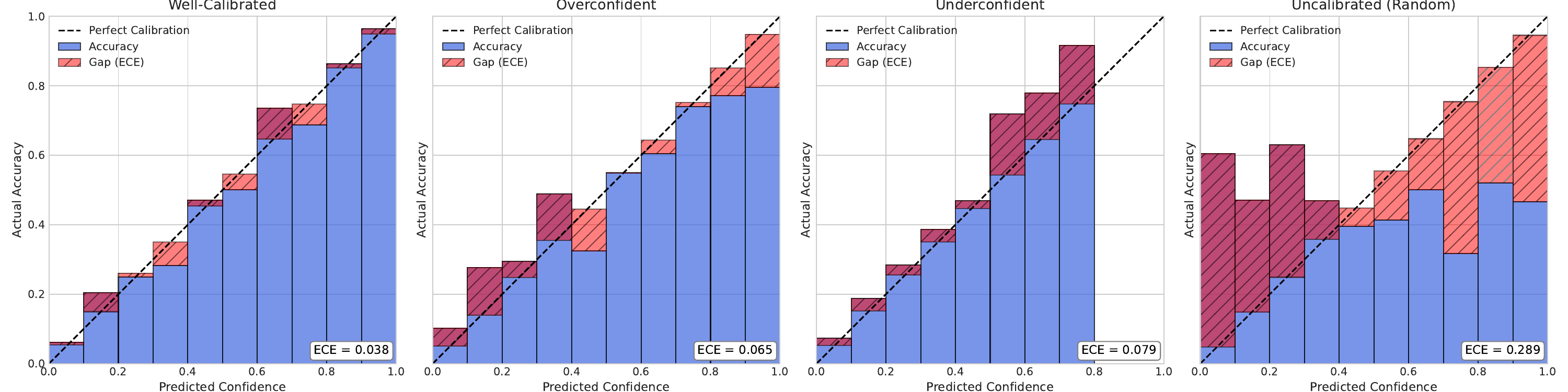}
    \caption{An example of a Reliability Diagram. The blue bars represent the model's accuracy within each confidence bin, while the red bars show the gap to perfect calibration (the diagonal line).}
    \label{fig:reliability_diagram}
\end{figure}

To measure miscalibration directly, the most common metric is the \textbf{Expected Calibration Error (ECE)}~\cite{naeini2015obtaining, guo2017calibration}. ECE measures the difference between a model's average confidence and its actual accuracy. To compute it, predictions are first grouped into $M$ bins based on their confidence scores. For each bin $B_m$, the average confidence, $\text{conf}(B_m)$, is compared to the actual accuracy of the predictions within that bin, $\text{acc}(B_m)$. The ECE is the weighted average of the absolute differences across all bins:
\begin{equation}
    \text{ECE} = \sum_{m=1}^{M} \frac{|B_m|}{n} \left| \text{acc}(B_m) - \text{conf}(B_m) \right|
\end{equation}
where $n$ is the total number of predictions. A lower ECE signifies a better-calibrated model. A complementary metric is the \textbf{Maximum Calibration Error (MCE)}, which measures the worst-case deviation by taking the maximum of the differences:
\begin{equation}
    \text{MCE} = \max_{m=1, \dots, M} \left| \text{acc}(B_m) - \text{conf}(B_m) \right|
\end{equation}
These metrics are often visualised using \textbf{Reliability Diagrams}. As shown in Figure~\ref{fig:reliability_diagram}, this plot shows the actual accuracy for each confidence bin. For a perfectly calibrated model, the bars align perfectly with the diagonal line, where confidence equals accuracy.

In addition to calibration, a key aspect of our evaluation is a model's ability to distinguish in-domain data from out-of-distribution (OoD) data. 
This is framed as a binary classification task where the model's uncertainty score is used as a predictor. 
We evaluate this using two standard metrics: 
the \textbf{Area Under the Receiver Operating Characteristic curve (AUROC)} and 
the \textbf{Area Under the Precision-Recall curve (AUPRC)}~\cite{davis2006relationship}. 
The AUROC measures the trade-off between true positive and false positive rates, 
while the AUPRC is more informative for imbalanced datasets. 
For both metrics, a higher score indicates a more reliable uncertainty signal for OoD detection.

\subsection{Related Work in LLM Calibration}

Improving the calibration of neural networks is an active area of research. Several prominent techniques have been proposed, which can be broadly categorised as post-hoc methods or training-time regularisation.

The most common and effective post-hoc method is \textbf{Temperature Scaling}~\cite{guo2017calibration}. 
This simple technique learns a single scalar temperature parameter, $T$, on a held-out validation set. 
At inference time, the final logits of the model are divided by $T$ before the softmax function is applied. 
This ``softens'' the probability distribution, reducing the model's overconfidence without changing its accuracy. While more complex methods exist, Temperature Scaling remains a very strong baseline.

Another approach is to regularise the model during training to discourage it from producing overconfident predictions. A classic example is \textbf{Label Smoothing}~\cite{szegedy2016rethinking}. 
Instead of training on hard, one-hot labels (e.g., \texttt{[0, 1, 0]}), the model is trained on softened labels (e.g., \texttt{[0.05, 0.9, 0.05]}). This prevents the model from becoming excessively certain by discouraging the logits for the correct class from growing infinitely larger than others.

\subsection*{Towards Making MoE-based LLMs Know What They Don't Know}
In contrast to these approaches, which operate either as a post-processing step on the final output (Temperature Scaling) or as a modification to the training objective (Label Smoothing), the work in this thesis explores a fundamentally different, \textbf{architectural} solution. 
We hypothesise that miscalibration in MoE models can be addressed at a more foundational level, by improving the reliability of the expert selection mechanism itself. 
Rather than correcting the final output, we aim to build a more inherently calibrated model by introducing principled Bayesian uncertainty directly into the MoE router.

\section{Bayesian Machine Learning: \\ A Principled Approach to Uncertainty}
\label{sec:background_bml}

This final section of our background review introduces the mathematical and conceptual tools used to address the challenges of uncertainty and calibration. 
While standard machine learning often seeks a single set of ``best'' model parameters, a point estimate, the Bayesian paradigm takes a different approach. 
Instead of a single answer, it aims to derive a full probability distribution over all possible parameters. 
This distribution serves as a principled representation of the model's uncertainty, providing a foundation for building more reliable and robust systems.

\subsection{The Bayesian Framework}
\label{sec:background_bayes_framework}

\subsubsection{Prior, Likelihood, and Posterior}
Bayesian inference is a framework for updating our beliefs in light of new evidence. It involves three core components:
\begin{itemize}
    \setlength\itemsep{0.05em} 
    \item The \textbf{Prior Distribution}, $p(\theta)$, which represents our initial belief about the model parameters $\theta$ before observing any data. It often serves as a form of regularisation.
    \item The \textbf{Likelihood}, $p(\mathcal{D}|\theta)$, which is the probability of observing our dataset $\mathcal{D}$ given a specific set of parameters $\theta$.
    \item The \textbf{Posterior Distribution}, $p(\theta|\mathcal{D})$, which is our updated belief about the parameters after having observed the data.
\end{itemize}
These components are formally connected by \textbf{Bayes' Theorem}, which provides the mathematical engine for updating our beliefs:
\begin{equation}
    p(\theta|\mathcal{D}) = \frac{p(\mathcal{D}|\theta) p(\theta)}{p(\mathcal{D})}
\end{equation}

\subsubsection{The Challenge of the Marginal Likelihood}
While elegant, this framework presents a major practical challenge. The denominator in Bayes' Theorem, $p(\mathcal{D})$, is the \textbf{marginal likelihood}, also known as the \textbf{model evidence}. It is calculated by integrating over the entire parameter space:
\begin{equation}
    p(\mathcal{D}) = \int p(\mathcal{D}|\theta)p(\theta)d\theta
\end{equation}
For any non-trivial model like a neural network, where $\theta$ can represent millions or billions of parameters, this high-dimensional integral is computationally \textbf{intractable}. 
Since the marginal likelihood cannot be computed, the true posterior distribution is also inaccessible. This intractability is the central challenge in Bayesian deep learning and motivates the need for the approximation methods we will discuss next.

\subsection{Bayesian Neural Networks (BNNs)}

\begin{figure}[H]
    \centering
    \includegraphics[width=\textwidth]{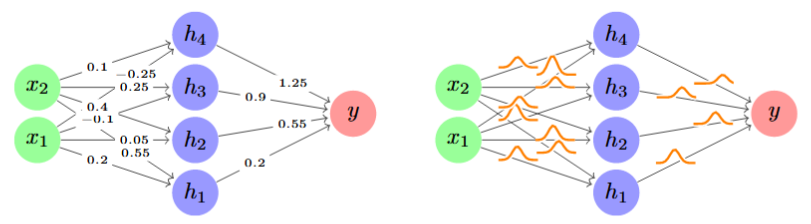}
    \caption{From Point Estimate to Weight Distribution: The Bayesian Neural Network Paradigm. 
    (A) A standard neural network learns a single set of weights, represented as a point estimate in weight space.
    (B) A Bayesian Neural Network learns a full posterior distribution over weights, capturing uncertainty and enabling more robust predictions.}
    \label{fig:bnn_from_point_to_dist}
\end{figure}

The general principles of Bayesian inference can be directly applied to neural networks, where the parameters $\theta$ correspond to the network's weights and biases, $W$. 
Instead of training to find a single, optimal point-estimate for these weights, a Bayesian Neural Network (BNN) aims to infer the full posterior distribution over them, $p(W|\mathcal{D})$, 
as illlustrated in Figure~\ref{fig:bnn_from_point_to_dist}
\footnote{Illustration taken from the Murphy textbook~\cite{murphy2024pml}.}.

\subsubsection{Weight-Space Posterior and Predictive Distribution}
The posterior distribution over the weights, $p(W|\mathcal{D})$, captures the model's \textbf{epistemic uncertainty}, 
that is, the uncertainty that arises from having limited training data. A wide posterior for a given weight indicates that many different values for that weight are plausible given the data, while a narrow posterior indicates high certainty.

To make a prediction for a new input $\mathbf{x}$, a BNN marginalises over this entire distribution of weights. The resulting \textbf{posterior predictive distribution} averages the outputs of an infinite ensemble of networks, each weighted by its posterior probability:
\begin{equation}
    p(y|\mathbf{x}, \mathcal{D}) = \int p(y|\mathbf{x}, W) p(W|\mathcal{D}) dW
\end{equation}
The variance of this predictive distribution provides a principled measure of the model's uncertainty in its output.

\subsubsection{An Overview of Approximation Methods}

As the true posterior $p(W|\mathcal{D})$ is intractable, BNNs must rely on approximation methods. The goal of these methods is to enable the approximation of the posterior predictive distribution, typically via Monte Carlo integration:
\begin{equation}
    p(y|\mathbf{x}, \mathcal{D}) = \int p(y|\mathbf{x}, W) p(W|\mathcal{D}) dW \approx \frac{1}{S} \sum_{s=1}^{S} p(y|\mathbf{x}, W^s)
\end{equation}
where $W^s$ are samples from a distribution that approximates the true posterior. The key difference between methods lies in how they obtain these samples.

\paragraph{Hamiltonian Monte Carlo (HMC)}
MCMC methods like Hamiltonian Monte Carlo (HMC)~\cite{neal2011mcmc} are a class of algorithms that can, given enough computation, generate samples that converge to the true posterior $p(W|\mathcal{D})$. HMC is a gold-standard method that uses principles from Hamiltonian dynamics to explore the parameter space efficiently and produce high-quality samples. However, its significant computational cost makes it impractical for the vast parameter spaces of modern LLMs.

\paragraph{MC Dropout}
A highly scalable alternative is Monte Carlo Dropout~\cite{gal2016dropout}, which reinterprets dropout as approximate Bayesian inference. The key insight is to keep dropout active during inference. Each of the $S$ stochastic forward passes, with its unique random dropout mask, is treated as a sample from an approximate weight posterior. The resulting predictions are then averaged to approximate the predictive distribution, where each $W^s$ represents the base weights with the $s$-th dropout mask applied.

\paragraph{Stochastic Weight Averaging Gaussian (SWAG)}
SWAG~\cite{maddox2019simple} approximates the posterior with a multivariate Gaussian distribution, $\mathcal{N}(\boldsymbol{\mu}_{\text{SWAG}}, \boldsymbol{\Sigma}_{\text{SWAG}})$, by leveraging the trajectory of weights during SGD training. After an initial convergence phase, the first and second moments of the weight iterates are collected to form the mean and a low-rank plus diagonal covariance. Inference is performed by drawing $S$ weight samples, $W^s \sim \mathcal{N}(\boldsymbol{\mu}_{\text{SWAG}}, \boldsymbol{\Sigma}_{\text{SWAG}})$, and averaging their predictions.

\paragraph{Deep Ensembles}
Deep Ensembles~\cite{lakshminarayanan2017simple} provide a powerful, non-explicitly Bayesian approach. The method involves training an ensemble of $M$ identical networks independently from different random initialisations. This collection of trained models, $\{W_1, \dots, W_M\}$, is treated as an empirical sample from the true posterior. The predictive distribution is approximated by averaging the predictions of all $M$ models in the ensemble (i.e., where $S=M$ and $W^s$ is the weight matrix of the $s$-th model).

These scalable methods provide computationally feasible ways to approximate the weight posterior. An alternative family of approximation methods, which reframes the problem as one of optimisation, is \textbf{Variational Inference}, which we will detail next.

\subsection{Variational Inference (VI)}

The final piece of theoretical background we require is \textbf{Variational Inference (VI)}, 
a powerful and widely used alternative to MCMC for approximating intractable posterior distributions~\cite{jordan1999introduction}. 
Instead of drawing samples, VI reframes the inference problem as one of optimisation, making it a natural fit for the gradient-based methods used in deep learning.

\subsubsection{Core Idea: Posterior Approximation via Optimisation}

The goal of VI is to approximate a complex and intractable true posterior, $p(\boldsymbol{z}|\boldsymbol{x})$, with a simpler, tractable distribution, $q_\phi(\boldsymbol{z})$, from a chosen family of distributions. 
The parameters $\phi$ of this ``variational distribution'' are optimised to make it as close as possible to the true posterior. 
This closeness is measured by the \textbf{Kullback-Leibler (KL) Divergence}.

Directly minimising the KL divergence is not possible, as its definition still contains the intractable posterior $p(\boldsymbol{z}|\boldsymbol{x})$. However, we can derive an alternative objective. The log marginal likelihood of the data, $\log p(\boldsymbol{x})$, can be decomposed as follows:
\begin{align}
    \log p(\boldsymbol{x}) &= \log \int p(\boldsymbol{x}|\boldsymbol{z})p(\boldsymbol{z})d\boldsymbol{z} \notag \\
    &= \log \int q_\phi(\boldsymbol{z}) \frac{p(\boldsymbol{x}|\boldsymbol{z})p(\boldsymbol{z})}{q_\phi(\boldsymbol{z})} d\boldsymbol{z} \notag \\
    &\ge \int q_\phi(\boldsymbol{z}) \log \frac{p(\boldsymbol{x}|\boldsymbol{z})p(\boldsymbol{z})}{q_\phi(\boldsymbol{z})} d\boldsymbol{z} \quad \impb{\text{(Jenson's Inequality)}} \notag \\
    &= \mathbb{E}_{q_\phi(\boldsymbol{z})}\left[\log p(\boldsymbol{x}|\boldsymbol{z})\right] - \KL\left[q_\phi(\boldsymbol{z}) || p(\boldsymbol{z})\right] := \mathcal{L}(\phi).
    \label{eq:elbo_derivation}
\end{align}
This gives us the \textbf{Evidence Lower Bound (ELBO)}, $\mathcal{L}(\phi)$. 
As its name and the math suggest, ELBO is a lower bound on the log marginal likelihood.
Besides, there's also a connection between optimising ELBO and the original intention of optimising KL divergence between $q_\phi(\boldsymbol{z})$ and $p(\boldsymbol{z}|\boldsymbol{x})$:
\begin{align}
\log p(\boldsymbol{x}) - \KL(q_\phi(\boldsymbol{z})||p(\boldsymbol{z}|\boldsymbol{x})) 
&= \log p(\boldsymbol{x}) - \mathbb{E}_{q_\phi(\boldsymbol{z})}\left[ \log \frac{q_\phi(\boldsymbol{z})}{p(\boldsymbol{z}|\boldsymbol{x})} \right] \notag \\
&= \log p(\boldsymbol{x}) + \mathbb{E}_{q_\phi(\boldsymbol{z})}\left[ \log \frac{p(\boldsymbol{x}|\boldsymbol{z})p(\boldsymbol{z})}{q_\phi(\boldsymbol{z})p(\boldsymbol{x})} \right] \quad \impb{\text{(Bayes' Theorem)}} \notag \\
&= \mathbb{E}_{q_\phi(\boldsymbol{z})}[\log p(\boldsymbol{x}|\boldsymbol{z})] - \KL(q_\phi(\boldsymbol{z})||p(\boldsymbol{z})) = \mathcal{L}(\phi).
\label{eq:elbo_kl_connection}
\end{align}

Crucially, because $\log p(\boldsymbol{x})$ is a constant with respect to $\phi$, maximising the ELBO is equivalent to minimising the KL divergence
\footnote{Equations~\ref{eq:elbo_derivation} and~\ref{eq:elbo_kl_connection} are adapted from lecture note~\cite{li2022vae}.}.

The ELBO is typically written in a more intuitive form:
\begin{equation}
    \mathcal{L}(\phi) = \underbrace{\mathbb{E}_{q_\phi(\boldsymbol{z})}[\log p(\boldsymbol{x}|\boldsymbol{z})]}_{\text{Reconstruction Term}} - \underbrace{\KL(q_\phi(\boldsymbol{z}) || p(\boldsymbol{z}))}_{\text{Regularisation Term}}
\end{equation}
The \textbf{reconstruction term} encourages the model to explain the observed data, while the \textbf{regularisation term} keeps the approximate posterior close to the prior $p(\boldsymbol{z})$.

\subsubsection{Structuring \(q_\phi\): Multivariate Gaussian and the Mean-Field Assumption}

A primary design choice in VI is the family of distributions used for the approximate posterior, $q_\phi(\boldsymbol{z})$. 
A common and flexible choice is the \textbf{multivariate Gaussian distribution}, $\mathcal{N}(\boldsymbol{z} | \boldsymbol{\mu}_\phi, \boldsymbol{\Sigma}_\phi)$, as it can capture both the central tendency and the variance of the latent variables. 
When the prior is chosen to be a standard multivariate normal, $p(\boldsymbol{z}) = \mathcal{N}(\boldsymbol{z} | \mathbf{0}, I)$, the KL divergence term in the ELBO has a convenient analytical solution:
\begin{equation}
    \KL\left(\mathcal{N}(\boldsymbol{\mu}_\phi, \boldsymbol{\Sigma}_\phi) || \mathcal{N}(\mathbf{0}, I)\right) = \frac{1}{2} \left( \text{tr}(\boldsymbol{\Sigma}_\phi) + \boldsymbol{\mu}_\phi^\top\boldsymbol{\mu}_\phi - k - \log|\boldsymbol{\Sigma}_\phi| \right)
\end{equation}
where $k$ is the dimensionality of the latent space $\boldsymbol{z}$.

However, for high-dimensional latent spaces common in deep learning, parameterising and computing with a full-rank covariance matrix $\boldsymbol{\Sigma}_\phi$ is often computationally prohibitive. A standard and effective simplification is the \textbf{mean-field assumption}~\cite{bishop2006pattern}. This assumes that the posterior distribution factorises across its dimensions, i.e., $q_\phi(\boldsymbol{z}) = \prod_i q_{\phi_i}(z_i)$. For a Gaussian, this is equivalent to constraining the covariance matrix to be diagonal, $\boldsymbol{\Sigma}_\phi = \text{diag}(\boldsymbol{\sigma}_\phi^2)$.

This simplification significantly reduces the computational complexity. The KL divergence for the mean-field case reduces to a simple sum over the dimensions, avoiding all expensive matrix operations like determinants or inversions:
\begin{equation}
    \KL\left(\mathcal{N}(\boldsymbol{\mu}_\phi, \text{diag}(\boldsymbol{\sigma}_\phi^2)) || \mathcal{N}(\mathbf{0}, I)\right) = \frac{1}{2} \sum_{i=1}^{k} \left( \mu_{{\phi}_i}^2 + \sigma_{{\phi}_i}^2 - \log(\sigma_{{\phi}_i}^2) - 1 \right)
\end{equation}
This tractable and efficient formulation is a cornerstone of most practical applications of VI in deep learning.
However, if the dimensionality of the latent space is tractable, it is possible to model the full-rank covariance matrix by parameterising it via its \textbf{Cholesky decomposition}~\cite{deisenroth2020mml}. 
This more expressive approach, which we detail later in our Methodology section~\ref{sec:methodology_fcvr}, allows the model to capture correlations between the latent variables.

\subsubsection{Amortised VI: VAE Case Study}

In the traditional formulation of VI, a separate set of variational parameters $\phi$ must be optimised for each data point. 
For large datasets, this is computationally infeasible. \textbf{Amortised VI} solves this by learning a single global function, an \textbf{inference network}, that maps any input data point $\mathbf{x}$ to the parameters of its approximate posterior, $q_\phi(\boldsymbol{z}|\mathbf{x})$. 
The cost of training this network is thus ``amortised'' over the entire dataset.

The quintessential example of this approach is the \textbf{Variational Autoencoder (VAE)}~\cite{kingma2013auto}. 
A VAE is a generative model composed of two neural networks: 
an \textbf{encoder} ($q_\phi(\boldsymbol{z} | \mathbf{x})$) that learns to map inputs to a latent distribution, 
and a \textbf{decoder} ($p_\theta(\mathbf{x} | \boldsymbol{z})$) that learns to reconstruct the inputs from samples of that distribution. Typically, the latent distribution is assumed to be a mean-field Gaussian, so the encoder network has two heads to predict the mean $\boldsymbol{\mu}_\phi(\mathbf{x})$ and the log-variance $\log\boldsymbol{\sigma}^2_\phi(\mathbf{x})$.

\begin{figure}[H]
    \centering
    \begin{tikzpicture}[
        >=Latex,
        node distance=1cm and 1cm,
        state/.style={circle, draw, minimum size=1cm, inner sep=0pt}, 
        obs/.style={state, fill=gray!60}, 
        param/.style={text height=1.5ex, text depth=.25ex}, 
        plate/.style={draw, rounded corners, thick, inner sep=0.5cm}, 
        conn/.style={->, thick}, 
        inf_conn/.style={->, thick, dashed} 
    ]

    \node[state] (z) {$\boldsymbol{z}$};
    \node[obs, below=of z] (x) {$\mathbf{x}$};
    \node[param, left=of z] (phi) {$\phi$};
    \node[param, right=of x] (theta) {$\theta$};

    \draw[conn] (z) -- (x);
    \draw[conn] (theta) -- (x); 
    
    \draw[inf_conn] (x) to[bend left] (z); 
    \draw[inf_conn] (phi) -- (z); 

    \node[plate, fit=(x)(z), label={[anchor=south east]south east:$\times N$}] (plate) {};

    \end{tikzpicture}
    \caption{Probabilistic Graphical Model of the Variational Autoencoder (VAE). The solid lines represent the generative model $p_\theta(\mathbf{x}|\mathbf{z})$, while the dashed lines represent the VI model (encoder) $q_\phi(\mathbf{z}|\mathbf{x})$.}
    \label{fig:pgm_vae}
\end{figure}
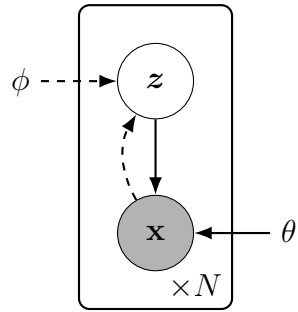

The VAE's structure is represented by the probabilistic graphical model in 
Figure~\ref{fig:pgm_vae} 
\footnote{PGM adapted from \cite{kingma2013auto}. Note that in our depiction, latent prior $p(\boldsymbol{z})$ is not parameterised by $\theta$.}.
This PGM clarifies how the two networks are trained jointly by maximising the ELBO. 
The reconstruction term, $\mathbb{E}_{q_\phi(\boldsymbol{z}|\mathbf{x})}[\log p_\theta(\mathbf{x}|\boldsymbol{z})]$, corresponds directly to the generative path of the model (solid arrows), 
forcing the decoder (parametrised by $\theta$) to accurately reconstruct the input $\mathbf{x}$ from the latent code $\boldsymbol{z}$. 
The regularisation term, $\KL(q_\phi(\boldsymbol{z}|\mathbf{x}) || p(\boldsymbol{z}))$, corresponds to the inference path (dashed arrows), 
forcing the encoder's output (parametrised by $\phi$) to stay close to a simple prior, $p(\boldsymbol{z})$.

\begin{figure}[H]
    \centering
    \includegraphics[width=\textwidth]{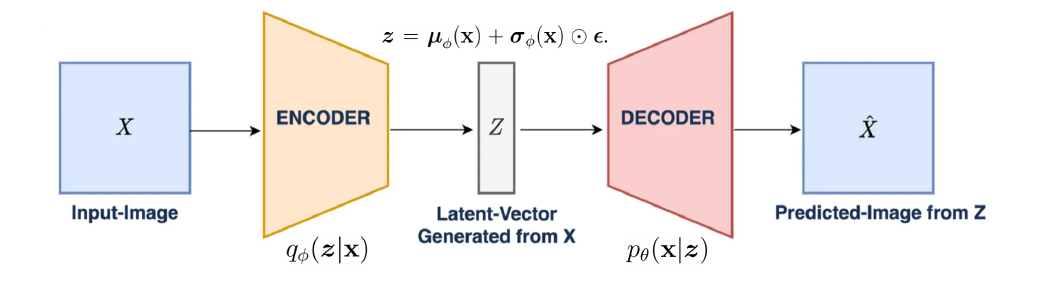}
    \caption{Schematic of the Variational Autoencoder (VAE) architecture.}
    \label{fig:vae_schematic}
\end{figure}

To optimise the ELBO, we must backpropagate gradients through the sampling step $\boldsymbol{z} \sim q_\phi(\boldsymbol{z}|\mathbf{x})$, which is non-differentiable. The VAE enables this with the \textbf{reparameterisation trick}. For a Gaussian latent variable, a sample is drawn by first sampling a standard noise variable $\boldsymbol{\epsilon} \sim \mathcal{N}(\textbf{0}, I)$ and then computing the sample as $\boldsymbol{z} = \boldsymbol{\mu}_\phi(\mathbf{x}) + \boldsymbol{\sigma}_\phi(\mathbf{x}) \odot \boldsymbol{\epsilon}$. This separates the stochasticity from the network parameters, creating a differentiable path for gradients.
The entire VAE schematic is illustrated in Figure~\ref{fig:vae_schematic}
\footnote{VAE Schematic adapted from~\cite{biswal2023dive}.}
.

A common modification to the VAE objective is the introduction of a hyperparameter $\beta$ to scale the KL divergence term, a model known as a $\beta$-VAE~\cite{higgins2017beta}. 
\begin{equation}
    \mathcal{L}_{\beta\text{-VAE}} = \mathbb{E}_{q_\phi(\boldsymbol{z}|\mathbf{x})}[\log p_\theta(\mathbf{x}|\boldsymbol{z})] - \beta \cdot \KL(q_\phi(\boldsymbol{z}|\mathbf{x}) || p(\boldsymbol{z}))
\end{equation}
This can be a crucial tool for preventing \textbf{posterior collapse}, a failure mode where the KL term is minimised too aggressively, causing the latent variables to become uninformative.

This amortised encoder-decoder architecture provides a direct conceptual blueprint for the Variational Routers developed in Section~\ref{sec:methodology_logit_space}.

\chapter{Motivation}
\label{chap:motivation}


This chapter outlines two motivational experiments designed to understand the limitations of deterministic routing strategies in current MoE-based language models. 
The results reveal a fundamental \textbf{brittleness} in the standard routing mechanism under purturbation, 
while also demonstrating the clear \textbf{potential} of introducing stochasticity.
Besides, since current LLMs are stacked with multiple MoE layers,
the experiments are conducted across the network's depth to identify which layers are most sensitive to these issues.
Together, these findings motivate the central goal of this thesis: 
to develop a principled Bayesian routing approach for better uncertainty quantification, aiming to achieve robust expert selection and calibrated output confidence.

\section{Motivation 1: Brittleness of Deterministic Routing}
\label{sec:motivation1}

\label{sec:motivation1}

Our first experiment investigates a fundamental hypothesis: if a router has learned a robust mapping from input representations to expert selections, its decisions should be stable under minimal, non-semantic perturbations. A significant change in expert selection in response to meaningless noise would reveal that the routing mechanism is brittle and inherently unreliable. This section details the experiment designed to quantify this brittleness across the depth of the network.

\subsection{Methodology}
\begin{figure}[H]
    \centering
    \includegraphics[width=0.9\textwidth]{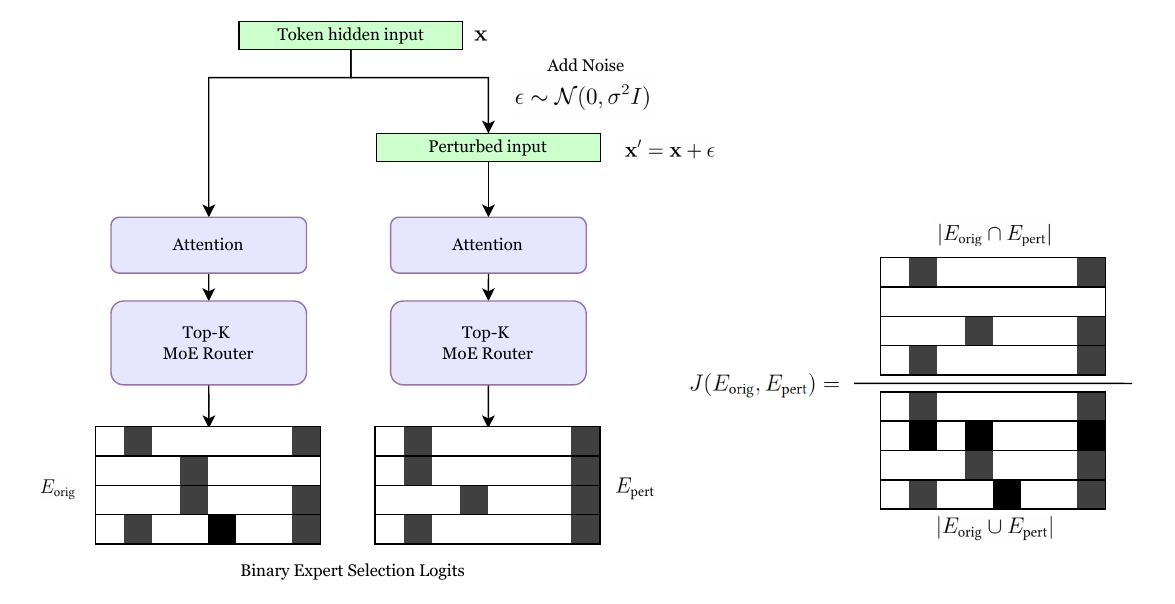}
    \caption{Experimental setup for quantifying the brittleness of deterministic routing at one MoE layer.}
    \label{fig:motivation1_methodology}
\end{figure}

The experiment is conducted on our fine-tuned MAP baseline model using a randomly sampled subset of data from our In-Domain (ID) test set. 
The experimental methodology is illustrated in Figure~\ref{fig:motivation1_methodology}.

To test stability, we introduce a minimal perturbation to the input of each MoE transformer layer. For each token embedding $\mathbf{x}$, a perturbed version $\mathbf{x'}$ is generated by adding Gaussian noise:
\begin{equation}
\mathbf{x'} = \mathbf{x} + \epsilon, \quad \text{where } \epsilon \sim \mathcal{N}(0, \sigma^2 I)
\end{equation}

To ensure the noise is meaningful yet non-semantic, 
the choice of standard deviation $\sigma$ is in proportion to the average L2 norm of the token embeddings, $\bar{L}$. We test multiple noise levels defined by a scaling factor $\gamma$:
\begin{equation}
    \sigma = \gamma \cdot \bar{L}, \quad \text{where } \gamma \in \{0.001, 0.002, 0.005, 0.007, 0.01, 0.02, 0.05\}
\end{equation}

For each token and for each noise level $\gamma$, we record the set of $K$ experts selected for the original input ($E_{\text{orig}}$) and the perturbed input ($E_{\text{pert}}$) at every MoE layer. 
To quantify the change in expert selection, we compute the \textbf{Jaccard Similarity} between these two sets:
\begin{equation}
    J(E_{\text{orig}}, E_{\text{pert}}) = \frac{|E_{\text{orig}} \cap E_{\text{pert}}|}{|E_{\text{orig}} \cup E_{\text{pert}}|}
\end{equation}
A score of 1.0 indicates perfect stability, while a score of 0.0 indicates a complete change in the selected experts.

\subsection{Results and Observations}
\begin{figure}[H]
    \centering
    \includegraphics[width=0.9\textwidth]{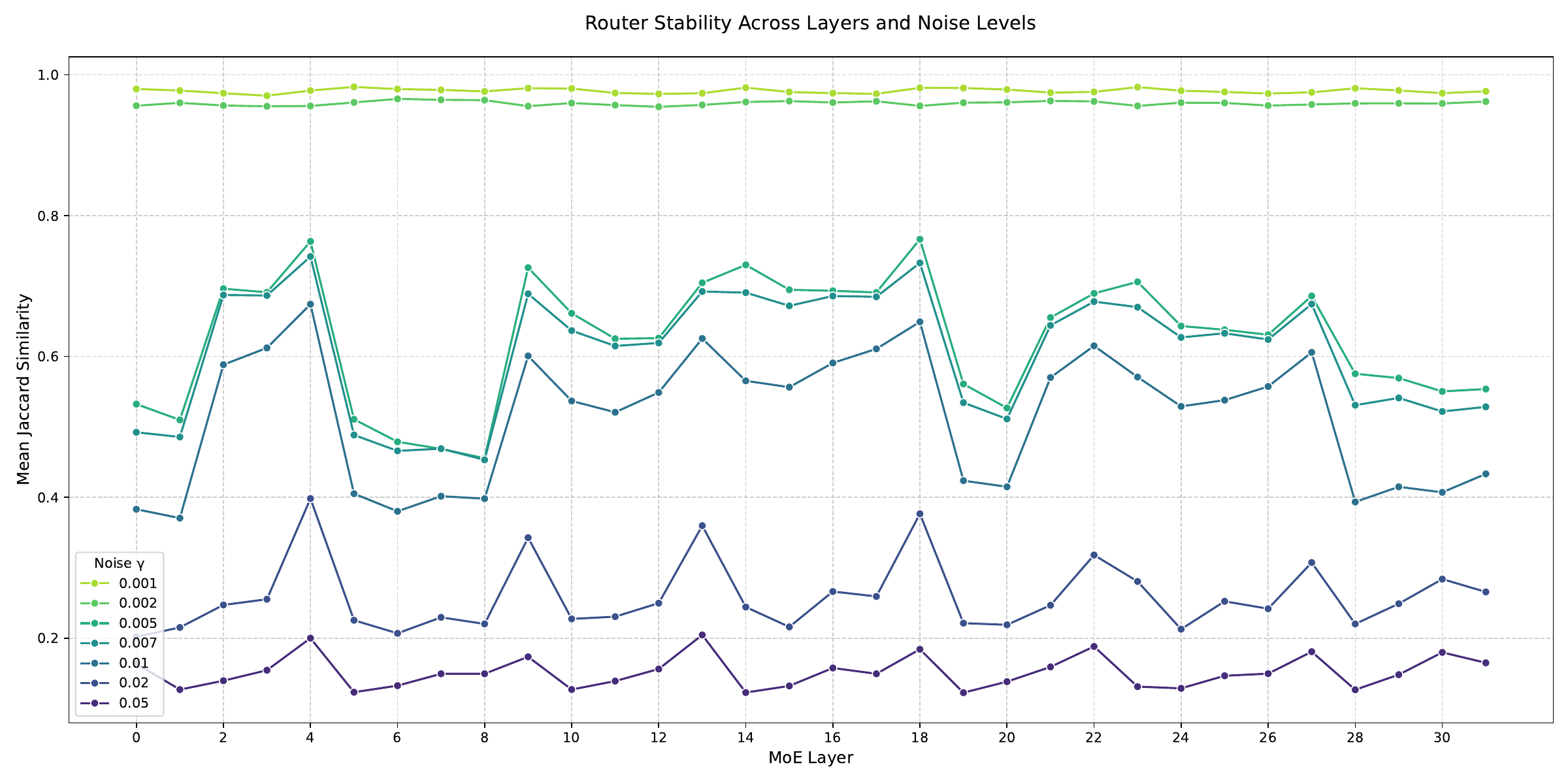}
    \caption{Mean Jaccard similarity across MoE layers for varying levels of input perturbation ($\gamma$). This plot reveals the sensitivity of each layer's router to noise.}
    \label{fig:jaccard_sensitivity}
\end{figure}

Figure \ref{fig:jaccard_sensitivity} shows the mean Jaccard similarity across all MoE layers for various noise levels. 
This sensitivity analysis reveals two key findings. 
\begin{enumerate}
    \setlength\itemsep{0.05em} 
    \item \textbf{General Instability}: Even a relatively very small amount of noise (e.g., $\gamma \ge 0.005$) is sufficient to cause a significant drop in stability, confirming the router's brittleness. 
    \item \textbf{Comparision Across Layers}: These results allow us to select an appropriate noise level for a more granular analysis: a noise level like $\gamma=0.01$ is sensitive enough to reveal instability without being so large that it saturates the effect across all layers.
\end{enumerate}

\begin{figure}[H]
    \centering
    \includegraphics[width=0.9\textwidth]{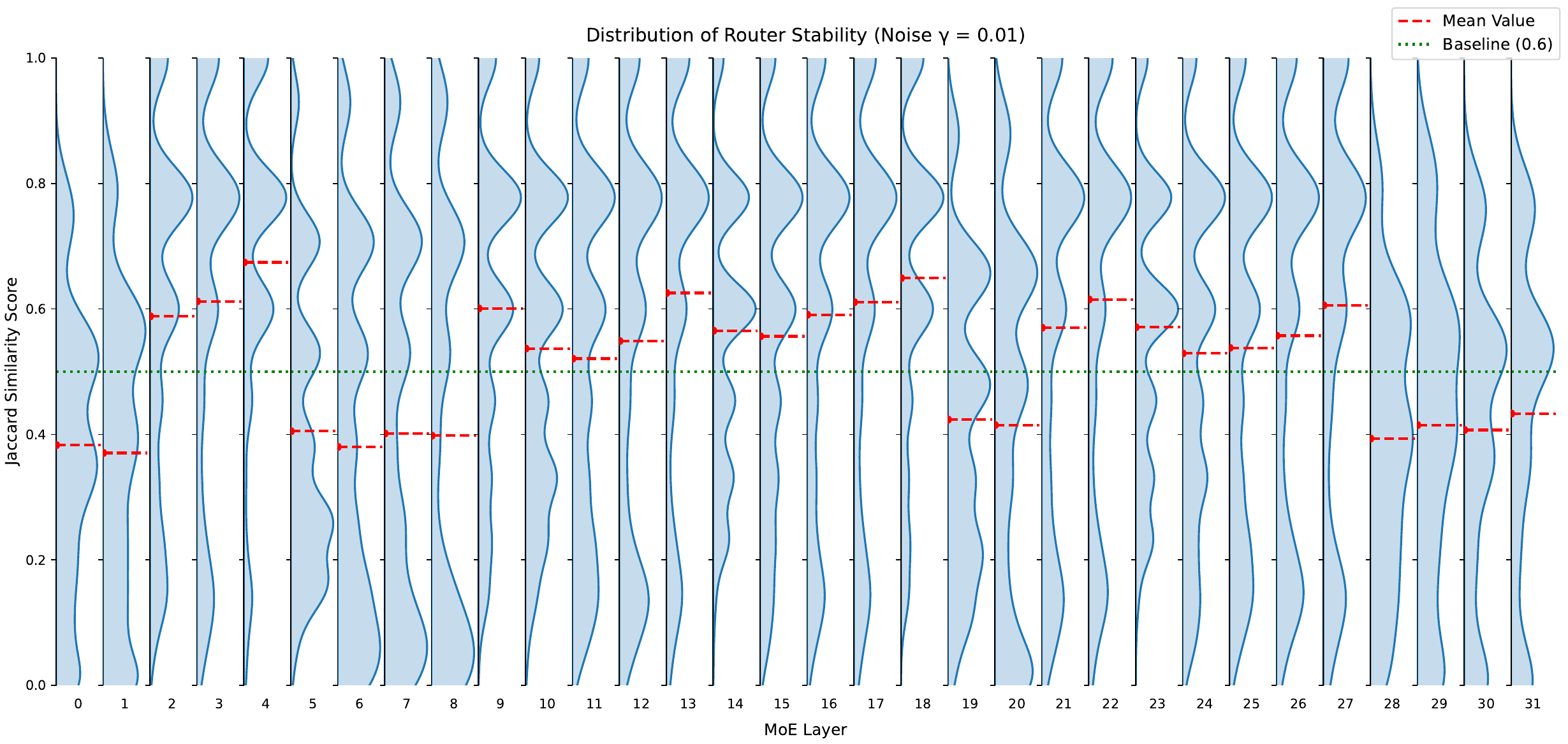}
    \caption{Distribution of token-level Jaccard similarity scores for each MoE layer at a fixed noise level ($\gamma=0.01$). This highlights that router instability is concentrated in specific layer groups.}
    \label{fig:jaccard_distribution}
\end{figure}

Using a fixed noise level of $\gamma=0.01$, 
we then analyze the full distribution of Jaccard scores at each layer, 
shown in Figure \ref{fig:jaccard_distribution}. 
This detailed view provides our main observation: 
\textit{The degree of instability is not uniform across the hierarchical network architecture.}
Instead, the brittleness appears to be concentrated in specific groups of layers. 
In our model, we observe pronounced instability at the \textbf{very beginning (Layers 0-1)}, 
in the \textbf{early-middle (Layers 5-8)}, the \textbf{late-middle (Layers 19-20)}, and most dramatically, at the \textbf{final layers (Layers 28-31)}. 
The distributions in these regions are skewed significantly towards lower Jaccard scores, 
indicating frequent changes in expert selection.

\subsection{Conclusion}
This experiment yields two critical conclusions that motivate our work. 
\begin{enumerate}
    \setlength\itemsep{0.05em} 
    \item Quantitatively confirming that the \textbf{standard deterministic routing mechanism is brittle}, as its decisions are sensitive to semantically meaningless small noise.
    \item Revealing that \textbf{instability is highly dependent on the layer's depth within the network}, which suggests that a Bayesian treatment can target specific susceptible layers rather than entire network
    \footnote{This observation is specific to the \texttt{ibm-granite-3B-MoE} model, which serves as the base model for all subsequent experiments. \
    For a more generalisable approach to layer selection, we also employ a last-$N$ layer selection strategy, as described in Section~\ref{sec:exp_layer_comparison}.}
    .
\end{enumerate}

\section{Motivation 2: Potentials of Stochastic Routing}
\label{sec:motivation2}

Having established the brittleness of the deterministic router, we now investigate whether introducing simple, ad-hoc stochasticity can lead to improvements in model behavior. 
If random noise in the selection process proves beneficial, it would provide a strong motivation for developing a principled Bayesian framework that can learn this stochasticity in a data-driven manner.

\subsection{Methodology}
\begin{figure}[H]
    \centering
    \includegraphics[width=\textwidth]{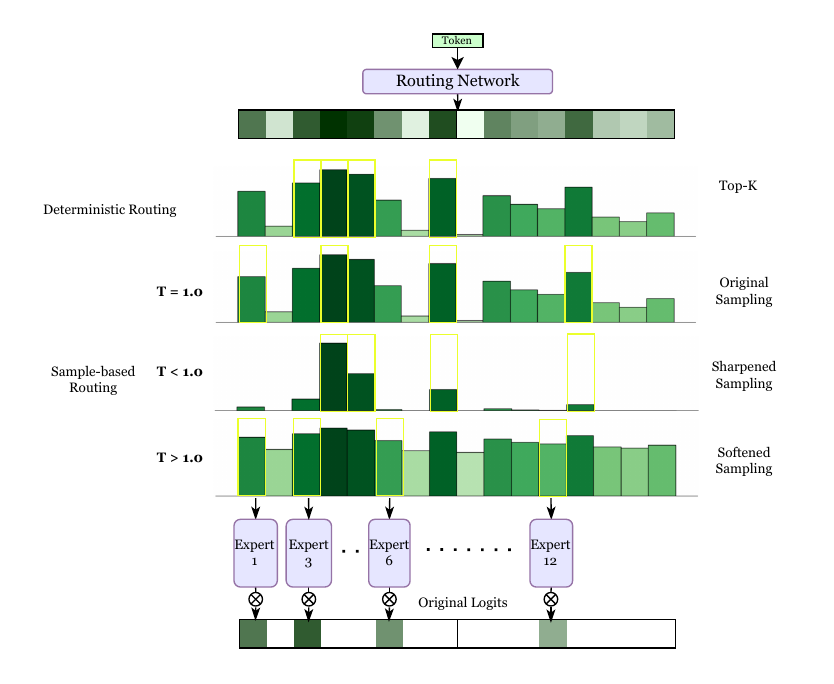}
    \caption{Experimental setup for introducing stochastic routing at a single MoE layer. The temperature parameter $T$ controls the level of randomness in expert selection.}
    \label{fig:motivation2_methodology}
\end{figure}

This experiment modifies the expert selection mechanism within a single MoE layer at a time, while all other layers remain deterministic. 
The standard router computes logits and selects the experts with the Top-K highest values. 
We replace this deterministic selection with a stochastic sampling process (as illustrated in Figure~\ref{fig:motivation2_methodology}):
\begin{enumerate}
    \setlength\itemsep{0.05em} 
    \item \textbf{Temperature Scaling:} Raw logits from router are first scaled by a temperature parameter $T$. A temperature $T>1$ softens the distribution, increasing randomness, while $T<1$ sharpens it.
    \item \textbf{Probabilistic Sampling:} A probability distribution $\mathbf{p}$ is formed by applying the softma]x function to the scaled logits:
    \begin{equation}
        \mathbf{p} = \text{softmax}\left(\frac{\text{logits}}{T}\right)
    \end{equation}
    Instead of selecting the Top-K experts, we then sample $K$ experts without replacement from this distribution $\mathbf{p}$.
\end{enumerate}

This procedure is applied to each MoE layer individually across different runs. 
We evaluate the impact on the model's overall performance on our In-Domain (ID) test set using two key metrics: 
Accuracy (ACC) to measure task performance and Expected Calibration Error (ECE) to measure model calibration.

\subsection{Results and Observations}

\begin{figure}[H]
    \centering
    \includegraphics[width=\textwidth]{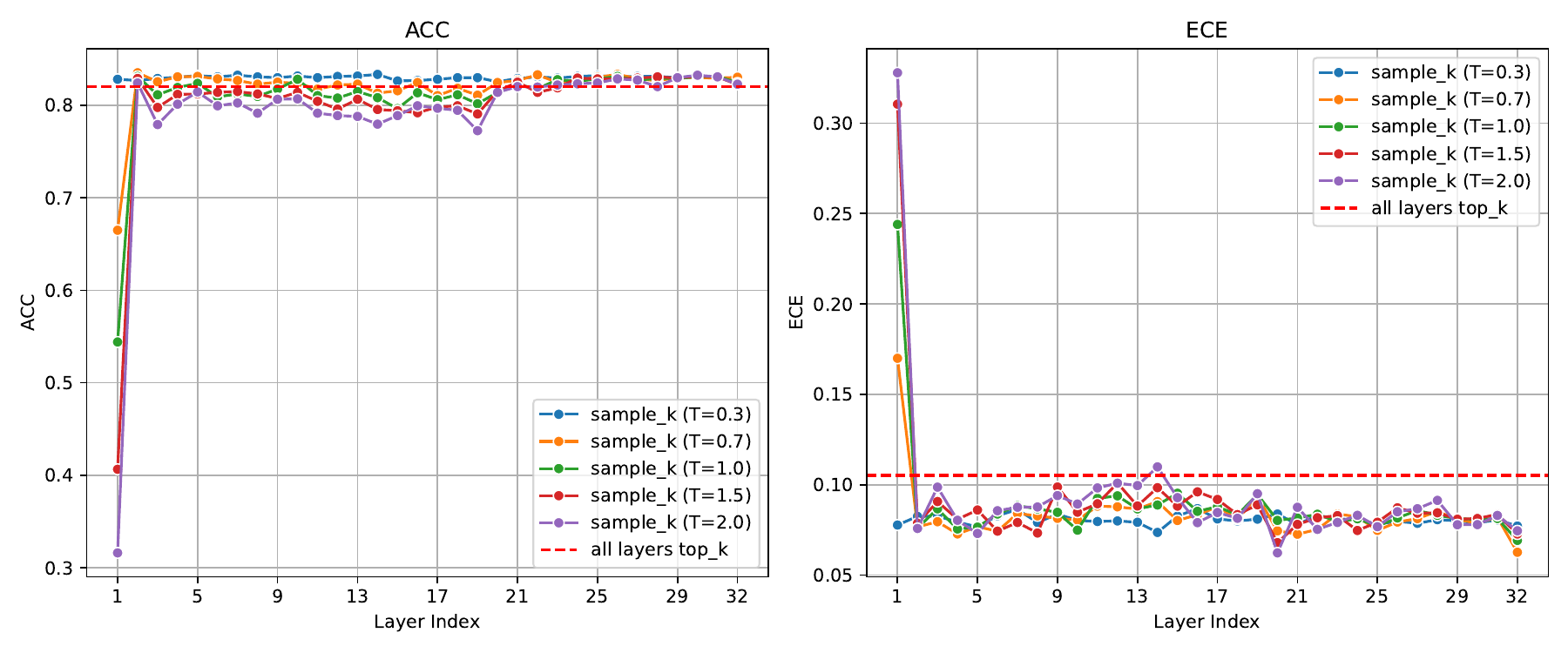}
    \includegraphics[width=\textwidth]{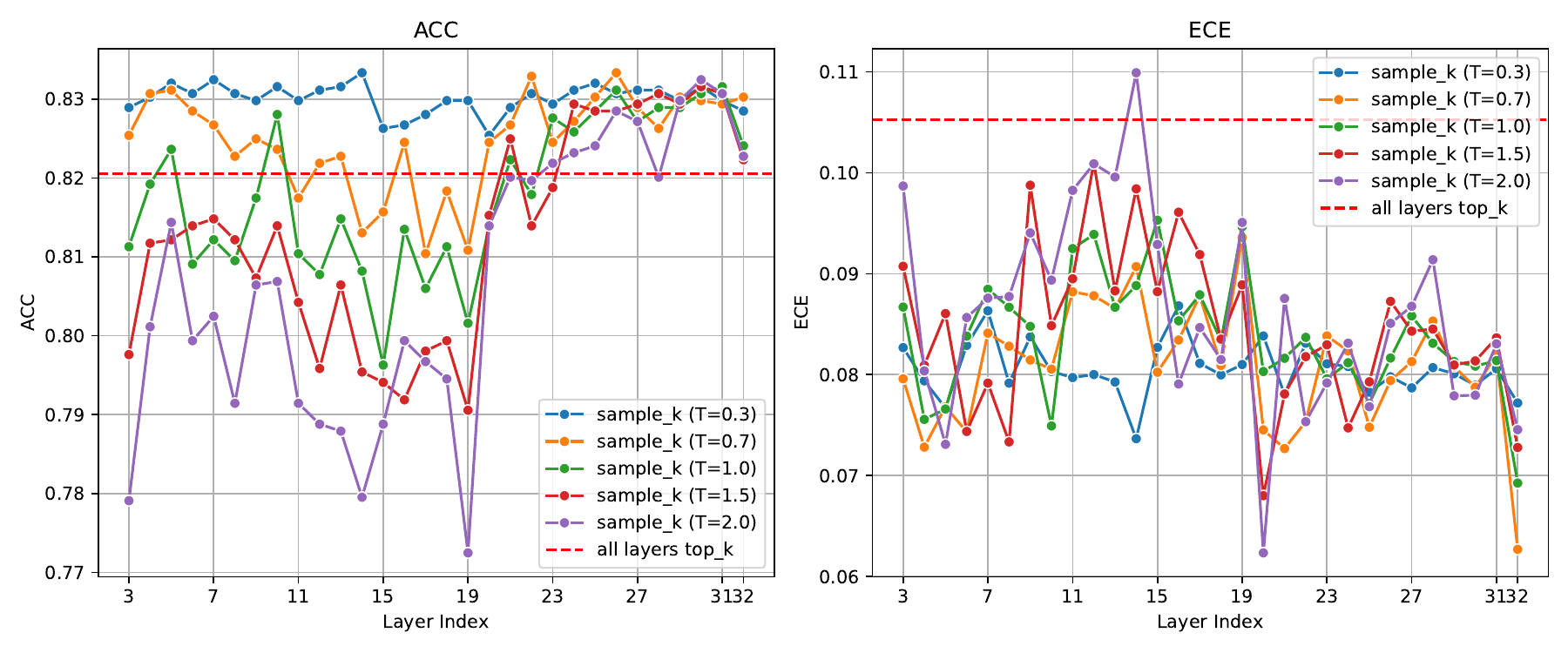}
    \caption{
        Model Accuracy (left) and ECE (right) when applying temperature-based stochastic routing at a single MoE layer at a time. 
        The top plot shows results for all layers, while the bottom plot excludes the first layer for more granular comparison in later layers.
        The dashed line represents the fully deterministic baseline.
    }
    \label{fig:stochastic_routing_results}
\end{figure}

The results of applying this stochastic routing strategy with various temperatures are shown in Figure~\ref{fig:stochastic_routing_results}. The plots display the model's Accuracy and ECE when stochasticity is introduced at each specific layer.

We draw two primary observations from these results:
\begin{enumerate}
    \setlength\itemsep{0.05em} 
    \item \textbf{Early Layers are Highly Sensitive:} 
    Introducing stochastic routing in the first two layers causes a significant degradation in model accuracy. These layers are likely responsible for learning fundamental, low-level representations, and their routing decisions are not robust to this type of random perturbation.
    \item \textbf{Stochasticity Improves Calibration in Later Layers:} 
    For the majority of the middle and later layers, a remarkable trend emerges. 
    Introducing stochasticity (especially with $T = 0.3$) leads to a consistent reduction in ECE compared to the deterministic baseline, \
    while the accuracy remains largely unchanged. 
    This suggests that replacing the overconfident `Top-K' selection with a more stochastic sampling process acts as a form of regularisation, 
    forcing the model to be less certain and, as a result, better calibrated.
\end{enumerate}

\subsection{Conclusion}

This experiment provides two insights that pave the way for this thesis.

\begin{enumerate}
    \setlength\itemsep{0.05em} 
    \item \textbf{Stochasticity can be beneficial.} 
          The fact that a simple, unprincipled injection of randomness can improve model calibration without sacrificing performance 
          strongly suggests that the deterministic router is suboptimal and motivates the need for a more sophisticated, principled Bayesian treatment,
          which has the potential of making better informed decision.
    \item \textbf{Early layers should not be selected for stochasticity.} The detrimental effect of stochasticity on early layers suggests that first layer would not be the apppropriate place to be probablistic.
          Instead, the focus should be on the middle and later layers, where stochasticity can reduce overconfidence without significantly impacting accuracy.
\end{enumerate}

\section{Chapter Summary}
\label{sec:motivation_summary}
These two motivational experiments paint a clear picture. 
The first demonstrates that the standard deterministic router is brittle, 
exhibiting significant instability in its expert selections in response to minimal, non-semantic input noise. 
This reveals a fundamental weakness in the current MoE paradigm.

Conversely, the second experiment shows that introducing simple, heuristic stochasticity in expert selection can be beneficial. 
Replacing the deterministic selection with temperature-based sampling can improve model reliability by reducing overconfidence (lower ECE) at a minimal cost to accuracy.

These findings create a compelling motivation for the work in this thesis. 
If deterministic routing is brittle, and simple, undirected randomness is beneficial, then a principled, data-driven approach to uncertainty should be even better. 
This thesis is designed to bridge this gap by replacing ad-hoc stochasticity with a formal Bayesian framework for MoE routing, aiming to achieve a new level of model robustness and reliability.
\chapter{Methodology:\\ Bayesian MoE Router}
\label{chap:methodology}

The preceding chapter established the core motivation for this work. 
This chapter details our proposed solution: a principled Bayesian framework designed to formalise stochasticity in MoE routing.

Our framework moves beyond single-point estimates by introducing probabilistic components into the routing pipeline. 
By modeling uncertainty in the \textbf{router's weights}, its \textbf{output logits} (similarity score), or the \textbf{final selection} process itself, each method induces a probabilistic belief over the expert choices. 
By doing so, we aim to achieve a more robust, well-calibrated expert selection mechanism, and extract better uncertainty signals to represent model's confidence.

\begin{figure}[H]
    \centering
    \includegraphics[width=\textwidth]{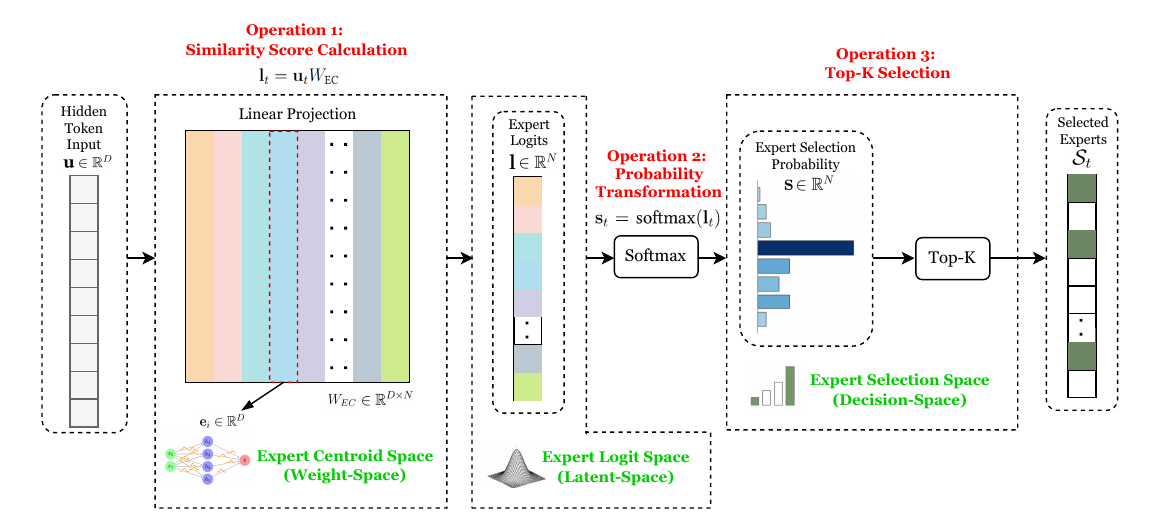}
    \caption{
        \textbf{Three Spaces for Bayesian Uncertainty in MoE Routing.}
        Illustration of the three distinct stages where uncertainty can be introduced: 
        (1) \emph{Expert Centroid Space} (weight-space), 
        (2) \emph{Expert Logit Space} (latent-space), and 
        (3) \emph{Expert Selection Space} (decision-space).
        Each corresponds to a different family of Bayesian routing methods described in this chapter.
    }
    \label{fig:bayesian_router_spaces}
\end{figure}

To systematically investigate this idea, 
we will present three distinct families of methods that introduce this uncertainty at different stages (as illustrated in Figure~\ref{fig:bayesian_router_spaces}): 
in the \textbf{expert centroid space} (weight-space), the \textbf{expert logit space} (latent-space), and the final \textbf{expert selection space} (decision-space). 
All methods are developed as efficient fine-tuning strategies designed to adapt a pre-trained MoE model, and this chapter will now detail each approach in turn.

\vspace{-0.5cm}

\section{Standard MoE Router: A Formal Definition}
\label{sec:methodology_standard_router}

Before detailing our Bayesian modifications, we formally define the standard deterministic routing process
\footnote{Already introduced in Chapter~\ref{chap:background}, but repeated here for clarity.}
. The pipeline begins by calculating a similarity score for each expert. For a given input token $\mathbf{u}_t$, the router computes a vector of unnormalized scores, or \textbf{logits} ($\mathbf{l}_t \in \mathbb{R}^N$), by projecting it with a learnable weight matrix, $W_{\text{EC}}$. This matrix is composed of $N$ column vectors, $W_{\text{EC}} = [\mathbf{e}_1, \dots, \mathbf{e}_N]$, where each vector $\mathbf{e}_i$ can be interpreted as a learnable centroid for an expert.
\begin{equation}
    \mathbf{l}_t = \mathbf{u}_t W_{\text{EC}}
\end{equation}
These logits are then transformed into a probability distribution over all $N$ experts using the softmax function, $\mathbf{s}_t = \text{softmax}(\mathbf{l}_t)$. Finally, a hard, deterministic \textbf{Top-K} selection mechanism is applied to this probability vector to identify the indices of the $K$ most probable experts. The probabilities for these selected experts are renormalized to sum to one, forming the final sparse \textbf{gating weights}, $\mathbf{g}_t$, which are used to compute the weighted sum of expert outputs. This completes the deterministic pipeline that our subsequent Bayesian methods aim to improve upon.

\section{Bayesian Inference on Expert Centroid Space}
\label{sec:methodology_weight_space}

First famliy of methods in our framework introduces Bayesian uncertainty at the earliest stage of the routing pipeline: Token-Expert similarity score calculation. 
This approach targets the router's linear projection layer, treating its weight matrix of expert centroids, $W_{\text{EC}}$, as a random variable $W_{\text{EC}}$. 
By doing so, we reframe standard routing mechanism into its principled Bayesian counterpart.

\subsection{Core Idea: Bayesian Multinomial Logistic Regression}

The standard MoE router, effectively a multinomial logistic regression model, learns a single, deterministic set of Expert Centroid Vectors as the model's weights (a point estimate). This approach reframes the process through a Bayesian lens by treating the router's weight matrix of expert centroids, $W_{\text{EC}}$, as a random variable. By doing so, we reformulate the standard routing mechanism into its principled Bayesian counterpart.

The goal of the router is to produce an expert selection probability distribution, $\mathbf{s}_t$, for a given input token hidden state, $\mathbf{u}_t$. The inference process is formalised as computing the posterior predictive distribution by marginalising over the router's weight posterior, $p(W_{\text{EC}}|\mathcal{D})$, which is approximated via Monte Carlo sampling:
\begin{align}
    p(\mathbf{s}_t | \mathbf{u}_t, \mathcal{D}) 
    &= \int p(\mathbf{s}_t|\mathbf{u}_t, W_{\text{EC}}) \, p(W_{\text{EC}}|\mathcal{D}) \, dW_{\text{EC}} \notag \\
    &\approx \frac{1}{S} \sum_{s=1}^{S} p(\mathbf{s}_t|\mathbf{u}_t, W_{\text{EC}}^s), \quad \text{where } W_{\text{EC}}^s \sim p(W_{\text{EC}}|\mathcal{D})
    \label{eq:weight_space_mc_inference}
\end{align}

In the language of neural networks, this inference process is implemented by averaging the softmax outputs from $S$ weight samples:

\begin{equation}
    \mathbf{s}_t \approx \frac{1}{S} \sum_{s=1}^{S} \text{softmax}(\mathbf{u}_t W_{\text{EC}}^s), \quad \text{where } W_{\text{EC}}^s \sim p(W_{\text{EC}}|\mathcal{D})
    \label{eq:nn_weight_space_mc_inference}
\end{equation}

\begin{figure}[H]
    \centering
    \includegraphics[width=\textwidth]{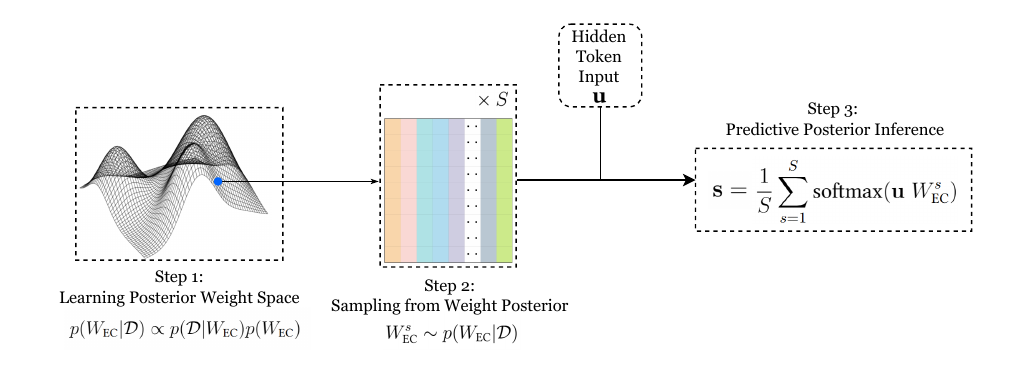}
    \caption{
        \textbf{Procedure for Bayesian MoE Routing on Expert Centroid Space.}
    }
    \label{fig:bayesian_expert_centroid_space}
\end{figure}

The entire process is illustrated in Figure~\ref{fig:bayesian_expert_centroid_space}.

This raises the central practical question: how can we obtain samples from the posterior distribution $p(W_{\text{EC}}|\mathcal{D})$? 
Since the true posterior is intractable to compute, we must rely on approximation methods. 
The following sections explore three distinct and powerful techniques for this purpose: \textbf{Monte Carlo Dropout}, \textbf{Stochastic Weight Averaging-Gaussian (SWAG)}, and \textbf{Deep Ensembles}.

\subsection{Method 1: MC Dropout Router (MCDR)}

Monte Carlo Dropout (MCD) is a straightforward and computationally efficient method for approximating the posterior predictive distribution.
Usually, stochastic dropout layers are employed during training as a regularisation, and are turned off during inference.
However, MC Dropout also performs random dropout at inference, effectively sampling from an approximate posterior distribution over the model weights.

In MoE Routing context, we apply dropout to the router's weight matrix $W_\text{EC}$ during both training and inference time, where each hidden unit is randomly dropped by sampling from a $\text{Bernoulli}(p)$ distribution.
Specifically, at inference time this procedure will be repeated $S$ times, each sampling results in a distinct model weight $W_\text{EC}^s$, thus achieving $S$ samples from posterior. 
Then by $S$ rounds of inference then averaging as in Eq.~\ref{eq:nn_weight_space_mc_inference}, we can obtain the final predictive distribution over experts.

\subsubsection*{In Practice}
For our implementation, we follow the standard and computationally efficient approach for MC Dropout. 
A dropout layer is inserted before the router's linear projection, applying a random binary mask to the input hidden state $\mathbf{u}_t$. 
The router is then fine-tuned, starting from the pre-trained MAP weights, by minimising a combined loss function that includes an L2 regularization term (weight decay):
\begin{equation}
    \mathcal{L}_{\text{MCDR}} = \mathcal{L}_{\text{task}} + \lambda ||W_{\text{EC}}||^2_F
\end{equation}
Here, $\mathcal{L}_{\text{task}}$ is the downstream task loss (e.g., cross-entropy), $||W_{\text{EC}}||^2_F$ is the squared Frobenius norm of the $D \times N$ expert centroid matrix, and $\lambda$ is the weight decay coefficient.

This specific training objective, combining dropout on the input units with L2 regularisation, is what allows the model to be interpreted as a form of 
\textbf{approximate variational inference for a deep Gaussian Process}~\cite{gal2016dropout}. 
At inference time, after obtaining the Monte Carlo average of the routing probabilities $\textbf{s}_t$, 
the standard deterministic \textbf{Top-K} mechanism is used to select the final set of experts.

\subsection{Method 2: Stochastic Weight Averaging Gaussian Router (SWAGR)}

The SWAG procedure begins after the router has been fine-tuned to convergence. 
We continue training for a number of epochs with a high, constant learning rate, collecting the expert centroid matrix $W_{\text{EC}}^s$ at each step $i$. 
The first two moments of these collected weights are used to define the approximate Gaussian posterior, $p(W_{\text{EC}}|\mathcal{D}) \approx \mathcal{N}(\bar{W}_{\text{EC}}, \Sigma_{\text{SWAG}})$. 
The mean of this posterior is the running average of the weights:
\begin{equation}
    \bar{W}_{\text{EC}} = \frac{1}{S}\sum_{s=1}^{S} W_{\text{EC}}^s
\end{equation}
The covariance matrix, $\Sigma_{\text{SWAG}}$, is constructed using the second moment of the iterates, capturing the geometry of the loss surface.

\subsubsection*{In Practice}
A crucial practical aspect of SWAG is the storage and computation of the covariance matrix. 
A full-rank covariance matrix for the $D \times N$ weights would be prohibitively large. 
Therefore, we use a \textbf{low-rank} plus \textbf{diagonal approximation}. 
This involves storing the running average of the weights ($\bar{W}_{\text{EC}}$), the running average of the squared weights (for the diagonal part), and a small number of recent weight vectors to form the low-rank deviation matrix. 
At inference time, we draw $S$ weight matrix samples $W_{\text{EC}}^s$ from this approximate Gaussian posterior. 
Each sample is used to calculate a logit vector, and the final routing probabilities are obtained by averaging the post-softmax outputs as described in Eq.~\ref{eq:nn_weight_space_mc_inference} as usual, followed by the standard Top-K selection.

\subsection{Method 3: Deep Ensembles of Routers (DER)}

The third method, the Deep Ensemble Router, is an implicit and non-parametric approach to approximating the posterior predictive distribution, following the work of Lakshminarayanan et al.~\cite{lakshminarayanan2017simple}. 
Instead of defining and approximating an explicit posterior distribution, this method leverages the diversity created by training multiple models independently.

The core idea is to treat the collection of independently trained models as a set of empirical samples from the true, unknown posterior distribution. 
Each of the $M$ routers in the ensemble is trained to convergence, finding a different mode in the loss landscape. 
This collection of final weight matrices, 
$\{W_{\text{EC}}^1, \dots, W_{\text{EC}}^M\}$, is then assumed to be a representative set of samples from $p(W_{\text{EC}}|\mathcal{D})$.

\subsubsection*{In Practice}
To implement DER, we train an ensemble of $M$ separate router weights. Each member is fine-tuned from the same pre-trained MAP weights but with a different random seed for its optimiser state and data shuffling to encourage functional diversity. 
At inference time, an input token $\mathbf{u}_t$ is passed through all $M$ routers in the ensemble, 
producing $M$ distinct logit vectors. 
Each logit vector is passed through a softmax function, and the resulting $M$ probability distributions are averaged to approximate the Bayesian model average, as shown in Eq.~\ref{eq:nn_weight_space_mc_inference} still. 
This final, robust probability distribution is then used for the standard Top-K selection of experts.

\subsection{Summary of Centroid-Space Methods}
\textbf{Pros:} The methods in this category provide a principled approach to routing uncertainty by applying classic BNN techniques directly to expert centroid matrix $W_{\text{EC}}$. 
By approximating posterior over weights, these methods capture true epistemic uncertainty. 
Their main advantage lies in this strong theoretical grounding and, in the case of \textbf{MCDR}, their simplicity and ease of implementation.

\textbf{Cons:} A key conceptual limitation of this approach is its \textbf{indirectness}. 
These methods model uncertainty in the high-dimensional weight-space, which must then propagate through a linear transformation to induce a distribution on the low-dimensional logit-space,
subsequently making it an inefficient way to represent uncertainty.

This raises a natural question: 
Can we model the uncertainty more directly? 
Instead of modeling the cause (uncertainty in the weights), can we directly model the effect (uncertainty in the logits)? 
This motivation leads us to the next family of methods.

\section{Bayesian Inference on Expert Logit Space}
\label{sec:methodology_logit_space}

This section explores a more direct and potentially more expressive alternative: applying Bayesian inference directly to the logit space itself. 
By modeling a probability distribution over the logit vector $l$, the quantity that immediately governs the final expert selection, 
we can create a more targeted representation of routing uncertainty. 
This section will develop this idea, starting by framing it as a probabilistic graphical model and then detailing two specific implementations of this strategy.

\subsection{Core Idea: Amortised Variational Inference on the Logit Space}

\subsubsection*{Probabilistic Graphical Model (PGM) Framing}
\begin{figure}[H]
    \centering
    \begin{tikzpicture}[
        >=Latex, 
        node distance=0.8cm and 0.8cm, 
        state/.style={circle, draw, minimum size=1cm, inner sep=0pt}, 
        obs/.style={state, fill=gray!60}, 
        plate/.style={draw, rounded corners, thick, inner sep=0.5cm}, 
        conn/.style={->, thick}, 
    ]

    \node[obs] (x) {$x$};
    \node[state, right=of x] (u1) {$\mathbf{u}_1$};
    \node[state, right=of u1] (u2) {$\mathbf{u}_2$};
    \node[right=0.1cm of u2] (d1) {$\dots$};
    \node[state, right=0.1cm of d1] (ui) {$\mathbf{u}_i$};
    \node[state, right=of ui] (ui1) {$\mathbf{u}_{i+1}$};
    \node[right=0.1cm of ui1] (d2) {$\dots$};
    \node[state, right=0.1cm of d2] (uL) {$\mathbf{u}_L$};
    \node[state, right=of uL] (h) {$\mathbf{h}$};
    \node[obs, right=of h] (y) {$y$};

    \node[state, below=of u1] (L1) {$\mathbf{l}_1$};
    \node[state, below=of u2] (L2) {$\mathbf{l}_2$};
    \node[state, below=of ui] (Li) {$\mathbf{l}_i$};
    \node[state, below=of ui1] (Li1) {$\mathbf{l}_{i+1}$};
    \node[state, below=of uL] (LL) {$\mathbf{l}_L$};
    \node[below=1.5cm of d1] (d3) {$\cdots$};
    \node[below=1.5cm of d2] (d4) {$\cdots$};

    \draw[conn] (x) -- (u1);
    \draw[conn] (u1) -- (u2);
    \draw[conn] (ui) -- (ui1);
    \draw[conn] (uL) -- (h);
    \draw[conn] (h) -- (y);
    \draw[conn] (u1) -- (L1);
    \draw[conn] (u2) -- (L2);
    \draw[conn] (ui) -- (Li);
    \draw[conn] (ui1) -- (Li1);
    \draw[conn] (uL) -- (LL);
    \draw[conn] (L1) -- (u2);
    \draw[conn] (Li) -- (ui1);
    \draw[conn] (LL) -- (h);

    \draw[conn] ([yshift=-.8cm]L1.south) -- (L1);
    \node[below=0 of L1, yshift=-.8cm] {$\phi_{1}$};

    \draw[conn] ([yshift=-.8cm]L2.south) -- (L2);
    \node[below=0 of L2, yshift=-.8cm] {$\phi_{2}$};

    \draw[conn] ([yshift=-.8cm]Li.south) -- (Li);
    \node[below=0 of Li, yshift=-.8cm] {$\phi_{i}$};

    \draw[conn] ([yshift=-.8cm]Li1.south) -- (Li1);
    \node[below=0 of Li1, yshift=-.8cm] {$\phi_{i+1}$};

    \draw[conn] ([yshift=-.8cm]LL.south) -- (LL);
    \node[below=0 of LL, yshift=-.8cm] {$\phi_{L}$};

    \node[plate, fit=(u1) (L1) (uL) (LL) (h) (x) (y), label={[anchor=south east]south east:$\times N$}] (plate) {};

    \end{tikzpicture}
    \caption{PGM of the full hierarchical MoE LLM.}
    \label{fig:pgm_full_moe}
\end{figure}
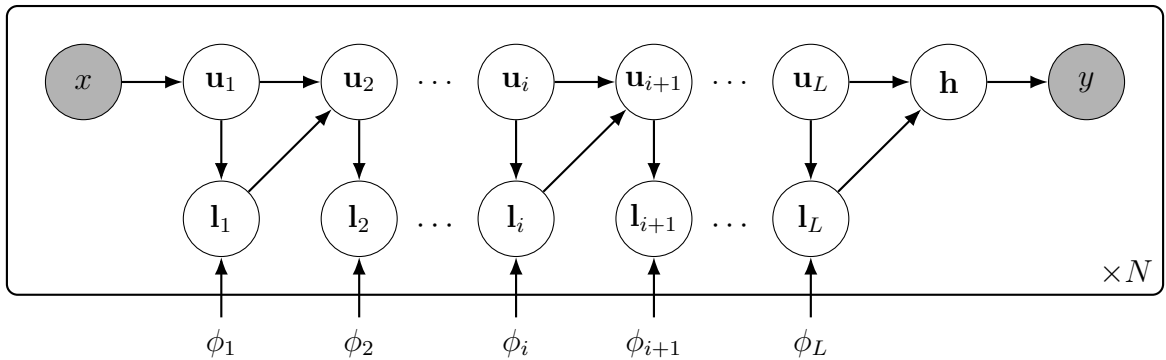

To formally ground our approach, we first view the entire MoE LLM as a deep, hierarchical latent variable model, as depicted in Figure~\ref{fig:pgm_full_moe}. 
In this model, the input sequence tokens $x$ and final output next token $y$ are observed variables, 
while the hidden states before each MoE layer, $\{\mathbf{u}_1, \mathbf{u}_2, \ldots, \mathbf{u}_L\}$, and the expert logit vectors at each MoE layer, $\{\mathbf{l}_1, \mathbf{l}_2, \ldots, \mathbf{l}_L\}$, are latent variables.
The final hidden state $\mathbf{h}$ before output projection is also a latent variable.
At each layer, hidden state $\mathbf{u}_i$ generates a latent logit vector $\mathbf{l}_i$, which in turn together determines the next hidden state $\mathbf{u}_{i+1}$.
Additionally, $L$ represents total number of MoE layers, and $N$ is the size of finetuning dataset. 

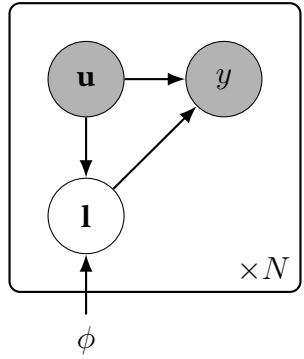
\begin{figure}[H]
    \centering
    \begin{tikzpicture}[
        >=Latex, 
        node distance=0.8cm and 0.8cm, 
        state/.style={circle, draw, minimum size=1cm, inner sep=0pt}, 
        obs/.style={state, fill=gray!60}, 
        plate/.style={draw, rounded corners, thick, inner sep=0.5cm}, 
        conn/.style={->, thick}, 
    ]

    \node[obs] (u) {$\textbf{u}$};
    \node[obs, right=of u] (y) {$y$};

    \node[state, below=of u] (l) {$\textbf{l}$};

    \draw[conn] (u) -- (y);
    \draw[conn] (u) -- (l);
    \draw[conn] (l) -- (y);

    \draw[conn] ([yshift=-.8cm]l.south) -- (l);
    \node[below=0 of l, yshift=-.8cm] {$\phi$};

    \node[plate, fit=(u) (y) (l), label={[anchor=south east]south east:$\times N$}] (plate) {};

    \end{tikzpicture}
    \caption{Simplified PGM for a single MoE layer used for our analysis.}
    \label{fig:pgm_simplified_moe}
\end{figure}

Inference on every logit space together would be challenging due to the hierarchical structure. 
To address this, we adopt a principled simplification: 
we analyse \textbf{one MoE layer at a time}, treating all other layers as \textbf{deterministic and frozen}. 
As the subsequent layers (Including all the following attention and MoE FFN mechanisms) are just deterministic functions of the current layer's output, 
we can simplify the graphical model to only the essential variables for our learning task, as shown in Figure~\ref{fig:pgm_simplified_moe}. 
The model reduces to inferring the latent logit vector $\textbf{l}$ for a given layer, conditioned on its observed input $\textbf{u}$ and the final observed task output $y$.

\subsubsection*{Variational Inference Formulation}
Our goal is to infer the posterior distribution over the logits, $p(\mathbf{l}|\mathbf{u}, y)$. 
As this is intractable, we use variational inference to approximate it with a tractable distribution, $q_\phi(\mathbf{l}|\mathbf{u})$. 
We assume this approximate posterior is a multivariate Gaussian.
The parameters $\phi$ of this distribution are learned by maximising the Evidence Lower Bound (ELBO):
\begin{equation}
    \mathcal{L}_\text{ELBO}(\phi) = \underbrace{\mathbb{E}_{q_\phi(\mathbf{l}|\mathbf{u})}[\log p(y|\mathbf{l}, \mathbf{u})]}_{\text{Reconstruction Term}} - \underbrace{\KL(q_\phi(\mathbf{l}|\mathbf{u}) || p(\mathbf{l} | \mathbf{u}))}_{\text{Regularisation Term}}
\end{equation}

Here, $p(\mathbf{l} | \mathbf{u})$ is the prior we choose for the logits, which will be defined later.

The \textbf{reconstruction term} corresponds to the downstream task loss, ensuring that the latent logits are useful for the final prediction. 
The \textbf{regularisation term} is the KL divergence between our learned posterior and a simple prior, which prevents the model from becoming overconfident.

\subsubsection*{Amortised Inference and Residual Learning}
Inspired by the Variational Autoencoder (VAE), 
we use a neural network, or the \textbf{variational router}, to perform \textbf{amortised inference}.
This network learns a single function that maps any input token $\mathbf{u}$ directly to the parameters of its corresponding posterior $q_\phi(\mathbf{l}|\mathbf{u})$,
which corresponds to $\boldsymbol{\mu}_{\text{post}}(\textbf{u})$ and $\boldsymbol{\Sigma}_{\text{post}}(\textbf{u})$ in this case (Mutivriate Gaussian).

To make full use of the pre-trained routing weights in deterministic router, 
we implement the posterior mean inference network using a \textbf{residual learning} mechanism. 
Instead of predicting the posterior mean directly, the network predicts a residual correction, $\Delta\boldsymbol{\mu}_\phi(\cdot)$, which is added to the original deterministic logits, $\text{NN}_{\text{det}}(\cdot)$:
\begin{equation}
    \boldsymbol{\mu}_{\text{post}} = \text{NN}_{\text{det}}(\textbf{u}) + \Delta\boldsymbol{\mu}_\phi(\textbf{u})
\end{equation}
This formulation provides a significant computational benefit. By setting the prior $p(\mathbf{l} | \mathbf{u})$ to be a Gaussian centered on the deterministic logits, 
$p(\mathbf{l} | \mathbf{u}) = \mathcal{N}(\mathbf{l}|\text{NN}_{\text{det}}(\textbf{u}), I)$, the KL divergence term in the ELBO simplifies. 
The KL divergence between the full posterior and the prior becomes equivalent to the KL divergence between the learned residual and a standard normal prior
\footnote{Proof in Appendix~\ref{app:kl_proof}.}:
\begin{align}
    & \KL(\mathcal{N}(\text{NN}_{\text{det}}(\textbf{u}) + \Delta\boldsymbol{\mu}_\phi(\textbf{u}), \boldsymbol{\Sigma}_\text{post}) \,||\, \mathcal{N}(\text{NN}_{\text{det}}(\textbf{u}), I)) \notag \\
    = &\KL(\mathcal{N}(\Delta\boldsymbol{\mu}_\phi(\textbf{u}), \boldsymbol{\Sigma}_\text{post}) \,||\, \mathcal{N}(0, I))
\end{align}


\begin{figure}[H]
    \centering
    \includegraphics[width=\textwidth]{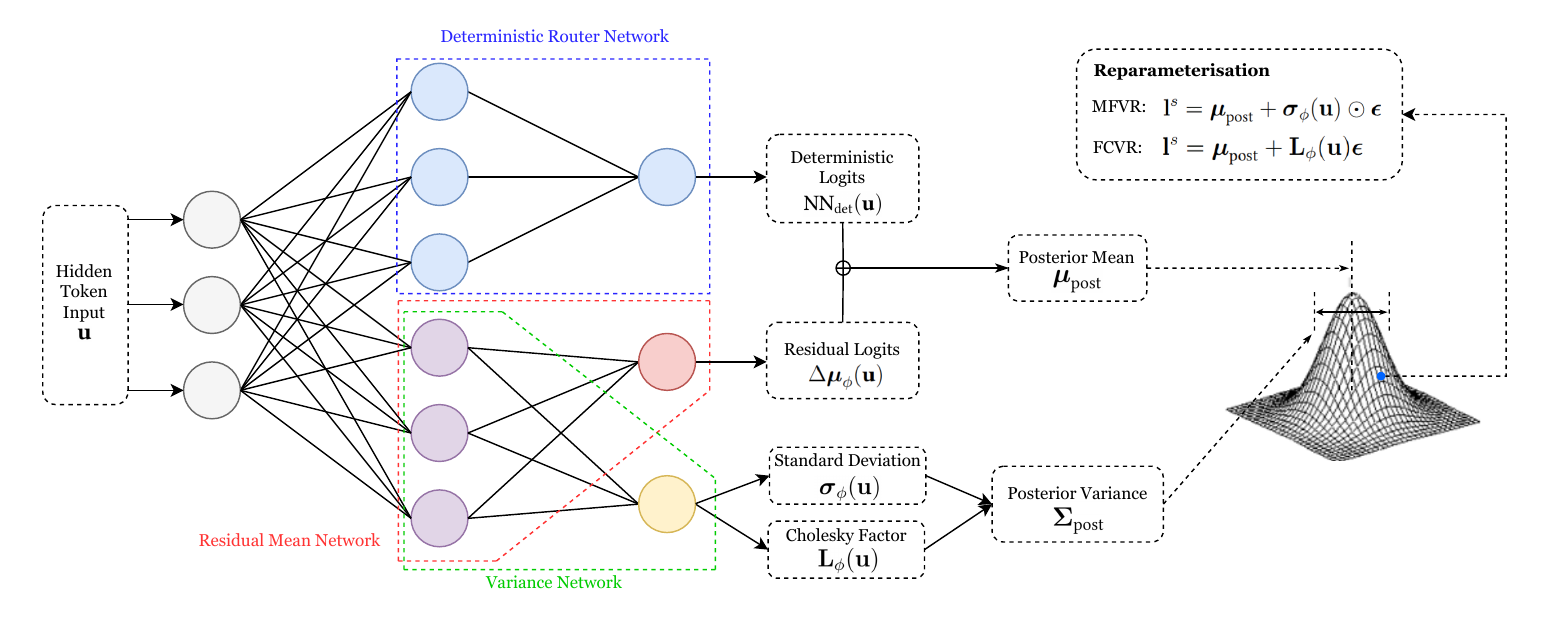}
    \caption{
        \textbf{Variational Router Illustration.}
        Variational router predicts a Gaussian posterior over the logits, 
        with a mean given by the deterministic logits plus a learned residual and variance. 
        A sample from this posterior is drawn by reparameterisation trick, 
        and resulting logits are used to compute routing probabilities.
    }
    \label{fig:variational_router}
\end{figure}

\subsection{Method 4: The Mean-Field Variational Router (MFVR)}

The Mean-Field Variational Router (MFVR) is the first and simplest implementation of our logit-space framework. It is based on the \textbf{mean-field assumption}, 
which posits that the posterior distribution over the logits can be factorised into independent univariate Gaussians for each of the $N$ experts. 
This implies that the covariance matrix of our approximate posterior, $\boldsymbol{\Sigma}_{\text{post}}(\mathbf{u})$, is a diagonal matrix.

\subsubsection*{Reparameterisation Trick}
To implement this, the variational router has a network head that outputs the log-standard deviation vector, $\log\boldsymbol{\sigma}_\phi(\cdot)$. 
A sample from the posterior is then generated using the standard element-wise reparameterisation trick:
\begin{equation}
    \mathbf{l}^s = \boldsymbol{\mu}_{\text{post}} + \boldsymbol{\sigma}_\phi(\mathbf{u}) \odot \boldsymbol{\epsilon}, \quad \text{where } \boldsymbol{\epsilon} \sim \mathcal{N}(0, I)
\end{equation}

\subsubsection*{Loss Function}
The parameters of the variational router, $\phi$, are learned by minimising a loss function derived from a single-sample Monte Carlo estimate of the ELBO. 
Since KL divergence between two diagonal Gaussians has a closed-form solution, the KL loss for this mean-field case simplifies to:
\begin{equation}
    \mathcal{L}_\text{MF-KL} = \frac{1}{2} \sum_{i=1}^{N} \left( (\Delta\mu_i)^2 + \sigma_i^2 - \log(\sigma_i^2) - 1 \right)
\end{equation}
where:
\begin{itemize}
    \setlength\itemsep{0.05em} 
    \item $N$ is the total number of experts.
    \item $\Delta\mu_i$ is the $i$-th component of the learned residual mean vector $\Delta\boldsymbol{\mu}_\phi(\mathbf{u})$.
    \item $\sigma_i^2$ is the $i$-th component of the learned variance vector $\boldsymbol{\sigma}^2_\phi(\mathbf{u})$.
\end{itemize}

A hyperparameter, $\beta$, is introduced to scale the KL term, similar to its use in Variational Auto Encoders (VAEs)~\cite{kingma2013auto} to balance the reconstruction and regularisation objectives:

\begin{equation}
    \mathcal{L}_{\text{MFVR}} = \mathcal{L}_{\text{task}} + \beta \cdot \mathcal{L}_\text{MF-KL}
\end{equation}

\subsubsection*{Training and Inference Sampling}

At training time, for each input token $\mathbf{u}$, we perform a single reparameterisation trick in logit space to obtain a sample of the logits, $\mathbf{l}^s$, 
then perform end-to-end training to update variational router's parameters $\phi$.

At inference time, we want a more accurate approximation of the posterior predictive distribution on the expert selection probablity, 
so we perform $S$ independent reparameterisation samples, $\{\mathbf{l}^1, \mathbf{l}^2, \ldots, \mathbf{l}^S\}$,
and average their post-softmax outputs to obtain the final routing probability.

\begin{figure}[H]
    \centering
    \includegraphics[width=\textwidth]{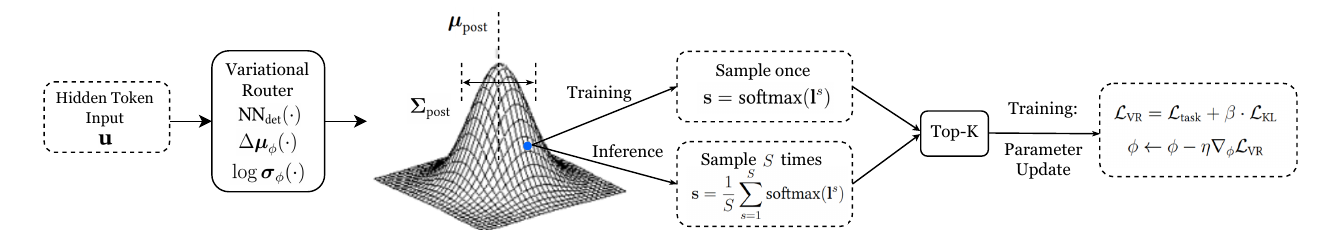}
    \caption{
        \textbf{Training and Inference Procedures for Variational Router.}
        Comparison of the training and inference data flows for the Variational Router. 
        During training (top), a single sample is used to compute a stochastic loss. 
        During inference (bottom), multiple samples are drawn and their post-softmax probabilities are averaged to produce a robust routing decision.
    }
    \label{fig:variational_router_training_inference}
\end{figure}

The training and inference procedures are illustrated in Figure~\ref{fig:variational_router_training_inference} and detailed in Algorithm~\ref{alg:mfvr}.

\subsection{Method 5: The Full-Covariance Variational Router (FCVR)}
\label{sec:methodology_fcvr}

The Full-Covariance Variational Router (FCVR) is a more expressive extension that relaxes the mean-field assumption. 
By modeling a \textbf{full-rank covariance matrix}, the FCVR can capture potential correlations between the logits of different experts, allowing for a richer and more flexible approximate posterior.

\subsubsection*{Reparameterisation Trick}
To ensure the covariance matrix remains positive semi-definite, 
the variational router is trained to output the elements of its \textbf{Cholesky factor}, $\mathbf{L}_\phi(\mathbf{u})$, where:
\begin{equation}
    \boldsymbol{\Sigma}_\text{post} = \mathbf{L}_\phi(\mathbf{u}) \mathbf{L}_\phi(\mathbf{u})^\top
\end{equation}

The reparameterization trick for the multivariate case is then used to generate a sample:
\begin{equation}
    \mathbf{l}^s = \boldsymbol{\mu}_{\text{post}} + \mathbf{L}_\phi(\mathbf{u}) \boldsymbol{\epsilon}, \quad \text{where } \boldsymbol{\epsilon} \sim \mathcal{N}(0, I)
\end{equation}

\subsubsection*{Loss Function}
The parameters of the Full-Covariance Variational Router are also learned by minimising the loss function derived from the ELBO. 
The key difference lies in the KL divergence term, which now measures the divergence between two full-rank multivariate Gaussians.
 This also has a closed-form analytical solution:
\begin{equation}
    \mathcal{L}_\text{FC-KL} = \frac{1}{2} \left( \text{tr}(\boldsymbol{\Sigma}_\text{post}) + ||\Delta\boldsymbol{\mu}||_2^2 - N - \log|\boldsymbol{\Sigma}_\text{post}| \right)
\end{equation}
where:
\begin{itemize}
    \setlength\itemsep{0.05em} 
    \item $N$ is the total number of experts.
    \item $\text{tr}(\boldsymbol{\Sigma}_\text{post})$ is the trace of the covariance matrix.
    \item $||\Delta\boldsymbol{\mu}||_2^2$ is the squared L2 norm of the residual mean vector $\Delta\boldsymbol{\mu}_\phi(\mathbf{u})$.
    \item $\log|\boldsymbol{\Sigma}_\text{post}|$ is the log-determinant of the covariance matrix, which can be computed efficiently from the Cholesky factor as $2 \sum_i \log(\text{diag}(\mathbf{L_\phi(\textbf{u})})_i)$.
\end{itemize}
As with the mean-field case, a hyperparameter $\beta$ is used to scale the KL term, yielding the final loss function:
\begin{equation}
    \mathcal{L}_{\text{FCVR}} = \mathcal{L}_{\text{task}} + \beta \cdot \mathcal{L}_\text{FC-KL}
\end{equation}

\subsubsection*{Training and Inference Sampling}
The training and inference procedures for the FCVR are identical to those of the MFVR, as detailed in Algorithm~\ref{alg:fcvr}. The only difference is the specific reparameterisation step used to generate the logit sample $\mathbf{l}^s$, which now incorporates the full Cholesky factor to capture correlations.

\begin{figure}[H]
    \centering
    \begin{minipage}{0.48\textwidth}
        \begin{algorithm}[H]
            \caption{MFVR Training and Inference}
            \label{alg:mfvr}
            \begin{algorithmic}[1]
                \Statex \textbf{Training (one step for input $\mathbf{u}$, target $y$):}
                \State $\mathbf{l}_{\text{det}} \leftarrow \text{NN}_{\text{det}}(\mathbf{u})$
                \State \impb{$\Delta\boldsymbol{\mu}, \boldsymbol{\sigma} \leftarrow \Delta\boldsymbol{\mu}_{\phi}(\mathbf{u}), \boldsymbol{\sigma}_{\phi}(\mathbf{u})$}
                \State $\boldsymbol{\mu}_{\text{post}} \leftarrow \mathbf{l}_{\text{det}} + \Delta\boldsymbol{\mu}$
                \State $\boldsymbol{\epsilon} \sim \mathcal{N}(0, I)$
                \State \impb{$\mathbf{l}^s \leftarrow \boldsymbol{\mu}_{\text{post}} + \boldsymbol{\sigma} \odot \boldsymbol{\epsilon}$}
                \State Select experts using $\text{Top-K}(\text{softmax}(\mathbf{l}^s))$, get model final output $\hat{y}$
                \State \impb{Compute $\mathcal{L}_{\text{MFVR}}$ using $\hat{y}$ and $y$}
                \State \impb{Update $\phi$ using $\nabla_\phi \mathcal{L}_{\text{MFVR}}$}
                \Statex
                \Statex \textbf{Inference (for input $\mathbf{u}$):}
                \State $\mathbf{l}_{\text{det}} \leftarrow \text{NN}_{\text{det}}(\mathbf{u})$
                \State \impb{$\Delta\boldsymbol{\mu}, \boldsymbol{\sigma} \leftarrow \Delta\boldsymbol{\mu}_{\phi}(\mathbf{u}), \boldsymbol{\sigma}_{\phi}(\mathbf{u})$}
                \State $\boldsymbol{\mu}_{\text{post}} \leftarrow \mathbf{l}_{\text{det}} + \Delta\boldsymbol{\mu}$
                \State $\mathbf{p}_{\text{avg}} \leftarrow \mathbf{0}$
                \For{$s = 1$ to $S$}
                \State $\boldsymbol{\epsilon '} \sim \mathcal{N}(0, I)$
                    \State \impb{$\mathbf{l}^s \leftarrow \boldsymbol{\mu}_{\text{post}} + \boldsymbol{\sigma} \odot \boldsymbol{\epsilon '}$}
                    \State $\mathbf{p}_{\text{avg}} \leftarrow \mathbf{p}_{\text{avg}} + \text{softmax}(\mathbf{l}^s)$
                \EndFor
                \State Select experts using $\text{Top-K}(\frac{\mathbf{p}_{\text{avg}}}{S})$
            \end{algorithmic}
        \end{algorithm}
    \end{minipage}
    \hfill 
    \begin{minipage}{0.48\textwidth}
        \begin{algorithm}[H]
            \caption{FCVR Training and Inference}
            \label{alg:fcvr}
            \begin{algorithmic}[1]
                \Statex \textbf{Training (one step for input $\mathbf{u}$, target $y$):}
                \State $\mathbf{l}_{\text{det}} \leftarrow \text{NN}_{\text{det}}(\mathbf{u})$
                \State \impb{$\Delta\boldsymbol{\mu}, \mathbf{L} \leftarrow \Delta\boldsymbol{\mu}_{\phi}(\mathbf{u}), \mathbf{L}_{\phi}(\mathbf{u})$}
                \State $\boldsymbol{\mu}_{\text{post}} \leftarrow \mathbf{l}_{\text{det}} + \Delta\boldsymbol{\mu}$
                \State $\boldsymbol{\epsilon} \sim \mathcal{N}(0, I)$
                \State \impb{$\mathbf{l}^s \leftarrow \boldsymbol{\mu}_{\text{post}} + \mathbf{L} \boldsymbol{\epsilon}$}
                \State Select experts using $\text{Top-K}(\text{softmax}(\mathbf{l}^s))$, get model final output $\hat{y}$
                \State \impb{Compute $\mathcal{L}_{\text{FCVR}}$ using $\hat{y}$ and $y$}
                \State \impb{Update $\phi$ using $\nabla_\phi \mathcal{L}_{\text{FCVR}}$}
                \Statex
                \Statex \textbf{Inference (for input $\mathbf{u}$):}
                \State $\mathbf{l}_{\text{det}} \leftarrow \text{NN}_{\text{det}}(\mathbf{u})$
                \State \impb{$\Delta\boldsymbol{\mu}, \mathbf{L} \leftarrow \Delta\boldsymbol{\mu}_{\phi}(\mathbf{u}), \mathbf{L}_{\phi}(\mathbf{u})$}
                \State $\boldsymbol{\mu}_{\text{post}} \leftarrow \mathbf{l}_{\text{det}} + \Delta\boldsymbol{\mu}$
                \State $\mathbf{p}_{\text{avg}} \leftarrow \mathbf{0}$
                \For{$s = 1$ to $S$}
                    \State $\boldsymbol{\epsilon'} \sim \mathcal{N}(0, I)$
                    \State \impb{$\mathbf{l}^s \leftarrow \boldsymbol{\mu}_{\text{post}} + \mathbf{L} \boldsymbol{\epsilon'}$}
                    \State $\mathbf{p}_{\text{avg}} \leftarrow \mathbf{p}_{\text{avg}} + \text{softmax}(\mathbf{l}^s)$
                \EndFor
                \State Select experts using $\text{Top-K}(\frac{\mathbf{p}_{\text{avg}}}{S})$
            \end{algorithmic}
        \end{algorithm}
    \end{minipage}
\end{figure}

\subsection{Summary of Logit-Space Methods}
The logit-space methods provide a more direct and expressive approach to routing uncertainty. 
By placing a learned, input-dependent Gaussian distribution directly over the expert logits, 
these methods, particularly \textbf{FCVR}, can capture complex correlations and provide a rich representation of the model's belief, 
leading to state-of-the-art performance.

However, this approach still faces a key limitation: 
The distribution that results from applying the softmax function to a Gaussian is still intractable. 
This forces us to rely on Monte Carlo sampling at inference time,
drawing multiple samples from the logit space and averaging their post-softmax probabilities,
which can be computationally expensive.

This leads to a final, crucial question: 
is it possible to introduce principled, input-dependent stochasticity without the need for multi-sample Monte Carlo averaging?
Also, taking inspiration from our earlier motivation experiments in Section~\ref{sec:motivation2}, 
this motivates the final family of methods, which operate directly on the expert selection space.

\section{Bayesian Inference on Expert Selection Space}
\label{sec:methodology_selection_space}

A prominent challenge of modeling uncertainty in the logit space is that the softmax of a Gaussian distribution is intractable. This necessitates the use of Monte Carlo sampling to approximate the posterior predictive distribution over the post-softmax routing probabilities, which we refer to as the \textbf{expert selection space}. This raises a natural question: can we model the uncertainty of the routing decision more directly in this final selection space?

\subsection{Core Idea: Learning Input-Dependent Temperature}
Our key inspiration comes from the motivation experiment in Section~\ref{sec:motivation2}. We observed that replacing the deterministic Top-K selection with a Sample-K strategy, 
governed by a global temperature parameter $T$, could improve model calibration. 
However, a single, fixed temperature is a blunt instrument, the optimal level of stochasticity is likely token-dependent. 
An easy token should be routed with high confidence (low temperature), while an ambiguous or out-of-distribution token should be routed with high uncertainty (high temperature).

This motivates a natural extension: to learn an input-dependent temperature, $T(\mathbf{u})$, allowing the model to dynamically control the stochasticity of its own routing decisions. 
The job of learning this variational temperature function is delegated to a neural network, and we call this approach the \textbf{Variational Temperature Sampling Router (VTSR)}.

\begin{figure}[H]
    \centering
    \includegraphics[width=\textwidth]{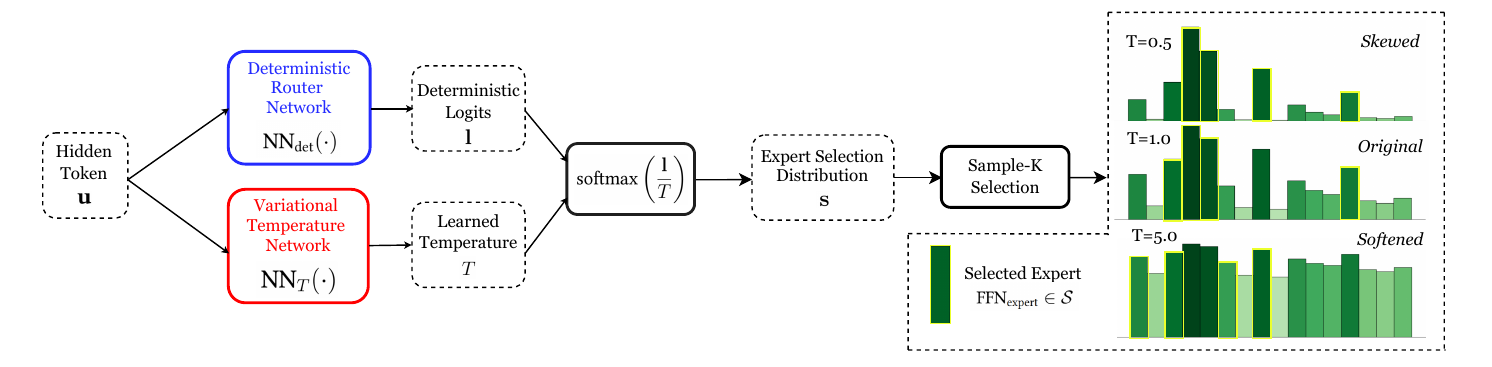}
    \caption{
        \textbf{Variational Temperature Sampling Router (VTSR).}
        Illustration of the VTSR approach: a neural network predicts an input-dependent temperature that scales the deterministic logits. This scaled distribution is then used for sampling experts, allowing the model to adapt its routing uncertainty based on the input token.
    }
    \label{fig:temperature_sampling_router}
\end{figure}

\subsection{Method 6: Variational Temperature Sampling Router (VTSR)}

The Variational Temperature Sampling Router is a pragmatic method designed to learn an optimal, input-dependent level of routing stochasticity. \
It consists of a small neural network that takes the token embedding $\mathbf{u}$ as input and outputs a single positive scalar value, the temperature $T = \text{NN}_T(\textbf{u})$. 
This temperature is then used to scale the deterministic logits generated by the original deterministic routing network $\mathbf{l} = \text{NN}_\text{det}(\mathbf{u})$ before a sampling operation, as opposed to the deterministic Top-K operation, selects the final experts.
Schematics of the VTSR approach is illustrated in Figure~\ref{fig:temperature_sampling_router}.


\subsubsection*{Training with the Gumbel-Softmax Trick}
A key challenge during training is that the process of sampling $K$ experts from the temperature-scaled distribution is non-differentiable, which breaks the flow of gradients. 
To overcome this, we employ the \textbf{Gumbel-Softmax trick}
\footnote{We don't explain details of Gumbel-Softmax trick here due to space limit, please refer to original papers~\cite{jang2016categorical, maddison2016concrete} for more information.}
(also known as the Concrete distribution). 
This technique provides a continuous, differentiable approximation to the discrete sampling process, allowing gradients to flow back to both the main router weights and the temperature prediction network.
At inference time, we use the learned temperature to scale the logits and perform multinomial sampling without Gumbel noise or relaxation.

\subsubsection*{Regularisation to Prevent Deterministic Collapse}
A network trained to predict $T(\mathbf{u})$ could learn to minimise the task loss by simply setting the temperature to be very low for all inputs, 
effectively collapsing back to a deterministic Top-K router. 
To prevent this, we introduce a regularisation term to the loss function that encourages the model to maintain a degree of uncertainty. 
Inspired by the uncertainty modeling work of Kendall \& Gal in~\cite{kendall2017uncertainties}, we penalise low temperatures by minimising the expected log-temperature, approximated as a within-batch average:
\begin{equation}
    \mathcal{L}_{\text{temp}} = - \frac{1}{B} \sum_{i=1}^B \log(\text{NN}_T(\mathbf{u}_i))
\end{equation}
where $B$ is the batch size and $T(\mathbf{u}_i)$ is the predicted temperature for the $i$-th input in the batch.
This regularisation term can be interpreted as encouraging entropy in the routing policy, forcing the model to only become confident (low temperature) when there is sufficient evidence in the data. 
The final training objective is a weighted sum of the task loss and this regularization term:
\begin{equation}
    \mathcal{L}_{\text{VTSR}} = \mathcal{L}_{\text{task}} + \beta \cdot \mathcal{L}_{\text{reg}}
\end{equation}
At inference time, we use the predicted temperature $T(\mathbf{u})$ to scale the logits and then perform a direct (non-Gumbel) sampling of $K$ experts from the resulting softmax distribution.

\subsection{Summary of the Selection-Space Method}

The key advantage of the final method, the \textbf{Variational Temperature Sampling Router (VTSR)}, is its exceptional efficiency. 
By learning an input-dependent temperature to control a single sampling step, it introduces principled stochasticity without the computational overhead of Monte Carlo averaging, making it ideal for latency-critical applications.

However, this theoretical elegance is offset by practical instability. 
Our experiments found the training to be challenging, with the learned temperature often suffering from \textbf{posterior collapse} even with regularisation. 
This resulted in a less reliable uncertainty signal for OoD detection compared to the more robust variational methods.

Ultimately, the value of the VTSR lies in its novel conceptual contribution: it successfully decouples routing stochasticity from multi-sample inference. 
While it requires further research to stabilise its training, it represents a promising and computationally efficient direction for future work.

\section{Chapter Summary}

This chapter has introduced a comprehensive framework for applying principled Bayesian uncertainty to the Mixture-of-Experts routing mechanism. 
We have detailed three distinct families of methods, each targeting a different conceptual space in the routing pipeline: 
the \textbf{Expert Centroid Space} (weight-space), 
the \textbf{Expert Logit Space} (latent-space), and 
the \textbf{Expert Selection Space} (decision-space).

\begin{table}[H]
    \centering
    \caption{A comprehensive summary of the proposed Bayesian routing methods.}
    \label{tab:methodology_summary}
    \resizebox{\textwidth}{!}{\begin{tabular}{lllccc}
        \toprule
        \textbf{Family} & \textbf{Model} & \textbf{Bayesian Technique} & \textbf{Source of Uncertainty} & \textbf{Requires Extra NN?} & \textbf{Inference Mechanism} \\
        \midrule
        \multirow{3}{*}{\makecell[l]{Expert Centroid \\ (Weight-Space)}}
        & MCDR & MC Dropout & Weights & No & MC Sampling (Dropout) \\
        & SWAGR & SWAG & Weights & No & MC Sampling (Weights) \\
        & DER & Deep Ensembling & Weights & No & MC Sampling (Ensemble) \\
        \midrule
        \multirow{2}{*}{\makecell[l]{Expert Logit \\ (Latent-Space)}}
        & MFVR & Variational Inference & Logits & Yes & Reparameterised MC Sampling (Logits) \\
        & FCVR & Variational Inference & Logits & Yes & Reparameterised MC Sampling (Logits) \\
        \midrule
        \makecell[l]{Expert Selection \\ (Decision-Space)}
        & VTSR & \makecell[l]{Beyesian Decision Theory \\ (Temperature-Sampling)}  & Selection Policy & Yes & Direct Sampling (Single) \\
        \bottomrule
    \end{tabular}}
\end{table}

As summarised in Table~\ref{tab:methodology_summary}, these approaches offer a clear spectrum of trade-offs. 
The weight-space methods build upon classic, well-understood BNN techniques. 
The logit-space methods provide a more direct and expressive way to model uncertainty over the routing decision itself, at the cost of an additional inference network. 
Finally, the selection-space method presents a uniquely efficient alternative that avoids Monte Carlo averaging. 

Having established the theoretical and architectural foundations of these methods, we now turn to a rigorous empirical evaluation of their performance in the next chapter.


\chapter{Experiments and Analysis}
\label{chap:experiments}

This chapter presents the comprehensive empirical evaluation of the Bayesian routing methods developed in Chapter~\ref{chap:methodology}. 
The primary goal is to rigorously assess their performance against standard baselines across a range of critical evaluation criteria.

Our experiments are designed to test three core hypotheses:
\begin{enumerate}
    \setlength\itemsep{0.05em}
    \item \textbf{Stability Hypothesis:} Bayesian routing methods, by modeling uncertainty, will exhibit greater stability against input perturbations compared to the brittle, deterministic router.
    \item \textbf{Calibration Hypothesis:} The proposed methods will improve model calibration on in-distribution tasks without significantly harming predictive accuracy.
    \item \textbf{OoD Detection Hypothesis:} The uncertainty signals derived from Bayesian routers will be more effective for Out-of-Distribution (OoD) detection than those from the deterministic baseline.
\end{enumerate}
To investigate these hypotheses, this chapter is structured as follows. We first detail the complete experimental setup. 
We then present the results for our three main performance experiments: Routing Stability, In-Distribution Calibration, and OoD Detection. 
Following this, we provide a comparative analysis of our layer selection strategies and a rigorous efficiency analysis of the methods' computational overhead. 
Finally, we conclude with a summary of our findings.

\section{Experimental Setup}
\label{sec:exp_setup}

This section details the common components: base model, datasets, and evaluation metrics. 
These are used across all subsequent experiments to ensure a fair and rigorous comparison of our proposed methods against established baselines.

\subsection{Model, Baselines, and Proposed Methods}

\paragraph{Base Model}
All experiments are conducted using the \textbf{IBM Granite-3.1 3B Instruct} model, an open-source, 3-billion parameter, decoder-only Mixture-of-Experts model designed for instruction-following tasks~\cite{ibm_granite_3b_2024}. Our Bayesian methods are applied as fine-tuning strategies on top of the pre-trained weights of this model.

\paragraph{Baselines}
We compare our methods against two key baselines:
\begin{enumerate}
    \setlength\itemsep{0.05em} 
    \item \textbf{Deterministic Router:} The standard, unmodified Granite-3.1 router, which uses a deterministic Top-K selection mechanism. This serves as our primary baseline.
    \item \textbf{Temperature Sampling:} A non-Bayesian stochastic baseline that uses a fixed, globally-tuned temperature to scale the logits before sampling experts, as explored in Chapter~\ref{chap:motivation}.
\end{enumerate}

\paragraph{Proposed Methods}
We evaluate the six Bayesian routing methods developed in Chapter~\ref{chap:methodology}: 
the three weight-space methods (\textbf{MCDR}, \textbf{SWAGR}, \textbf{DER}),
two logit-space methods (\textbf{MFVR}, \textbf{FCVR})
and one selection-space method (\textbf{VTSR}).

\subsection{Datasets and Tasks}

All evaluations are performed on the \textbf{Multiple-Choice Question Answering (MCQA)} task across a suite of seven distinct datasets. 
These datasets test a range of reasoning skills, from commonsense knowledge to expert-level domains. 
A brief description of each is provided below, with full details on data format, preprocessing and splits available in Table~\ref{tab:mcqa_datasets_summary}, Appendix~\ref{app:models_and_datasets}.
\begin{itemize}
    \setlength\itemsep{0.05em} 
    \item \textbf{OpenBookQA (OBQA)}~\cite{mihaylov2018can}: A commonsense reasoning dataset requiring scientific knowledge from an open book of elementary-level science facts.
    \item \textbf{AI2 Reasoning Challenge (ARC)}~\cite{clark2018think}: A dataset of challenging, grade-school-level science questions. We use both the difficult \textbf{ARC-Challenge} set and the simpler \textbf{ARC-Easy} set.
    \item \textbf{SciQ}~\cite{welbl2017crowdsourcing}: A dataset containing crowdsourced science exam questions covering a broad range of topics in physics, chemistry, and biology.
    \item \textbf{MedMCQA}~\cite{pal2022medmcqa}: A large-scale medical entrance exam dataset. We use a subset of questions from the \textbf{Medicine} subject area, which requires expert clinical knowledge.
    \item \textbf{MMLU (Massive Multitask Language Understanding)}~\cite{hendrycks2020measuring}: A benchmark designed to measure knowledge across a vast range of subjects. We use the \textbf{Professional Law} subset for our experiments.
\end{itemize}

Our experiments are structured into two distinct evaluation settings:
\vspace{-.3cm}
\paragraph{In-Distribution (ID) Evaluation} For the primary calibration and performance analysis, we fine-tune and evaluate the model separately on four distinct datasets, treating each as an independent in-distribution task: \textbf{OBQA}, \textbf{ARC-Challenge}, \textbf{SciQ}, and \textbf{MedMCQA-Med}.
\vspace{-.2cm}
\paragraph{Out-of-Distribution (OoD) Evaluation} For OoD detection experiments, the model is fine-tuned solely on \textbf{OBQA}. We then test its ability to distinguish this in-domain data from two types of distributional shifts:
\vspace{-.2cm}
\begin{itemize}
    \setlength\itemsep{0.05em} 
    \item \textit{Small Shift (Formal Science):} \textbf{ARC-Challenge} and \textbf{ARC-Easy}.
    \item \textit{Large Shift (Expert Domains):-} \textbf{MedMCQA-Med} and \textbf{MMLU-Law}.
\end{itemize}

\subsection{Evaluation Metrics}
To test our hypotheses, we employ a suite of metrics to measure model stability, calibration, and OoD detection performance.
\vspace{-.2cm}
\begin{itemize}
    \setlength\itemsep{0.05em} 
    \item \textbf{Routing Stability:} Measured using the \textbf{Jaccard Similarity} between the expert sets selected for an original input and its perturbed version.
    \item \textbf{Performance and Calibration:} Measured using standard classification and calibration metrics:
    \vspace{-.5cm}
    \begin{itemize}
        \setlength\itemsep{0.05em} 
        \item \textbf{Accuracy:} The proportion of correct answers.
        \item \textbf{Negative Log-Likelihood (NLL):} Measures the quality of the predicted probabilities.
        \item \textbf{Expected Calibration Error (ECE):} The primary metric for miscalibration, measuring the difference between confidence and accuracy.
        \item \textbf{Maximum Calibration Error (MCE):} Measures the worst-case calibration error in any confidence bin.
    \end{itemize}
    \item \textbf{Out-of-Distribution Detection:} Measured by treating the task as a binary classification problem (ID vs. OoD) based on an uncertainty score. We report:
    \vspace{-.2cm}
    \begin{itemize}
        \setlength\itemsep{0.05em} 
        \item \textbf{AUROC:} The Area Under the Receiver Operating Characteristic curve.
        \item \textbf{AUPRC:} The Area Under the Precision-Recall curve.
    \end{itemize}
\end{itemize}

\section{Implementation Details and Training Strategy}
\label{sec:exp_implementation}

This section details the specific choices made during the implementation of our experiments, including the entire training procedure to guarantee fair comparison, which layers were modified, and the key tuning considerations required for each of the proposed Bayesian methods.

\subsection{Training Pipeline}
To create a strong deterministic baseline and ensure a fair comparison, we employ a multi-stage fine-tuning process.

\vspace{-.2cm}
\paragraph{Deterministic Router Fine-Tuning (MAP Baseline)}
Our process begins by adapting the pre-trained Granite-3.1 model to our in-distribution MCQA task. This is done in two stages:
\begin{enumerate}
    \setlength\itemsep{0.05em} 
    \item First, we perform an efficient \textbf{LoRA (Low-Rank Adaptation)}~\cite{hu2021lora} fine-tuning of the attention layers' Key, Value, and Query (KVQ) projection matrices. This adapts the model's core representations to the task domain.
    \item Second, with the adapted attention layers frozen, we conduct a full-parameter fine-tuning of all MoE router linear layers. This yields our strong, deterministic baseline router with \textbf{Maximum a Posteriori (MAP)} weights.
\end{enumerate}

\vspace{-.2cm}
\paragraph{Bayesian Router Fine-Tuning}
All of our proposed Bayesian methods are then trained as a final fine-tuning step. Each Bayesian router is initialised with the weights from the converged MAP baseline and then trained further according to its specific objective (e.g., with dropout active, using the ELBO loss, etc.). This ensures that any observed improvements are due to the Bayesian treatment itself, rather than differences in initialisation or general training.

\subsection{MoE Layer Selection Strategies}
A key research question when modifying a deep architecture like an MoE-LLM is not just \textit{how} to intervene, but \textit{where}. To investigate this, we evaluate three distinct strategies for choosing which MoE router layers to make Bayesian:
\begin{enumerate}
    \setlength\itemsep{0.05em} 
    \item \textbf{Susceptible Layers (Primary Strategy):} Our main approach is to apply the Bayesian treatment only to the layers identified as most brittle in our motivational stability analysis (Chapter~\ref{chap:motivation}). This tests the hypothesis that a targeted intervention is most effective. All main results in this chapter are reported using this strategy.
    \item \textbf{Last Layer (Heuristic):} A simple heuristic where only the final MoE layer in the network is made Bayesian. This targets the layer responsible for the highest level of semantic abstraction.
    \item \textbf{Last-5 Layers (Heuristic):} A more general heuristic that applies the Bayesian modification to a block of the final five MoE layers, without relying on a prior stability analysis.
\end{enumerate}
A comparative analysis of these three strategies is presented in Section~\ref{sec:exp_layer_comparison} to validate our primary approach.

\subsection{Method-Specific Tuning and Considerations}
Each of our proposed Bayesian methods has unique hyperparameters that require careful tuning to ensure both stability and optimal performance.

\vspace{-.2cm}
\paragraph{MC Dropout Router (MCDR)} 
The most critical hyperparameter for MCDR is \textbf{dropout rate, $p$}. 
After experimentation, a rate of $p=0.05$ was selected. A MC sample number of $S=35$ was used.

\vspace{-.2cm}
\paragraph{Deep Ensembles of Routers (DER)} 
For DER, key parameter is \textbf{number of ensemble members, $M$}. 
While a larger ensemble yields better performance, this comes at a linear cost in both computation and memory. 
For computational feasibility, our experiments were conducted with $M=10$.

\vspace{-.2cm}
\paragraph{Variational Routers (MFVR \& FCVR)} The crucial hyperparameter for the variational routers is the \textbf{KL-divergence weight, $\beta$}, in the ELBO loss function. This term balances the task-specific reconstruction loss against the regularisation of the latent logit space. Careful tuning is required to prevent \textbf{posterior collapse}.

\vspace{-.2cm}
\paragraph{Variational Temperature Router (VTSR)} Similarly, the VTSR has a regularisation weight, $\beta$, for its $\mathbb{E}[\log(T(\mathbf{x}))]$ term. This is essential for preventing the learned temperature from collapsing towards zero, which would revert the model to a deterministic state.\\

All code to reproduce our experiments, including the specific hyperparameter configurations for each method, 
is available at our public repository\footnote{\url{https://github.com/albus-li/albus-bayesian-moe-router}}.

\section{Experiment 1: Stability Under Perturbation}

\subsection{Goal and Methodology}

The first experiment directly tests our \textbf{Stability Hypothesis}: that the proposed Bayesian routing methods are more robust to minor input perturbations than the standard deterministic router. A robust router should maintain a consistent expert selection policy when faced with semantically meaningless noise, while a brittle router will exhibit erratic changes.

To measure this, we adopt the same methodology as our motivational experiment in Chapter~\ref{chap:motivation}. We inject a small amount of calibrated Gaussian noise into the input of the target MoE router layer. We then measure the change in the set of selected experts between the original and perturbed input using the \textbf{Jaccard Similarity}. This process is repeated for all methods across a large sample of test tokens, and the mean Jaccard Similarity is reported.

\subsection{Results and Analysis}

\begin{figure}[H]
    \centering
    \includegraphics[width=0.8\textwidth]{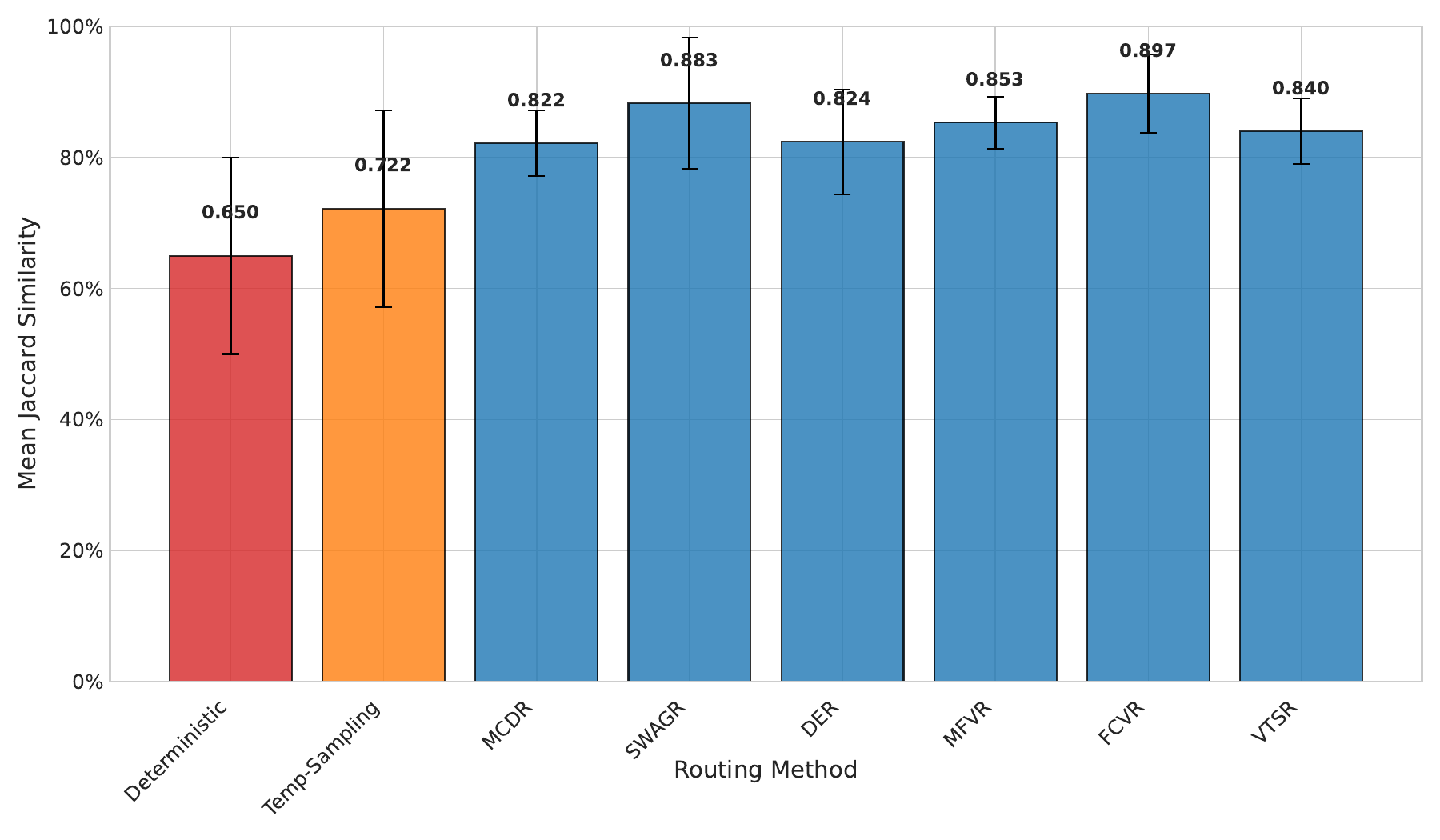}
    \caption{Mean Jaccard Similarity for each routing method under input perturbation, evaluated on the OBQA dataset. Higher scores indicate greater stability. Error bars represent the standard deviation across the test set.}
    \label{fig:stability_results}
\end{figure}

The results of the stability experiment are presented in Figure~\ref{fig:stability_results}. These scores were obtained by fine-tuning the susceptible layers of the \textbf{ibm-granite-3b} model on the \textbf{OBQA} dataset. The final Jaccard Similarity for each method is the average score across all modified layers and test tokens.

As hypothesised, the \textbf{deterministic router} exhibits the lowest stability, confirming its brittle nature with a mean Jaccard Similarity of only \textbf{0.650}. The simple \textbf{temperature sampling} baseline offers a modest improvement to \textbf{0.722}, suggesting that even ad-hoc stochasticity helps mitigate brittleness.

All proposed Bayesian methods demonstrate a substantial and statistically significant improvement in routing stability over both baselines. 
The logit-space methods proved to be particularly effective, with the \textbf{FCVR} achieving the highest stability of all methods at \textbf{0.897}, followed closely by the \textbf{MFVR} at \textbf{0.853}. 
Among the weight-space methods, \textbf{SWAGR} was a top performer with a score of \textbf{0.883}. 
The other methods, including \textbf{VTSR} (\textbf{0.840}), \textbf{DER} (\textbf{0.824}), and \textbf{MCDR} (\textbf{0.822}), also provided strong and reliable improvements.

This experiment provides compelling evidence in support of our stability hypothesis. The results quantitatively demonstrate that modelling uncertainty with a range of different Bayesian methods leads to a more robust and reliable expert selection mechanism compared to the deterministic approach.

\section{Experiment 2: In-Distribution Calibration}
\subsection{Goal and Methodology}

This experiment tests our \textbf{Calibration Hypothesis}: that the proposed Bayesian routing methods can improve model calibration on in-distribution (ID) tasks without significantly harming predictive accuracy. 
A well-calibrated model is crucial for trustworthiness, as its predictive confidence should accurately reflect its likelihood of being correct.

The evaluation is conducted on our suite of in-distribution MCQA datasets. We measure performance using standard metrics: 
\textbf{Accuracy (ACC)} for predictive performance, and \textbf{Negative Log-Likelihood (NLL)}, 
\textbf{Expected Calibration Error (ECE)}, and \textbf{Maximum Calibration Error (MCE)} to quantify calibration. 
We also use \textbf{Reliability Diagrams} for a visual assessment of calibration.

\subsection{Results and Analysis}

We tested our proposed Bayesian methods and the baselines on all four in-distribution datasets. 
The routers displayed a consistent pattern of behaviour across all settings. 
For clarity, we present the results from the \textbf{OpenBookQA (OBQA)} dataset here as a representative example. 
The full results for all four datasets are detailed in Table~\ref{tab:id_full_results}, Appendix~\ref{app:id_full_results}.

\begin{figure}[H]
    \centering
    \includegraphics[width=\textwidth]{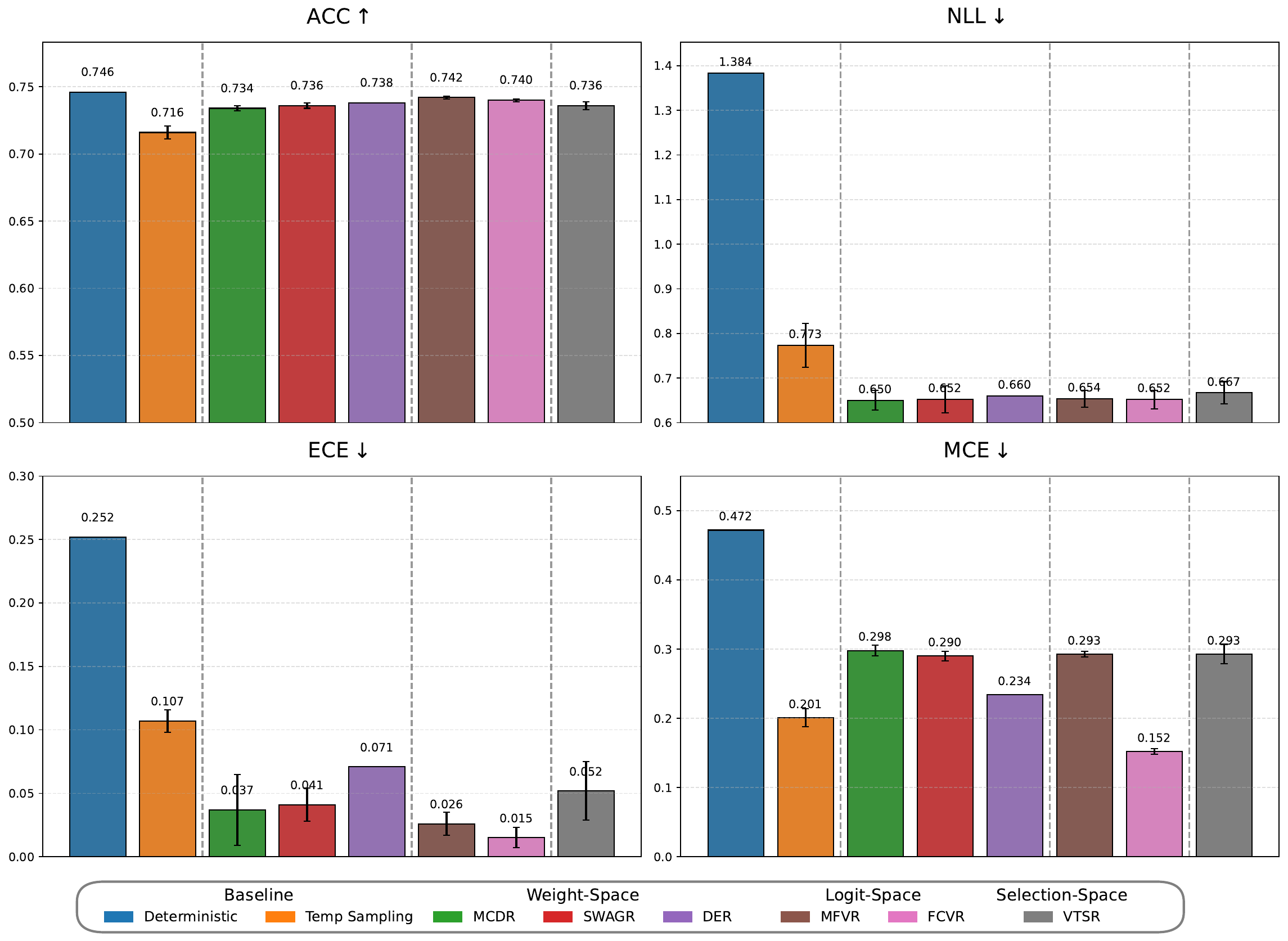}
    \caption{In-distribution performance and calibration results on the \textbf{OpenBookQA (OBQA)} dataset.}
    \label{fig:id_calibration_obqa}
\end{figure}

The primary quantitative results for OBQA are summarised in Figure~\ref{fig:id_calibration_obqa}
\footnote{
    Metrics for every method (exluding deterministic baseline and DER) are averaged over 5 stochastic forward passes.
    Standard deviations are shown as error bars.
}. 
A key finding is that all of our proposed Bayesian methods maintain \textbf{Accuracy} on par with the strong deterministic baseline.
This is a crucial distinction from the `Temp-Sampling' baseline, which improves calibration but at a notable cost to accuracy, highlighting the trade-offs of using unprincipled stochasticity.

The benefits of our approach become evident in the probabilistic and calibration metrics. 
For \textbf{Negative Log-Likelihood (NLL)}, the \textbf{MC Dropout Router} was the top performer. 
This is a particularly noteworthy result, as MCDR is simple to implement and demonstrates that an effective probabilistic model does not necessarily require a complex architecture. 
As our primary metric for miscalibration, the \textbf{Expected Calibration Error (ECE)} is substantially reduced by all Bayesian methods. 
The logit-space methods performed exceptionally well, with \textbf{FCVR} reducing the ECE by over \textbf{94\%} compared to the deterministic baseline.

Overall, this experiment provides strong evidence in support of our calibration hypothesis. The results show that by introducing principled uncertainty into the routing mechanism, we can significantly improve the calibration of MoE models without compromising their core predictive accuracy.

\section{Experiment 3: Out-of-Distribution Detection}
\label{sec:exp_ood}

\subsection{Goal and Methodology}

This experiment evaluates our \textbf{OoD Detection Hypothesis} by investigating how our proposed Bayesian routers improve the model's ability to distinguish in-domain (ID) from out-of-distribution (OoD) data. 
We designed four distinct OoD detection tasks in total: two representing a \textbf{small distributional shift} (ID: OBQA vs. OoD: ARC-C / ARC-E) 
and two representing a \textbf{large distribution shift} (ID: OBQA vs. OoD: MMLU-Law / MedMCQA). 
To ensure a clear demonstration of the main findings, we present the results for one representative large-shift task, 
\textit{ID: OBQA vs. OoD: MedMCQA-Med}, in this section. 
The complete results for all four OoD tasks can be found in Appendix~\ref{app:ood_detection}.

The evaluation is structured as two distinct sub-experiments, each testing a specific aspect of uncertainty. The task is framed as a binary classification problem where a model-derived uncertainty score is used to classify inputs, with performance measured by \textbf{AUROC} and \textbf{AUPRC}. Based on their strong performance in the in-distribution calibration experiments, we focus our analysis on four standout Bayesian methods: \textbf{MCDR} (as the most effective weight-space method), \textbf{MFVR}, \textbf{FCVR}, and \textbf{VTSR}.

\subsection{Experiment 3a: Improving Standard Uncertainty Signal}

Our first hypothesis is that the uncertainty introduced by a Bayesian router will propagate through the network, making the standard uncertainty signal—the entropy of the final prediction over the vocabulary—more reliable. 
To test this, we compare the OoD detection performance using the \textbf{final vocabulary entropy} from our standout Bayesian methods against the same signal from the deterministic baseline. 
The results, shown in Table~\ref{tab:ood_vocab_entropy}, demonstrate a clear improvement across all evaluated methods.

\begin{wraptable}{l}{0.5\textwidth}
    \centering
    \caption{OoD detection performance using the \textbf{final vocabulary entropy} on the OBQA vs. MedMCQA task. Best results are in bold.}
    \label{tab:ood_vocab_entropy}
    \begin{tabular}{lcc}
        \toprule
        \textbf{Method} & \textbf{AUROC} $\uparrow$ & \textbf{AUPRC} $\uparrow$ \\
        \midrule
        Deterministic & 0.762 & 0.727 \\
        \midrule
        MCDR & 0.793 & 0.737 \\
        MFVR & 0.844 & 0.782 \\
        FCVR & \textbf{0.853} & \textbf{0.802} \\
        VTSR & 0.812 & 0.791 \\
        \bottomrule
    \end{tabular}
\end{wraptable}

The FCVR method achieves the highest scores, but all Bayesian approaches show a significant gain in both AUROC and AUPRC over the deterministic model. This suggests that a more robust internal routing mechanism leads to a more calibrated and reliable final prediction distribution, which in turn serves as a better signal for OoD detection. 

This finding is crucial as it validates the idea that improving an internal component of the model can have a positive, measurable impact on final output's reliability. 
\\

\subsection{Experiment 3b: Router-Level Uncertainty as Signal}
Inspired by work~\cite{li2024your} showing that MoE routing probabilities can serve as meaningful representations, our second hypothesis is that the router's internal uncertainty can be leveraged as a novel and superior signal for OoD detection. 
We test if method-specific signals
\footnote{Details of each method-specific signals are provided in Appendix~\ref{app:ood_detection}.}
that directly capture the router's epistemic uncertainty (e.g., logit variance) outperform the naive entropy of the expert selection probabilities.

\begin{table}[H]
    \centering
    \caption{Comparison of different \textbf{router-level} uncertainty signals for OoD detection on the OBQA vs. MedMCQA task. The best signal for each method is in bold.}
    \label{tab:ood_router_signals}
    \begin{tabular}{llcc}
        \toprule
        \textbf{Method} & \textbf{Router-Level Signal Type} & \textbf{AUROC} $\uparrow$ & \textbf{AUPRC} $\uparrow$ \\
        \midrule
        Deterministic & Expert Selection Entropy & 0.679 & 0.645 \\
        \midrule
        \multirow{2}{*}{\textbf{MCDR}} 
        & Expert Selection Entropy & 0.684 & 0.651 \\
        & MC Logit Variance & 0.786 & 0.723 \\
        \midrule
        \multirow{2}{*}{\textbf{MFVR}}
        & Expert Selection Entropy & 0.682 & 0.637 \\
        & Inferred Logit Variance & 0.835 & \textbf{0.793} \\
        \midrule
        \multirow{2}{*}{\textbf{FCVR}} 
        & Expert Selection Entropy & 0.692 & 0.642 \\
        & Inferred Logit Variance & \textbf{0.844} & 0.773 \\
        \midrule
        \multirow{2}{*}{\textbf{VTSR}} 
        & Expert Selection Entropy & 0.683 & 0.643 \\
        & Inferred Temperature & 0.512 & 0.492 \\
        \bottomrule
    \end{tabular}
\end{table}

This detailed analysis reveals several key insights. 
A surprising finding is that \textbf{expert selection entropy}, when used as an uncertainty signal, shows only marginal improvements for Bayesian methods compared to deterministic baseline. 
This suggests that simply making the routing process probabilistic is not, by itself, sufficient to create a powerful OoD signal at the post-softmax level.

The true benefit of our framework is revealed when we examine the \textbf{method-specific uncertainty signals}. 
For every method that provides such a signal, it \textbf{consistently and significantly outperforms} the naive expert selection entropy. 
As shown in Table~\ref{tab:ood_router_signals}, the `Logit Variance' for MCDR, MFVR and FCVR are demonstrably better OoD detectors. 
This confirms our core hypothesis: the internal, pre-softmax uncertainty about the logits provides a richer and more reliable measure of the model's confidence than the entropy of the final probabilities.

Furthermore, the poor performance of the `Inferred Temperature' from the VTSR provides a crucial diagnostic insight. 
The model's failure to produce a high temperature for OoD inputs indicates that the training objective is dominated by the task loss, causing the regularisation term to be ignored. This is a classic symptom of \textbf{posterior collapse}, where the model learns to make its uncertainty signal uninformative (i.e., always predicting a low temperature) to achieve a lower overall loss. This highlights the challenges in training such a direct signal and reinforces the effectiveness of the more implicit uncertainty captured by the logit-space and weight-space methods.

\section{Ablation Study: \\ Comparative Analysis of Layer Selection}
\label{sec:exp_layer_comparison}

The main results presented in the preceding sections were generated using our primary \textbf{Susceptible Layers} strategy. This section provides a detailed ablation study to validate that methodological choice. For each of our standout Bayesian methods (\textbf{MCDR, MFVR, FCVR, and VTSR}), we compare its performance when applied using three different layer selection strategies:
\begin{enumerate}
    \setlength\itemsep{0.05em} 
    \item \textbf{Susceptible Layers (Primary):} Targeted approach based on stability analysis in Chapter~\ref{chap:motivation}.
    \item \textbf{Last Layer Only (Heuristic):} A simple heuristic targeting only the final MoE layer.
    \item \textbf{Last-5 Layers (Heuristic):} A more general heuristic targeting a block of final five MoE layers.
\end{enumerate}
We evaluate these strategies using the single key metric from each of our three main experiments, with results averaged across all relevant datasets.

The results of this comparison are summarised in Table~\ref{tab:layer_selection_comparison}. The findings show a clear and consistent trend across all evaluated methods: the targeted \textbf{Susceptible Layers} strategy almost always yields the best performance. For nearly every method, this strategy achieves the highest mean Jaccard Similarity, the lowest mean ECE, and the highest mean AUROC.

While the ``Last-5 Layers'' heuristic provides a reasonable improvement, it rarely matches the performance of the more targeted approach. 
The ``Last Layer Only'' strategy is clearly suboptimal, suggesting that intervening at a single, final layer is insufficient to address the model's systemic brittleness. 
These findings validate our primary methodological choice, demonstrating that a targeted application of Bayesian methods to the layers most prone to instability is more effective than using simpler heuristics.

\begin{table}[H]
    \centering
    \caption{Comparative analysis of layer selection strategies for each standout Bayesian method. The AUROC metric is calculated using the final vocabulary entropy. Best result for each method is in bold.}
    \label{tab:layer_selection_comparison}
    \begin{tabular}{l l S[table-format=1.3, detect-weight] 
                   S[table-format=1.3, detect-weight] 
                   S[table-format=1.3, detect-weight]}
        \toprule
        \textbf{Method} & \textbf{Layer Selection Strategy} & {\textbf{Jaccard} $\uparrow$} & {\textbf{ECE} $\downarrow$} & {\textbf{AUROC (Voc. Ent.)} $\uparrow$} \\
        \midrule
        \multirow{3}{*}{\textbf{MCDR}} 
        & Susceptible layers & \textbf{0.822} & \textbf{0.037} & \textbf{0.793} \\
        & Last 5 Layers      & 0.793 & 0.113 & 0.773 \\
        & Last Layer Only    & 0.752 & 0.135 & 0.762 \\
        \midrule
        \multirow{3}{*}{\textbf{MFVR}} 
        & Susceptible layers & \textbf{0.853} & \textbf{0.026} & \textbf{0.844} \\
        & Last 5 Layers      & 0.821 & 0.121 & 0.808 \\
        & Last Layer Only    & 0.779 & 0.205 & 0.778 \\
        \midrule
        \multirow{3}{*}{\textbf{FCVR}} 
        & Susceptible layers & \textbf{0.897} & \textbf{0.015} & \textbf{0.853} \\
        & Last 5 Layers      & 0.872 & 0.103 & 0.811 \\
        & Last Layer Only    & 0.783 & 0.194 & 0.783 \\
        \midrule
        \multirow{3}{*}{\textbf{VTSR}} 
        & Susceptible layers & \textbf{0.840} & \textbf{0.052} & \textbf{0.812} \\
        & Last 5 Layers      & 0.832 & 0.142 & 0.789 \\
        & Last Layer Only    & 0.732 & 0.168 & 0.773 \\
        \bottomrule
    \end{tabular}
\end{table}

\section{Practicality: \\ Efficiency Analysis of Bayesian Routers}
\label{sec:conclusion_efficiency}

This section will provide a rigorous quantitative discussion of the memory and computational costs of the proposed Bayesian routing methods. 
To be considered practical, the overhead of these methods must be negligible relative to the scale of the base model. 
This analysis will show that this is indeed the case.
\begin{itemize}
    \setlength\itemsep{0.05em} 
    \item $L$: MoE (Mixture of Experts) layer number
    \item $N$: Number of experts
    \item $D$: Model hidden dimension
    \item $S$: Number of Monte Carlo samples
    \item $M$: Number of ensemble members
    \item $H$: Hidden dimension within additional networks 
        ($\text{NN}_\mu$, $\text{NN}_\sigma$ in MFVR/FCVR, $\text{NN}_\text{temp}$ in VTSR)
    \item $B$: Batch size
    \item $T$: Sequence length
\end{itemize}

\subsection{Memory Overhead}

To assess the practicality of our methods, we first analyse their memory footprint. 
In the context of large-scale MoE models, the most critical metric is not the on-disk storage size, 
but the \textbf{activation memory}, the total number of parameters that must be actively held in GPU memory to perform an inference pass~\cite{shazeer2017outrageously},
which is the principle we will adopt for our analysis
\footnote{For some sample-based methods, number of activated parameters during inference can exceed that of stored parameters.}
.

\paragraph{Weight-Space Methods}
The inference-time memory cost for weight-space methods is driven by the need to generate multiple samples of the router weights.
\begin{itemize}
    \setlength\itemsep{0.05em} 
    \item \textbf{MCDR} is exceptionally efficient. As dropout is implemented as a mask on the input activations, it requires \textbf{zero} additional weight parameters to be loaded into memory.
    \item \textbf{SWAGR} requires loading $S$ samples of the expert centroid matrix, $W_{\text{EC}}$, for parallel processing. The total additional activation memory for $L$ modified layers is therefore $L \times (S-1) \times D \times N$.
    \item \textbf{DER} also requires loading all $M$ ensemble members, resulting in an additional memory cost of $L \times (M-1) \times D \times N$.
\end{itemize}

\paragraph{Logit and Selection-Space Methods}
For these methods, the primary memory overhead is the fixed cost of the additional inference network's parameters, which must be loaded into memory.
\begin{itemize}
    \setlength\itemsep{0.05em} 
    \item \textbf{MFVR} requires a one-hidden-layer MLP with a hidden dimension $H$ and two output heads of size $N$, for a total of $L \times (D \cdot H + 2 \cdot H \cdot N)$ additional parameters.
    \item \textbf{FCVR} is similar, but one output head must parameterise the Cholesky factor, which has $\frac{N(N+1)}{2}$ elements. The cost is $L \times (D \cdot H + H \cdot N + H \cdot \frac{N(N+1)}{2})$.
    \item \textbf{VTSR} requires only a small network to predict a scalar, for a cost of $L \times (D \cdot H + H \cdot 1)$ parameters.
\end{itemize}

Table~\ref{tab:memory_overhead_quant} quantifies these theoretical costs within the context of the \textbf{Granite-3B-MoE}
\footnote{$D=1536$, $N=40$, $L_\text{total}=32$} 
model, assuming the modification of $L=10$ layers and hyperparameters of $S=35$, $M=10$ and $H= \frac{D}{4}$.

\begin{table}[H]
    \centering
    \caption{Theoretical activation memory overhead for each Bayesian router, quantified for the Granite-3B MoE model and shown as a percentage of the total $\sim$800M activated parameters during inference.}
    \label{tab:memory_overhead_quant}
    \begin{tabular}{l l r r}
        \toprule
        \textbf{Method} & \textbf{Theoretical Formula} & \textbf{Actual Add. Params} & \textbf{\% of Total Model} \\
        \midrule
        MCDR & 0 & 0 & 0.00\% \\
        SWAGR & $L(S-1)DN$ & $\sim$20.9M & $\sim$2.61\% \\
        DER & $L(M-1)DN$ & $\sim$5.5M & $\sim$0.69\% \\
        \midrule
        MFVR & $L(DH + 2HN)$ & $\sim$6.2M & $\sim$0.78\% \\
        FCVR & $L(DH + HN + H\frac{N(N+1)}{2})$ & $\sim$9.2M & $\sim$1.15\% \\
        \midrule
        VTSR & $L(DH + H)$ & $\sim$5.9M & $\sim$0.74\% \\
        \bottomrule
    \end{tabular}
\end{table}

\subsection{Computation Overhead}

Next, we analyse the computational cost of each method in terms of floating-point operations (FLOPs). 
The primary source of computational cost in our networks is matrix multiplication. 
The FLOPs required to multiply a $p \times r$ matrix with an $r \times q$ matrix is approximately $2prq$. 
Therefore, a single forward pass for one token through a router's linear layer ($W_{EC} \in \mathbb{R}^{D \times N}$) requires approximately $2DN$ FLOPs. 
In our analysis, we consider costs of activation functions negligible.

\paragraph{Weight-Space Methods}
The overhead for these methods comes from the need to perform multiple forward passes through the router to generate samples.
\begin{itemize}
    \setlength\itemsep{0.05em} 
    \item \textbf{MCDR and SWAGR:} Both require $S$ forward passes. The additional cost over the single baseline pass is $L \times (S-1) \times 2DN$ FLOPs.
    \item \textbf{DER:} It requires $M$ forward passes, for an additional cost of $L \times (M-1) \times 2DN$ FLOPs.
\end{itemize}

\paragraph{Logit-Space Methods}
These methods incur overhead from both their additional inference network and the sampling process.
\begin{itemize}
    \setlength\itemsep{0.05em} 
    \item \textbf{MFVR:} Double-head one-hidden-layer MLP adds approximately $2DH + 4HN$ FLOPs. 
          Reparameterisation trick for $S$ samples adds $S \times 2N$ FLOPs. 
          Total overhead is the sum of these two.
    \item \textbf{FCVR:} MLP cost is higher due to the larger Cholesky factor output head, costing roughly $2DH + 2HN + 2H\frac{N(N+1)}{2}$ FLOPs. 
          The reparameterisation requires a matrix-vector product, adding $S \times 2N^2$ FLOPs.
\end{itemize}

\paragraph{Selection-Space Method}
\begin{itemize}
    \item \textbf{VTSR:} The temperature prediction network adds approximately $2DH + 2H$ FLOPs. This is followed by $N$ divisions to scale the logits
    \footnote{Our theoretical FLOPs analysis does not include the cost of averaging multiple post-softmax outputs. 
        If this is considered from a theoretical analysis standpoint, VTSR would be even more efficient, as it does not require sampling.}.
\end{itemize}

Table~\ref{tab:flops_overhead_quant} summarises the theoretical overhead of each method and contextualises it as a percentage of the total FLOPs
\footnote{Actual Additional FLOPs are measured and calcuated via \texttt{fvcore} python library.}
required for a full forward pass of the Granite-3B-MoE model.

\begin{table}[H]
    \centering
    \caption{Theoretical and experimental computational overhead of Bayesian routers.}
    \label{tab:flops_overhead_quant}
    \resizebox{\textwidth}{!}{
    \begin{tabular}{lccc}
        \toprule
        \textbf{Method} & \textbf{Theoretical FLOPs Overhead (Big-O)} & \textbf{Actual Add. FLOPs (GFLOPs Per Token)} & \textbf{\% of Total Model} \\
        \midrule
        MCDR & $O(LSDN)$ & 0.0208 & 2.32\% \\
        SWAGR & $O(LSDN)$ & 0.0208 & 2.32\% \\
        DER & $O(LMDN)$ & 0.0059 & 0.66\% \\
        \midrule
        MFVR & $O(L(DH + HN + SN))$ & 0.0069 & 0.77\% \\
        FCVR & $O(L(DH + HN^2 + SN^2))$ & 0.0096 & 1.07\% \\
        \midrule
        VTSR & $O(L(DH + H + N))$ & 0.0060 & 0.67\% \\
        \bottomrule
    \end{tabular}
    }
\end{table}

\subsection{Parallelisation and Practical Trade-offs}

The theoretical FLOPs translate to real-world latency based on how well the computation can be parallelised on a GPU. 
The $S$ sampling steps required for most of our methods are \textbf{embarrassingly parallelisable}~\cite{li2022branch}.

\begin{itemize}
    \setlength\itemsep{0.05em} 
    \item \textbf{MCDR:} Highly efficient; the input batch can be expanded by a factor of $S$ and processed in a single pass with different dropout masks.
    \item \textbf{DER and SWAGR:} The $S$ forward passes use different weight matrices, which is less efficient but still parallelisable.
    \item \textbf{MFVR and FCVR:} Monte Carlo sampling occurs after the parameters of the logit distribution ($\boldsymbol{\mu}, \boldsymbol{\Sigma}$) have been computed. This is very efficient, as only the small reparameterisation step needs to be parallelised, involving vector-scalar operations for MFVR and more expensive matrix-vector operations for FCVR.
    \item \textbf{VTSR:} The exception, as its single-pass inference requires no parallel sampling strategy, making its latency profile fundamentally different and more efficient.
\end{itemize}

This analysis culminates in the qualitative summary of trade-offs presented in Table~\ref{tab:tradeoff_summary}. 
The \textbf{FCVR} offers state-of-the-art performance at a moderate computational cost. 
\textbf{MCDR} provides a solid baseline improvement for almost no implementation overhead. 
While \textbf{VTSR} offers a uniquely compelling low-latency profile, its performance was hampered by training instability and temperature collapse in our experiments. 
Despite these current limitations, we believe the underlying concept of learning a direct, input-dependent routing stochasticity is powerful. 
It remains a fascinating and promising area for future work, focussed on the development of more stable training methods.

\begin{table}[H]
    \centering
    \caption{A qualitative summary of the trade-offs between performance and practicality for all evaluated methods.}
    \label{tab:tradeoff_summary}
    \begin{tabular}{lcccc}
        \toprule
        \textbf{Method} & \textbf{Calibration} $\uparrow$ & \textbf{OoD Detection} $\uparrow$& \textbf{Memory Overhead} $\downarrow$& \textbf{FLOPs Overhead} $\downarrow$\\
        \midrule
        \rowcolor{gray!20} MCDR & High & Medium & Negligible & High \\
        SWAGR & Medium & Medium & High & High \\
        DER & Medium & Medium & Low & Low \\
        \midrule
        MFVR & High & High & Low & Low \\
        \rowcolor{gray!20} FCVR & Very High & High & Medium & Medium \\
        \midrule
        \rowcolor{gray!20} VTSR & High & Low & Low & Low \\
        \bottomrule
    \end{tabular}
\end{table}

\section{Chapter Summary}
\label{sec:exp_summary}

This chapter presented a comprehensive empirical evaluation of our proposed Bayesian routing methods, assessing their performance on routing stability, model calibration, and out-of-distribution detection, as well as their practical efficiency.

The results from our experiments provide strong, consistent evidence in support of our core hypotheses. We demonstrated that all proposed Bayesian methods significantly improve \textbf{routing stability} and lead to substantial gains in \textbf{ID calibration} without harming predictive accuracy. Furthermore, we showed that the internal uncertainty signals derived from the Bayesian routers are highly effective for \textbf{OoD detection}, decisively outperforming the standard baselines.

This performance, however, must be weighed against practical costs. Our efficiency analysis revealed a clear spectrum of trade-offs. The logit-space approaches, particularly the \textbf{FCVR}, consistently provided the strongest performance but at a moderate computational cost. In contrast, the \textbf{MCDR} offered a solid improvement for a negligible implementation overhead, while the \textbf{VTSR} proved to be exceptionally efficient from a latency perspective. Our ablation study on layer selection further validated our targeted approach, showing that applying these methods to the layers most prone to instability yields the best results.

Taken together, these findings demonstrate that introducing principled Bayesian uncertainty into the MoE routing mechanism is a viable, effective, and computationally tractable strategy for building more reliable, calibrated, and robust Large Language Models.
\chapter{Discussion and Conclusion}
\label{chap:conclusion}

This thesis has presented a comprehensive empirical evaluation of a novel Bayesian routing framework designed to improve the reliability of Mixture-of-Experts (MoE) models. The experiments conducted in Chapter~\ref{chap:experiments} provide strong evidence in support of our core hypotheses.

Our results first demonstrated that the standard deterministic router is inherently brittle, whereas all proposed Bayesian methods significantly improve \textbf{routing stability} under input perturbation. On in-distribution tasks, these methods achieve substantial gains in \textbf{model calibration}, as measured by ECE and MCE, without sacrificing predictive accuracy. Furthermore, the uncertainty signals derived directly from the Bayesian routers proved to be highly effective for \textbf{Out-of-Distribution (OoD) detection}, decisively outperforming both the final-layer entropy and the internal signal from the deterministic baseline. Finally, our comparative analysis validated our targeted approach, showing that applying these methods to the layers most susceptible to instability yields the best overall performance.

These collective findings confirm that introducing principled uncertainty into the MoE routing mechanism is an effective strategy for enhancing model reliability, providing a strong foundation for the subsequent discussion on the practical trade-offs and broader implications of this work.

\section{Limitations and Future works}

While the results presented in this thesis provide strong evidence for the benefits of Bayesian routing, the scope of this work has several limitations. These limitations, however, naturally define promising and critical directions for future research.

\paragraph{Generalisability Across Models and Tasks}
Our empirical evaluation was conducted on a single base model, the Granite-3B-MoE, and was focused primarily on Multiple-Choice Question Answering tasks. 
While this provided a controlled environment for rigorous analysis, it limits the generalisability of our findings.
A crucial finding is that not all MoE architectures demonstrate a significant layer-wise susceptibility difference, as seen in the Granite-3B-MoE.
If so, optimal susceptible layer selection strategy might not be as obvious.
A crucial next step is to validate these methods across a broader range of MoE architectures, 
such as those from the DeepSeek-MoE~\cite{deepseek-v3} and Qwen-MoE~\cite{qwen2.5-moe} families, and on more diverse downstream tasks. 
This would be essential to confirm that improved routing reliability translates to performance gains across the wider LLM ecosystem.

\paragraph{Modelling Correlations in Weight-Space}
All the weight-space methods evaluated implicitly assume independence among all model weight scalars, 
which subsequently assume independence between the posteriors of the expert centroid vectors.
However, it is highly plausible that expert centroids are correlated: 
for instance, experts representing similar knowledge domains might occupy nearby or related regions in the embedding space. 
Future work could explore more structured Bayesian priors that explicitly model these correlations.

\paragraph{Stabilising the Variational Temperature Router}
Our experiments with the Variational Temperature Sampling Router (VTSR) highlighted a trade-off between theoretical elegance and practical stability. 
Its single-pass inference makes it exceptionally efficient, but its training proved challenging, often suffering from temperature collapse despite regularisation. 
This suggests that while the core concept of learning a direct, input-dependent stochasticity is powerful, it requires further research. 
Future work could focus on developing more advanced regularisation techniques or alternative training objectives to stabilise the learning of the temperature parameter.

\paragraph{Evaluation on Free-Form Generation}
The evaluation in this thesis was intentionally constrained to the MCQA setting to allow for rigorous and quantitative measurement of calibration. 
However, this does not capture the full range of LLM failure modes, particularly in open-ended, free-form generation. 
A critical direction for future work is to extend this evaluation to generative tasks. 
This would involve assessing the impact of Bayesian routers on reducing \textbf{hallucination}, 
improving the coherence of generated text under uncertainty, 
and leveraging the router's uncertainty signal to trigger safer behaviours, such as refusing to answer when the model ``knows it doesn't know".

\section{Conclusion}

The standard deterministic router in Mixture-of-Experts (MoE) models represents a critical vulnerability, where brittle, overconfident expert selections can undermine the reliability of the entire system. This thesis addressed this challenge by proposing and evaluating a structured \textbf{Bayesian routing framework}, demonstrating that a targeted application of principled uncertainty to the lightweight routing mechanism is a pragmatic and effective strategy for improving the trustworthiness of massive-scale LLMs.

Our empirical findings confirm the success of this approach. We systematically evaluated methods that introduce uncertainty at three distinct stages of the routing pipeline: 
in the \textbf{Weight-Space}, the \textbf{Logit-Space}, and the \textbf{Selection-Space}. 
The results showed that methods across all three categories successfully 
enhanced routing stability, improved model calibration, and provided a superior signal for out-of-distribution detection. 
The analysis also revealed a clear spectrum of trade-offs: the \textbf{Full-Covariance Variational Router (FCVR)} delivered state-of-the-art performance, 
while methods like \textbf{MC Dropout Router(MCDR)} offered significant gains for minimal effort, 
and the \textbf{Variational Temperature Router (VTSR)} introduced a promising, highly efficient new direction.

Ultimately, this work provides a practical, architectural pathway toward building more reliable and self-aware language models. Equipping our models with the ability to quantify their own uncertainty is not a peripheral feature but a foundational requirement for their safe and responsible deployment. The \textbf{Bayesian Mixture of Experts} framework developed in this thesis represents a significant and tangible step towards ``\textit{\textbf{making LLMs know what they don't know}}''.

\newpage
\addcontentsline{toc}{chapter}{Bibliography}
\bibliography{references}

\fancyhead{}
\chapter*{Declarations}
\addcontentsline{toc}{chapter}{Declarations}

\paragraph{Use of Generative AI}
In the preparation of this thesis, the author utilised the Generative AI model Gemini, developed by Google, as a writing and research assistant. The model's assistance was primarily in the following areas:
\begin{itemize}
    \setlength\itemsep{0.05em}
    \item Early drafting based on detailed outlines and specific instructions provided by author.
    \item Proofreading for grammatical errors, typos, and clarity.
    \item Brainstorming and suggesting alternative structures for chapters, sections, and paragraphs to improve narrative flow.
    \item Generating illustrative code snippets, including LaTeX for tables, Python for visualisations, and TikZ for diagrams.
\end{itemize}
The conceptual framework, methodological and experimental design, analysis, scientific claims, and final conclusions are entirely the author's own.

\paragraph{Data and Code Availability}
To ensure the reproducibility of this research, all source code and experimental configurations have been made publicly available. This includes the implementation of the Bayesian routing methods, training scripts, and scripts for generating most figures presented in this thesis. 
The repository can be accessed at:
\begin{center}
    \url{https://github.com/albus-li/albus-bayesian-moe-router}
\end{center}

\paragraph{Ethical Considerations and Computational Resources}
All experiments were conducted on established, publicly available academic datasets, and no new private or sensitive user data was collected.
The computational experiments were performed on the Imperial College Department of Computing (DoC) GPU Cluster, utilising \textbf{NVIDIA Tesla A100 (80GB)} and \textbf{Tesla A40 (48GB)} GPUs. The author gratefully acknowledges the provision of these essential computational resources.
\appendix

\clearpage 
\chapter{Models \& Datasets}
\label{app:models_and_datasets}

This appendix provides detailed information on:
\begin{itemize}
    \item MCQA datasets used in this thesis (see Table~\ref{tab:mcqa_datasets_summary})
    \item Open-sourced state-of-the-art MoE-based LLMs' configurations (see Table~\ref{tab:moe_models_full_comparison})
    \footnote{Not all models listed are used in this thesis. In fact, we only use the IBM Granite MoE models for experiments. The full list is provided for completeness and future reference.}
\end{itemize}

\begin{landscape}
\footnotesize 
\begin{longtable}{@{} l p{4.0cm} p{12cm} l @{}} 
    \caption{Summary of Selected MCQA Datasets for Calibration and OoD Experiments}
    \label{tab:mcqa_datasets_summary}\\
    \toprule
    \textbf{Dataset} & \textbf{Knowledge Domain} & \textbf{Example Question, Choices, \& Answer} & \textbf{Split Sizes (Train/Val/Test)} \\
    \midrule
    \endfirsthead

    \multicolumn{4}{c}%
    {{\bfseries \tablename\ \thetable{} -- continued from previous page}} \\
    \toprule
    \textbf{Dataset} & \textbf{Knowledge Domain} & \textbf{Example Question, Choices, \& Answer} & \textbf{Split Sizes (Train/Val/Test)} \\
    \midrule
    \endhead

    \bottomrule
    \endfoot

    \endlastfoot

    \textbf{OBQA} & Commonsense Science Reasoning & 
        \textbf{Q:} A person wants to start saving money... After looking over their budget... they decide the best way to save money is to\dots \newline
        \textbf{C:} (A) make more phone calls; (B) quit eating lunch out; (C) buy less with monopoly money; (D) have lunch with friends \newline
        \textbf{A:} quit eating lunch out 
        & \makecell[tl]{
        \textbf{Original:} 4957 / 500 / 500 \\
        \textbf{ID:} 5000 / 50 / 500} \\
    \midrule

    \textbf{ARC-C} & Formal Science Education (Challenge) & 
        \textbf{Q:} An astronomer observes that a planet rotates faster after a meteorite impact. Which is the most likely effect of this increase in rotation? \newline
        \textbf{C:} (A) Planetary density will decrease.; (B) Planetary years will become longer.; (C) Planetary days will become shorter.; (D) Planetary gravity will become stronger. \newline
        \textbf{A:} Planetary days will become shorter. 
        & 
        \makecell[tl]{\textbf{Original:} 1119 / 299 / 1172 
        \\ \textbf{OoD-S:} 500 from 1172 
        } \\
    \addlinespace
    \textbf{ARC-E} & Formal Science Education (Easy) & 
        \textbf{Q:} Which statement best explains why photosynthesis is foundation of food webs? \newline
        \textbf{C:} (A) Sunlight is the source of energy for nearly all ecosystems.; (B) Most ecosystems are found on land instead of in water.; (C) Carbon dioxide is more available than other gases.; (D) The producers in all ecosystems are plants. \newline
        \textbf{A:} Sunlight is the source of energy for nearly all ecosystems. 
        & \makecell[tl]{
        \textbf{Original:} 2251 / 570 / 2376 
        \\ \textbf{OoD-S:} 500 from 2376 
        }\\
    \midrule

    \textbf{SciQ} & Broad STEM Knowledge & 
        \textbf{Q:} Compounds that are capable of Accuracyepting electrons, such as O2 or F2, are called what? \newline
        \textbf{C:} antioxidants; Oxygen; residues; oxidants \newline
        \textbf{A:} oxidants 
        & \makecell[tl]{
            \textbf{Original:} 11679 / 1000 / 1000 \\
            \textbf{ID:} 5000 / 50 / 500
        }\\
    \addlinespace
    \textbf{MMLU-Law} & Expert Legal Reasoning & 
        \textbf{Q:} One afternoon, a pilot was flying a small airplane when it suddenly ran out of gas... At trial, the pilot's attorney calls the consulting attorney to testify... The attorney's testimony is\dots \newline
        \textbf{C:} (A) admissible, because...; (B) admissible, because...; (C) inadmissible, because the attorney-client privilege prevents...; (D) inadmissible, because it was a statement... \newline
        \textbf{A:} inadmissible, because the attorney-client privilege prevents such a breach of confidential communications. 
        &\makecell[tl]{
            \textbf{Original:} 5 (dev) / 170 / 1534 \\
            \textbf{OoD-L:} 500 from 1534
        }\\
    \addlinespace
    \textbf{MedMCQA-Med} & Expert Medical Knowledge & 
        \textbf{Q:} Which of the following is derived from fibroblast cells? \newline
        \textbf{C:} (A) TGF-13; (B) MMP2; (C) Collagen; (D) Angiopoietin \newline
        \textbf{A:} Collagen 
        & \makecell[tl]{
        \textbf{Original:} 17887 / 295 / -- \\
        \textbf{ID:} 5000 / 50 / 500 \\
        \textbf{OoD-L:} 500
        } \\

    \bottomrule

    \end{longtable}

\vfill

\normalsize
\begin{table}[h]
    \centering
    \caption{Parameters and configurations of most famous modern open-source MoE-based LLMs.}
    \label{tab:moe_models_full_comparison}
    \resizebox{1.5\textwidth}{!}{
    \begin{tabular}{llcccccc}
        \toprule
        \textbf{Family} & \textbf{Model} & \textbf{\#Act. Exp.} & \textbf{\#Total Exp.} & \textbf{Act. Params} & \textbf{Total Params} & \textbf{\#Layers} & \textbf{Hid. Dim} \\
        \midrule
        \multirow{3}{*}{MoLM} 
        & ibm-research/MoLM-350M-4B & 2 & 32 & 350M & 4B & 24 & 1024 \\
        & ibm-research/MoLM-700M-4B & 4 & 32 & 700M & 4B & 24 & 1024 \\
        & ibm-research/MoLM-700M-8B & 2 & 32 & 700M & 8B & 48 & 1024 \\
        \midrule
        \multirow{2}{*}{OLMoE} 
        & allenai/OLMoE-1B-7B-0924-Instruct & 8 & 64 & 1B & 7B & 16 & 2048 \\
        & (with SFT \& DPO) & & & & & & \\
        \midrule
        \multirow{2}{*}{IBM Granite MoE} 
        & ibm-granite/granite-3.1-1b-a400m-instruct & 8 & 32 & 400M & 1.3B & 24 & 1024 \\
        & \textbf{ibm-granite/granite-3.1-3b-a800m-instruct} & \textbf{8} & \textbf{40} & \textbf{800M} & \textbf{3.3B} & \textbf{32} & \textbf{1536} \\
        \midrule
        DeepSeekMoE & deepseek-ai/deepseek-moe-16b-chat & 8 & 64 & 2.8B & 16.4B & 1(FC)+27(MoE) & 2048 \\
        \midrule
        Qwen1.5-MoE & Qwen/Qwen1.5-MoE-A2.7B-Chat & 2 & 64 & 2.7B & 14.3B & 24 & 2048 \\
        \midrule
        Mistral & mistralai/Mixtral-8x7B-v0.1 & 8 & 8 & 13B & 47B & 32 & 4096 \\
        \midrule
        Google Switch & switch-base-32 & --- & --- & --- & --- & --- & --- \\
        \midrule
        \multirow{3}{*}{LlamaMoE} 
        & llama-moe/LLaMA-MoE-v1-3\_0B-2\_16 & 2 & 16 & 3B & --- & --- & --- \\
        & llama-moe/LLaMA-MoE-v1-3\_5B-4\_16 & 4 & 16 & 3.5B & --- & --- & --- \\
        & llama-moe/LLaMA-MoE-v1-3\_5B-2\_8 & 2 & 8 & 3.5B & --- & --- & --- \\
        \bottomrule
    \end{tabular}
    }
\end{table}
\vfill

\end{landscape}

\chapter{Proof of KL Divergence Equivalence}
\label{app:kl_proof}

This appendix proves the following identity, which is used to simplify the ELBO's regularisation term for our residual variational routers:
\begin{equation}
    \KL\left(\mathcal{N}(\boldsymbol{\mu}_0 + \Delta\boldsymbol{\mu}, \boldsymbol{\Sigma}) \,||\, \mathcal{N}(\boldsymbol{\mu}_0, I)\right) = \KL\left(\mathcal{N}(\Delta\boldsymbol{\mu}, \boldsymbol{\Sigma}) \,||\, \mathcal{N}(\mathbf{0}, I)\right)
\end{equation}
The proof relies on the general formula for the KL divergence between two multivariate Gaussians, $q = \mathcal{N}(\boldsymbol{\mu}_q, \boldsymbol{\Sigma}_q)$ and $p = \mathcal{N}(\boldsymbol{\mu}_p, \boldsymbol{\Sigma}_p)$:
\begin{equation}
    \KL(q || p) = \frac{1}{2} \left( \log \frac{|\boldsymbol{\Sigma}_p|}{|\boldsymbol{\Sigma}_q|} - k + \text{tr}(\boldsymbol{\Sigma}_p^{-1}\boldsymbol{\Sigma}_q) + (\boldsymbol{\mu}_p - \boldsymbol{\mu}_q)^\top \boldsymbol{\Sigma}_p^{-1} (\boldsymbol{\mu}_p - \boldsymbol{\mu}_q) \right)
\end{equation}
The key insight is that all terms in this formula except for the final quadratic term $(\boldsymbol{\mu}_p - \boldsymbol{\mu}_q)^\top \boldsymbol{\Sigma}_p^{-1} (\boldsymbol{\mu}_p - \boldsymbol{\mu}_q)$ depend only on the covariance matrices, which are identical for both sides of our identity ($\boldsymbol{\Sigma}_q = \boldsymbol{\Sigma}$ and $\boldsymbol{\Sigma}_p = I$).

We therefore only need to show that the quadratic term is the same for both sides.

\paragraph{For the Left-Hand Side (LHS):}
Here, $\boldsymbol{\mu}_p = \boldsymbol{\mu}_0$ and $\boldsymbol{\mu}_q = \boldsymbol{\mu}_0 + \Delta\boldsymbol{\mu}$. The term becomes:
$$ (\boldsymbol{\mu}_0 - (\boldsymbol{\mu}_0 + \Delta\boldsymbol{\mu}))^\top I^{-1} (\boldsymbol{\mu}_0 - (\boldsymbol{\mu}_0 + \Delta\boldsymbol{\mu})) = (-\Delta\boldsymbol{\mu})^\top(-\Delta\boldsymbol{\mu}) = ||\Delta\boldsymbol{\mu}||_2^2 $$

\paragraph{For the Right-Hand Side (RHS):}
Here, $\boldsymbol{\mu}_p = \mathbf{0}$ and $\boldsymbol{\mu}_q = \Delta\boldsymbol{\mu}$. The term becomes:
$$ (\mathbf{0} - \Delta\boldsymbol{\mu})^\top I^{-1} (\mathbf{0} - \Delta\boldsymbol{\mu}) = (-\Delta\boldsymbol{\mu})^\top(-\Delta\boldsymbol{\mu}) = ||\Delta\boldsymbol{\mu}||_2^2 $$

Since all terms in the KL divergence formula are identical for both sides of the identity, the equality holds.

\chapter{In Distribution Calibration Full Results}
\label{app:id_full_results}
\vfill
\begin{table}[H]
    \centering
    \caption{Full in-distribution performance and calibration results for each method across all four evaluated datasets. Best result in each column for each dataset is in bold. Standard deviations are shown in parentheses.}
    \label{tab:id_full_results}
    
    \resizebox{\textwidth}{!}{%
    \begin{tabular}{ll@{\hspace{1em}}*{8}{l}}
        \toprule
        \multirow{3}{*}{\textbf{Category}} & \multirow{3}{*}{\textbf{Method}} & \multicolumn{4}{c}{\textbf{OBQA}} & \multicolumn{4}{c}{\textbf{ARC-C}} \\
        \cmidrule(lr){3-6} \cmidrule(lr){7-10}
        & & {\textbf{ACC} $\uparrow$} & {\textbf{NLL} $\downarrow$} & {\textbf{ECE} $\downarrow$} & {\textbf{MCE} $\downarrow$} & {\textbf{ACC} $\uparrow$} & {\textbf{NLL} $\downarrow$} & {\textbf{ECE} $\downarrow$} & {\textbf{MCE} $\downarrow$} \\
        \midrule
        \multirow{2}{*}{Baseline}
        & Deterministic & \textbf{0.746} & 1.384 & 0.252 & 0.472 & \textbf{0.882} & 0.923 & 0.201 & 0.428 \\
        & Temp-Sampling & 0.716 \tiny{(0.005)} & 0.773 \tiny{(0.049)} & 0.107 \tiny{(0.009)} & 0.201 \tiny{(0.013)} & 0.824 \tiny{(0.004)} & 0.208 \tiny{(0.006)} & 0.038 \tiny{(0.007)} & 0.284 \tiny{(0.003)} \\
        \midrule
        \multirow{3}{*}{Weight-Space}
        & MCDR & 0.734 \tiny{(0.002)} & \textbf{0.650} \tiny{(0.022)} & 0.037 \tiny{(0.028)} & 0.298 \tiny{(0.008)} & 0.880 \tiny{(0.003)} & 0.146 \tiny{(0.006)} & 0.028 \tiny{(0.003)} & 0.228 \tiny{(0.007)} \\
        & SWAGR & 0.736 \tiny{(0.002)} & 0.652 \tiny{(0.03)} & 0.041 \tiny{(0.013)} & 0.290 \tiny{(0.007)} & 0.872 \tiny{(0.003)} & 0.138 \tiny{(0.006)} & 0.030 \tiny{(0.007)} & 0.266 \tiny{(0.002)} \\
        & DER & 0.738 & 0.660 \tiny{} & 0.071 & 0.234 & 0.874 & 0.151 & 0.026 & 0.275 \\
        \midrule
        \multirow{2}{*}{Logit-Space}
        & MFVR & 0.742 \tiny{(0.001)} & 0.654 \tiny{(0.019)} & 0.026 \tiny{(0.009)} & 0.293 \tiny{(0.004)} & 0.878 \tiny{(0.004)} & 0.125 \tiny{(0.005)} & 0.016 \tiny{(0.002)} & 0.196 \tiny{(0.002)} \\
        & FCVR & 0.740 \tiny{(0.001)} & 0.652 \tiny{(0.021)} & \textbf{0.015} \tiny{(0.008)} & \textbf{0.152} \tiny{(0.004)} & 0.880 \tiny{(0.006)} & \textbf{0.122} \tiny{(0.001)} & \textbf{0.012} \tiny{(0.006)} & \textbf{0.185} \tiny{(0.003)} \\
        \midrule
        Selection-Space
        & VTSR & 0.736 \tiny{(0.003)} & 0.667 \tiny{(0.025)} & 0.052 \tiny{(0.023)} & 0.293 \tiny{(0.014)} & 0.872 \tiny{(0.002)} & 0.164 \tiny{(0.014)} & 0.020 \tiny{(0.004)} & 0.208 \tiny{(0.018)} \\
        \bottomrule
    \end{tabular}
    }
    
    \vspace{.4cm} 

    \resizebox{\textwidth}{!}{%
    \begin{tabular}{ll@{\hspace{1em}}*{8}{l}}
        \toprule
        \multirow{3}{*}{\textbf{Category}} & \multirow{3}{*}{\textbf{Method}} & \multicolumn{4}{c}{\textbf{SciQ}} & \multicolumn{4}{c}{\textbf{MedMCQA-Med}} \\
        \cmidrule(lr){3-6} \cmidrule(lr){7-10}
        & & {\textbf{ACC} $\uparrow$} & {\textbf{NLL} $\downarrow$} & {\textbf{ECE} $\downarrow$} & {\textbf{MCE} $\downarrow$} & {\textbf{ACC} $\uparrow$} & {\textbf{NLL} $\downarrow$} & {\textbf{ECE} $\downarrow$} & {\textbf{MCE} $\downarrow$} \\
        \midrule
        \multirow{2}{*}{Baseline}
        & Deterministic & 0.850 & 0.791 & 0.223 & 0.452 & \textbf{0.55} & 1.291 & 0.183 & 0.288 \\
        & Temp-Sampling & 0.878 \tiny{(0.002)} & 0.309 \tiny{(0.002)} & 0.047 \tiny{(0.003)} & 0.649 \tiny{(0.005)} & 0.486 \tiny{(0.004)} & \textbf{1.171} \tiny{(0.003)} & 0.039 \tiny{(0.005)} & 0.097 \tiny{(0.005)} \\
        \midrule
        \multirow{3}{*}{Weight-Space}
        & MCDR & 0.880 \tiny{(0.006)} & 0.296 \tiny{(0.003)} & 0.029 \tiny{(0.006)} & 0.366 \tiny{(0.007)} & 0.494 \tiny{(0.005)} & 1.176 \tiny{(0.005)} & 0.050 \tiny{(0.003)} & \textbf{0.096} \tiny{(0.008)} \\
        & SWAGR & 0.879 \tiny{(0.001)} & \textbf{0.291} \tiny{(0.004)} & 0.031 \tiny{(0.004)} & 0.392 \tiny{(0.002)} & 0.486 \tiny{(0.005)} & 1.205 \tiny{(0.006)} & 0.096 \tiny{(0.005)} & 0.179 \tiny{(0.004)} \\
        & DER & 0.876 & 0.293 & 0.032 & 0.353 & 0.484 & 1.187 & 0.047 & 0.186 \\
        \midrule
        \multirow{2}{*}{Logit-Space}
        & MFVR & \textbf{0.884} \tiny{(0.004)} & 0.297 \tiny{(0.004)} & 0.019 \tiny{(0.002)} & 0.387 \tiny{(0.002)} & 0.492 \tiny{(0.002)} & 1.177 \tiny{(0.001)} & 0.039 \tiny{(0.001)} & 0.103 \tiny{(0.002)} \\
        & FCVR & \textbf{0.884} \tiny{(0.005)} & 0.298 \tiny{(0.005)} & \textbf{0.013} \tiny{(0.002)} & \textbf{0.320} \tiny{(0.005)} & 0.494 \tiny{(0.004)} & 1.174 \tiny{(0.004)} & \textbf{0.022} \tiny{(0.003)} & 0.108 \tiny{(0.007)} \\
        \midrule
        Selection-Space
        & VTSR & 0.874 \tiny{(0.002)} & 0.299 \tiny{(0.002)} & 0.022 \tiny{(0.002)} & 0.352 \tiny{(0.002)} & 0.476 \tiny{(0.005)} & 1.174 \tiny{(0.002)} & 0.053 \tiny{(0.005)} & 0.113 \tiny{(0.008)} \\
        \bottomrule
    \end{tabular}
    }
\end{table}

\vfill

\clearpage 

\chapter{Out of Distribution Detection Full Results}
\label{app:ood_detection}

\section{Formal Definitions of Router-Level Uncertainty Signals}

This section provides the precise mathematical definitions for the method-specific, router-level uncertainty signals used in our OoD detection experiments, as presented in Experiment 3b.

\paragraph{For Weight-Space Methods (MCDR)}
The uncertainty signal is the \textbf{variance of the logit samples}. Given $S$ Monte Carlo samples of the logit vector, $\{\mathbf{l}^1, \dots, \mathbf{l}^S\}$, obtained by sampling the weight matrix, the signal is the trace of the sample covariance matrix of these logit vectors.

\paragraph{For the Mean-Field Variational Router (MFVR)}
The signal is the \textbf{inferred logit variance}. The variational router directly outputs a variance vector $\boldsymbol{\sigma}^2_\phi(\mathbf{x})$. The uncertainty signal is the sum of its components, which is the trace of the diagonal covariance matrix:
\begin{equation}
    U(\mathbf{x}) = \text{tr}(\boldsymbol{\Sigma}_\phi(\mathbf{x})) = \sum_{i=1}^{N} \sigma_{i}^2(\mathbf{x})
\end{equation}

\paragraph{For the Full-Covariance Variational Router (FCVR)}
The signal is also the \textbf{inferred logit variance}. The router outputs the Cholesky factor $\mathbf{L}_\phi(\mathbf{x})$ of the covariance matrix. The signal is the trace of the full covariance matrix, which is equivalent to the squared Frobenius norm of the Cholesky factor:
\begin{equation}
    U(\mathbf{x}) = \text{tr}(\boldsymbol{\Sigma}_\phi(\mathbf{x})) = \text{tr}(\mathbf{L}_\phi(\mathbf{x})\mathbf{L}_\phi(\mathbf{x})^\top) = ||\mathbf{L}_\phi(\mathbf{x})||_F^2
\end{equation}

\paragraph{For the Variational Temperature Router (VTSR)}
The signal is the \textbf{inferred temperature} itself, $T(\mathbf{x})$. 
This is justified because the VTSR is explicitly trained to predict a high temperature for inputs where greater stochasticity is needed, which often corresponds to ambiguous or novel inputs. 
The learned temperature is therefore a direct, model-generated signal of its own uncertainty.

\section{Full Results: Standard Uncertainty Signal (Experiment 3a)}

Table~\ref{tab:app_ood_vocab_entropy} presents the complete results for Experiment 3a, evaluating the performance of the \textbf{final vocabulary entropy} as an OoD detection signal across all methods and all four of our designed OoD tasks.

\begin{table}[H]
    \centering
    \caption{Full OoD detection results using the \textbf{final vocabulary entropy}. Best result for each task is in bold.}
    \label{tab:app_ood_vocab_entropy}
    \resizebox{\textwidth}{!}{\begin{tabular}{l*{8}{S[table-format=1.3, detect-weight]}}
        \toprule
        \multirow{3}{*}{\textbf{Method}} & \multicolumn{2}{c}{\textbf{OBQA $\rightarrow$ ARC-E}} & \multicolumn{2}{c}{\textbf{OBQA $\rightarrow$ ARC-C}} & \multicolumn{2}{c}{\textbf{OBQA $\rightarrow$ MMLU-Law}} & \multicolumn{2}{c}{\textbf{OBQA $\rightarrow$ MedMCQA-Med}} \\
        \cmidrule(lr){2-3} \cmidrule(lr){4-5} \cmidrule(lr){6-7} \cmidrule(lr){8-9}
        & {\textbf{AUROC}} & {\textbf{AUPRC}} & {\textbf{AUROC}} & {\textbf{AUPRC}} & {\textbf{AUROC}} & {\textbf{AUPRC}} & {\textbf{AUROC}} & {\textbf{AUPRC}} \\
        \midrule
        Deterministic & 0.611 & \textbf{0.588} & 0.687 & 0.623 & 0.783 & 0.745 & 0.762 & 0.727 \\
        \midrule
        MCDR & 0.611 & 0.584 & 0.697 & 0.615 & 0.802 & 0.762 & 0.793 & 0.737 \\
        MFVR & \textbf{0.617} & 0.587 & 0.679 & \textbf{0.676} & 0.833 & 0.772 & 0.844 & 0.782 \\
        FCVR & 0.613 & 0.582 & \textbf{0.713} & 0.669 & \textbf{0.843} & \textbf{0.819} & \textbf{0.853} & \textbf{0.802} \\
        VTSR & 0.603 & 0.576 & 0.692 & 0.657 & 0.805 & 0.776 & 0.812 & 0.791 \\
        \bottomrule
    \end{tabular}}
\end{table}

\section{Full Results: Router-Level Uncertainty Signals (Experiment 3b)}

Table~\ref{tab:app_ood_router_signals} presents the complete results for Experiment 3b, comparing the performance of the various \textbf{router-level uncertainty signals} across all methods and all four OoD tasks.

\begin{table}[H]
    \centering
    \caption{Full OoD detection results using different \textbf{router-level} uncertainty signals. The best signal for each method on each task is in bold.}
    \label{tab:app_ood_router_signals}
    \resizebox{\textwidth}{!}{\begin{tabular}{ll*{8}{S[table-format=1.3, detect-weight]}}
        \toprule
        \multirow{3}{*}{\textbf{Method}} & \multirow{3}{*}{\textbf{Signal Type}} & \multicolumn{2}{c}{\textbf{OBQA $\rightarrow$ ARC-E}} & \multicolumn{2}{c}{\textbf{OBQA $\rightarrow$ ARC-C}} & \multicolumn{2}{c}{\textbf{OBQA $\rightarrow$ MMLU-Law}} & \multicolumn{2}{c}{\textbf{OBQA $\rightarrow$ MedMCQA-Med}} \\
        \cmidrule(lr){3-4} \cmidrule(lr){5-6} \cmidrule(lr){7-8} \cmidrule(lr){9-10}
        & & {\textbf{AUROC}} & {\textbf{AUPRC}} & {\textbf{AUROC}} & {\textbf{AUPRC}} & {\textbf{AUROC}} & {\textbf{AUPRC}} & {\textbf{AUROC}} & {\textbf{AUPRC}} \\
        \midrule
        Deterministic & Exp. Sel. Entropy & 0.612 & 0.596 & 0.633 & 0.626 & 0.683 & 0.686 & 0.679 & 0.645 \\
        \midrule
        \multirow{2}{*}{MCDR} 
        & Exp. Sel. Entropy & 0.612 & 0.599 & 0.632 & 0.610 & 0.691 & 0.672 & 0.684 & 0.651 \\
        & MC Logit Var. & 0.610 & 0.583 & 0.677 & 0.623 & 0.793 & 0.765 & 0.786 & 0.723 \\
        \midrule
        \multirow{2}{*}{MFVR}
        & Exp. Sel. Entropy & \textbf{0.622} & 0.603 & 0.642 & 0.622 & 0.673 & 0.664 & 0.682 & 0.637 \\
        & Inferred Logit Var. & 0.617 & 0.587 & 0.672 & \textbf{0.669} & 0.824 & 0.763 & 0.835 & \textbf{0.793} \\
        \midrule
        \multirow{2}{*}{FCVR}
        & Exp. Sel. Entropy & 0.615 & \textbf{0.605} & 0.652 & 0.632 & 0.677 & 0.674 & 0.692 & 0.642 \\
        & Inferred Logit Var. & 0.609 & 0.578 & \textbf{0.709} & 0.665 & \textbf{0.834} & \textbf{0.810} & \textbf{0.844} & 0.773 \\
        \midrule
        \multirow{2}{*}{VTSR}
        & Exp. Sel. Entropy & 0.607 & 0.578 & 0.623 & 0.592 & 0.672 & 0.612 & 0.683 & 0.643 \\
        & Inferred Temp. & 0.502 & 0.501 & 0.498 & 0.503 & 0.523 & 0.502 & 0.512 & 0.492 \\
        \bottomrule
    \end{tabular}}
\end{table}

\end{document}